\documentclass[11pt, a4paper, logo, copyright]{googlecloud}

\pdftrailerid{redacted}

\makeatletter
\renewcommand\bibentry[1]{\nocitep{#1}{\frenchspacing\@nameuse{BR@r@#1\@extra@b@citeb}}}
\makeatother

\usepackage{xcolor, soul,todonotes} 
\definecolor{kmycolor}{rgb}{0.858, 0.188, 0.478}

\definecolor{llcolor}{rgb}{0.0, 0.5, 0.5}

\definecolor{yscolor}{rgb}{1.0, 0.6, 0.0}

\definecolor{dxlcolor}{rgb}{0.0, 0.0, 0.0}

\usepackage{kantlipsum, lipsum}
\usepackage{dsfont}

\definecolor{ourred}{HTML}{F19C99}
\definecolor{ourblue}{HTML}{7EA6E0}
\definecolor{plotred}{HTML}{f77189}
\definecolor{plotgreen}{HTML}{33b07a}
\definecolor{plotpurple}{HTML}{cc7af4}
\definecolor{plotmagenta}{HTML}{f565cc}
\definecolor{plotazure}{HTML}{38a9c5}

\definecolor{codegreen}{rgb}{0,0.6,0}
\definecolor{codegray}{rgb}{0.5,0.5,0.5}
\definecolor{codepurple}{rgb}{0.58,0,0.82}
\definecolor{backcolour}{rgb}{0.95,0.95,0.92}

\definecolor{Gray}{gray}{0.92}

\definecolor{stage1}{HTML}{34A853}
\definecolor{stage2}{HTML}{A680B8}
\definecolor{stage3}{HTML}{009999}

\usepackage{listings}

\lstdefinestyle{mystyle}{
    backgroundcolor=\color{backcolour},   
    commentstyle=\color{codegreen},
    keywordstyle=\color{magenta},
    numberstyle=\tiny\color{codegray},
    stringstyle=\color{codepurple},
    basicstyle=\ttfamily\scriptsize,
    breakatwhitespace=false,         
    breaklines=true,                 
    captionpos=b,                    
    keepspaces=true,                 
    numbersep=5pt,                  
    showspaces=false,                
    showstringspaces=false,
    showtabs=false,                  
    tabsize=2,
    frame=none,
    aboveskip=1pt,
    belowskip=1pt,
}
\usepackage[most]{tcolorbox}
\tcbset{enhanced,
boxrule=0pt,frame hidden,
borderline west={4pt}{0pt}{green!75!black},
top=0pt,bottom=0pt,
colback=green!10!white,
sharp corners}

\usepackage{colortbl}
\usepackage{float}

\usepackage{multirow}

\usepackage[authoryear, sort&compress, round]{natbib}

\usepackage[utf8]{inputenc} %
\usepackage{lmodern}
\usepackage[T1]{fontenc}    %
\usepackage{hyperref}       %
\hypersetup{
  colorlinks,
  citecolor=blue,
  linkcolor=red,
  urlcolor=blue}
\usepackage{url}            %
\usepackage{booktabs}       %
\usepackage{amsfonts}       %
\usepackage{nicefrac}       %
\usepackage{microtype}      %
\usepackage{wrapfig} %
\usepackage{graphicx} %
\usepackage{soul} %
\usepackage{enumitem}
\usepackage{amssymb,amsmath}
\usepackage{tikz, adjustbox}
\usepackage[most]{tcolorbox}
\usepackage{subcaption}
\usepackage{multirow}
\usepackage{arydshln}
\usepackage{float}
\usepackage[capitalize,noabbrev,nameinlink]{cleveref}
\usepackage{fancyvrb}

\usepackage{tcolorbox}

\usepackage{amsmath,amsfonts,bm}

\def\eqref#1{equation~\ref{#1}}

\def\1{\bm{1}}

\DeclareMathAlphabet{\mathsfit}{\encodingdefault}{\sfdefault}{m}{sl}
\SetMathAlphabet{\mathsfit}{bold}{\encodingdefault}{\sfdefault}{bx}{n}
 \usepackage{amsthm}
\usepackage{algorithm}
\usepackage{algpseudocode}
\usepackage{amsmath}
\usepackage{xcolor} 
\definecolor{darkgreen}{rgb}{0.0, 0.5, 0.0}
\usepackage{enumitem}
\setlist[itemize]{align=parleft,left=0pt..1em}
\usepackage{multirow}
\usepackage{media9} % For embedding video
\algrenewcommand{\algorithmiccomment}[1]{\textcolor{gray}{\hfill$\triangleright$ #1}}
\usepackage{booktabs}
\usepackage{graphicx}
\usepackage{booktabs}
\usepackage{subcaption}  % for subfigure environment (\begin{subfigure})
\usepackage{wrapfig}
\usepackage{arydshln}
\usepackage{tabularx}
\usepackage{caption}
\usepackage[most]{tcolorbox}
\usepackage{courier} % for fixed-width font
\tcbuselibrary{skins}

\theoremstyle{definition}

\usepackage{caption}
\usepackage{soul}

\usepackage{hyperref}

\tcbuselibrary{listingsutf8}
\usepackage{listings}
\usepackage{xcolor}
\UseRawInputEncoding
\definecolor{lavender}{RGB}{242, 242, 255}  % Light purple tone
\everymath{\displaystyle} % Force all math to be displaystyle

\usepackage{graphicx}       % For including images
\usepackage{subcaption}     % For subfigures
\usepackage{caption}        % For figure captions
\usepackage{ragged2e}       % For text alignment (e.g., \justify)
\usepackage{tikz}           % For drawing the timeline, though simpler text is fine
\usepackage{amsmath}        % For math environments, just in case (not strictly needed here)
\crefname{algorithm}{Alg.}{Algs.}
\Crefname{algorithm}{Alg.}{Algs.}

\usepackage{ragged2e} % for \RaggedRight
\definecolor{lightgray}{gray}{0.95}
\definecolor{myblue}{rgb}{0.1,0.1,0.6}

\usepackage{pifont}
\newcommand{\cmark}{\textcolor{green!60!black}{\ding{51}}} 
\newcommand{\xmark}{\textcolor{red!70!black}{\ding{55}}}

\usepackage{bbold}

\usepackage{listings}

\lstdefinestyle{mystyle}{
    backgroundcolor=\color{backcolour},   
    commentstyle=\color{codegreen},
    keywordstyle=\color{magenta},
    numberstyle=\tiny\color{codegray},
    stringstyle=\color{codepurple},
    basicstyle=\ttfamily\scriptsize,
    breakatwhitespace=false,         
    breaklines=true,                 
    captionpos=b,                    
    keepspaces=true,                 
    numbers=left,                    
    numbersep=5pt,                  
    showspaces=false,                
    showstringspaces=false,
    showtabs=false,                  
    tabsize=2,
    frame=none,
    aboveskip=1pt,
    belowskip=1pt,
}
\lstdefinestyle{plainins}{
    backgroundcolor=\color{white},   
    commentstyle=\color{codegreen},
    keywordstyle=\color{magenta},
    numberstyle=\tiny\color{codegray},
    stringstyle=\color{codepurple},
    basicstyle=\ttfamily\scriptsize,
    breakatwhitespace=false,         
    breaklines=true,                 
    captionpos=b,                    
    keepspaces=true,                 
    numbers=none,                    
    numbersep=5pt,                  
    showspaces=false,                
    showstringspaces=false,
    showtabs=false,                  
    tabsize=2,
    aboveskip=0pt,
    belowskip=0pt,
    frame=single
}
\lstdefinestyle{plainexam}{
    backgroundcolor=\color[HTML]{FFFCF3},   
    commentstyle=\color{codegreen},
    keywordstyle=\color{magenta},
    numberstyle=\tiny\color{codegray},
    stringstyle=\color{codepurple},
    basicstyle=\ttfamily\scriptsize,
    breakatwhitespace=false,         
    breaklines=true,                 
    captionpos=b,                    
    keepspaces=true,                 
    numbers=none,                    
    numbersep=5pt,                  
    showspaces=false,                
    showstringspaces=false,
    showtabs=false,                  
    tabsize=2,
    aboveskip=0pt,
    belowskip=0pt
}

\newcommand{\model}{\textsc{A${^2}$RD}}
\newcommand{\memory}{\textsc{MVMem}}
\newcommand{\dataset}{\textsc{LVbench-C}}

\title{\model{}: Agentic Autoregressive Diffusion for Long Video Consistency}
\correspondingauthor{Do Xuan Long: xuanlong.do@u.nus.edu; Yale Song, Long T. Le: \{yalesong, longtle\}@google.com}

\renewcommand{\today}{}

\author[1 2 *]{Do Xuan Long}
\author[1]{Yale Song}
\author[2]{Min-Yen Kan}
\author[1]{Tomas Pfister}
\author[1]{Long T. Le}
\affil[1]{Google Cloud AI Research}
\affil[2]{National University of Singapore}
\begin{document}

\begin{abstract}
% \ll{Suggested title: LongVeo: Agentic Autoregressive Video Generation with Multimodal Memory and Test-Time Scaling}
% \kmy{Consider starting with this sentence, the first two are motivation and not needed in an abstract.} 
% Existing methods suffer from semantic drift, where entities and environments change unintentionally, or content collapse, where narratives fail to progress meaningfully. 

Synthesizing consistent and coherent long video remains a fundamental challenge. Existing methods suffer from semantic drift and narrative collapse over long horizons. We present \textbf{\model{}}, an {A}gentic {A}uto-{R}egressive {D}iffusion architecture that decouples creative synthesis from consistency enforcement. \model{} formulates long video synthesis as a \emph{closed-loop} process that synthesizes and self-improves video segment-by-segment through a \emph{Retrieve--Synthesize--Refine--Update} cycle. It comprises three core components: \emph{(i) Multimodal Video Memory} that tracks video progression across modalities; \emph{(ii) Adaptive Segment Generation} that switches among generation modes for natural progression and visual consistency; and {\emph{(iii) Hierarchical Test-Time Self-Improvement}} that self-improves each segment at frame and video levels to prevent error propagation. We further introduce \textbf{$\dataset{}$}, a challenging benchmark with non-linear entity and environment transitions {to stress-test long-horizon consistency. Across public and $\dataset{}$ benchmarks spanning one- to ten-minute videos, $\model{}$ outperforms state-of-the-art baselines by up to 30\% in consistency and 20\% in narrative coherence. Human evaluations corroborate these gains while also highlighting notable improvements in motion and transition smoothness.}
\end{abstract}

\maketitle

%%%%%%%%%%%%%%%%%%%%%%%%%%%%%
\begin{figure*}[h!]
    \centering
\includegraphics[width=1\linewidth]{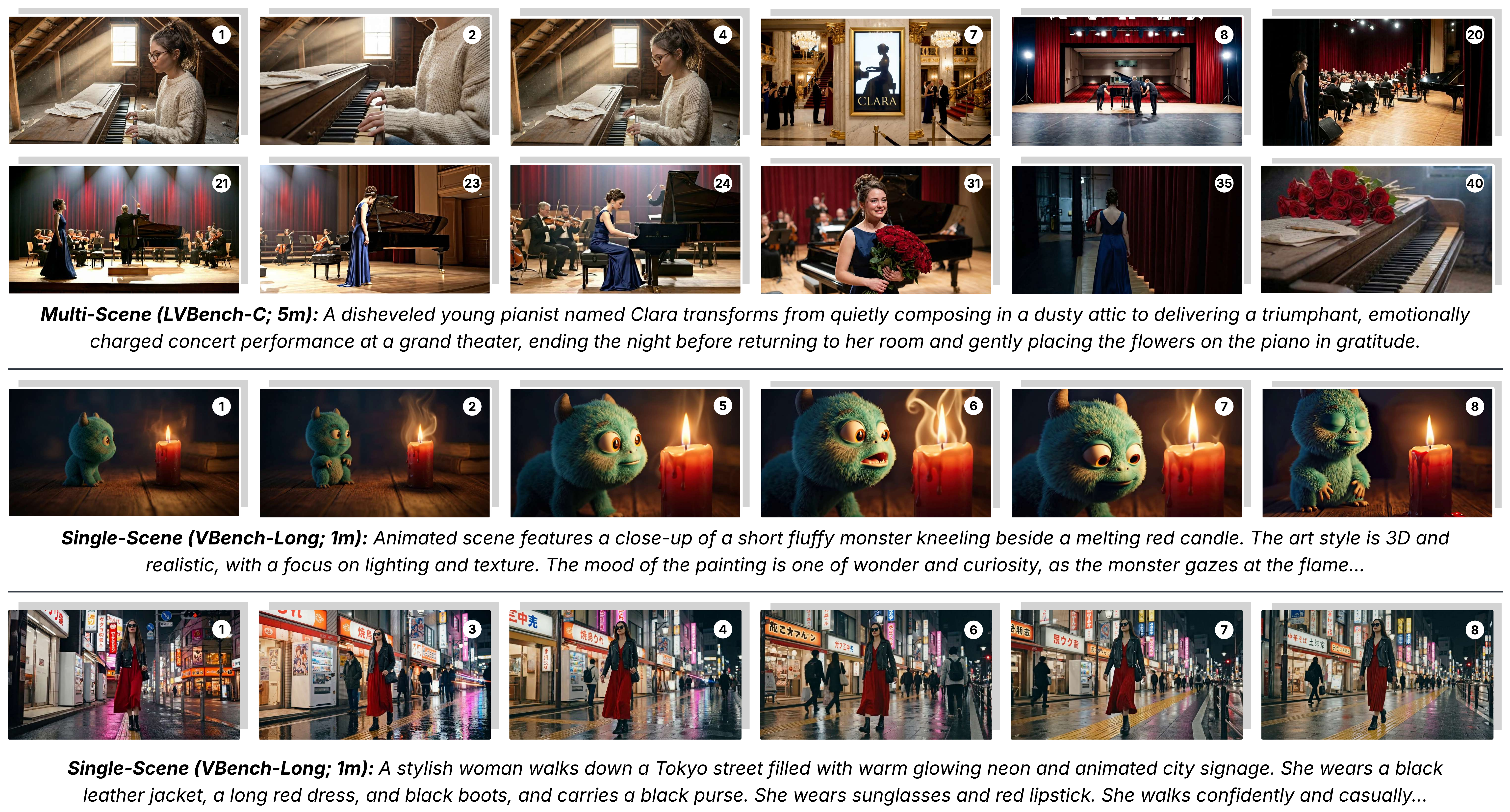} 
\caption{Examples of 1m and 5m videos generated by \model{} with Veo 3.1, showcasing consistent and coherent stories in static, dynamic, and multi-shot environments. Full and more storyboards are provided in \Cref{fig:baseline-examples-progression,fig:baseline-examples-consistency,sec:storyboard-examples}, with videos in the supplementary materials.}
% \vspace{-5mm}
\label{fig:teaser-figure}
\end{figure*}
%%%%%%%%%%%%%%%%%%%%%%%%%%%%%

\centerline{\small{\textbf{Project Page:} \url{http://dxlong2000.github.io/AARD}}}

\section{Introduction}
% \kmy{Do you really want to call this ``ultra-long''?  Later the technology can generate movie length outputs or longer?}
% \kmy{video generation -- need the accompanying videos and website to link.}

Video synthesis has emerged as a transformative capability in artificial intelligence, powering high-impact applications including cinematic storytelling, educational content, entertainment, and advertising \citep{elmoghany2025survey,ma2025controllable}. Although recent breakthroughs in diffusion models \citep{ho2022video,singer2022make,esser2023structure,brooks2024sora,wan2025wan,deepmind2025veo3,seedance2} have achieved remarkable fidelity for second-long clips, real-world applications demand minute- to hour-long videos. Scaling to coherent long video synthesis, however, remains a fundamental challenge. At its core are two fundamental problems: \emph{temporal consistency}, which requires models to track and preserve entities, environments, and motion dynamics, and \emph{narrative coherence}, which demands that videos evolve meaningfully over time.

\begin{figure*}[t!]
    \centering
\includegraphics[width=.95\linewidth]{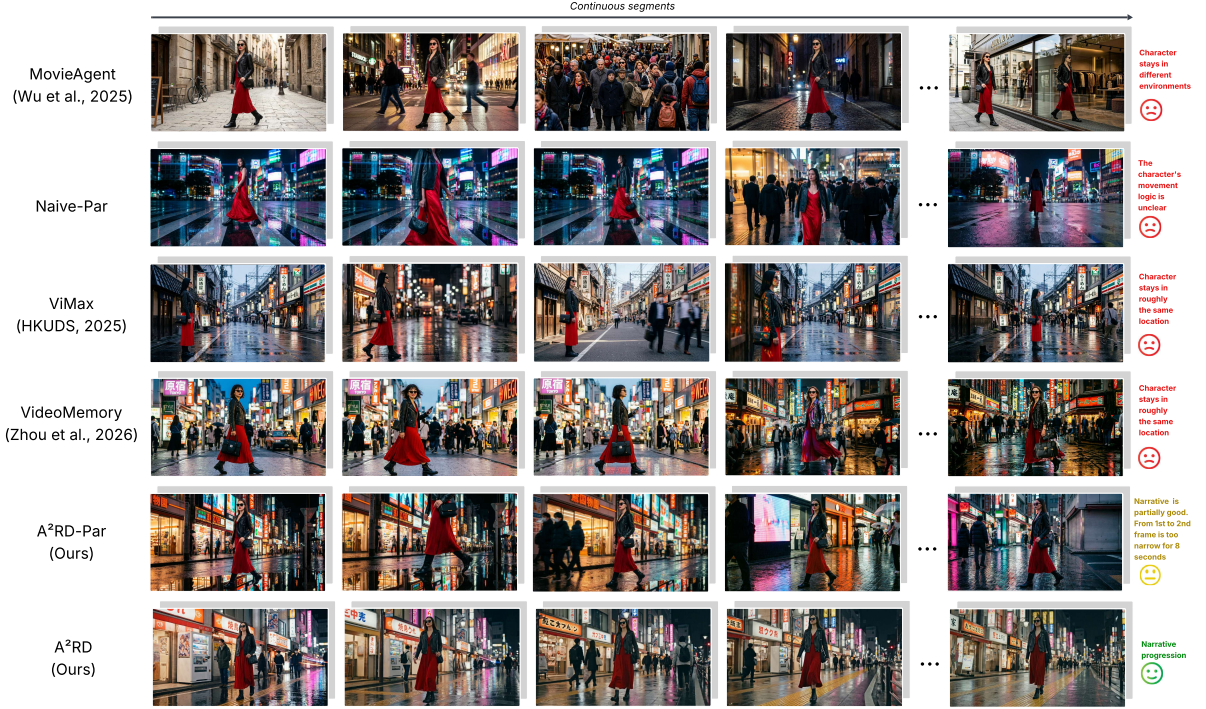} 
\caption{Narrative progression: a VBench-Long sample depicting a woman walking on a Japanese street. \model{} maintains coherent and continue story, while baselines exhibit poor entity and environment progression.}   
\label{fig:baseline-examples-progression}
\end{figure*}

\begin{figure*}[t!]
    \centering
\includegraphics[width=.95\linewidth]{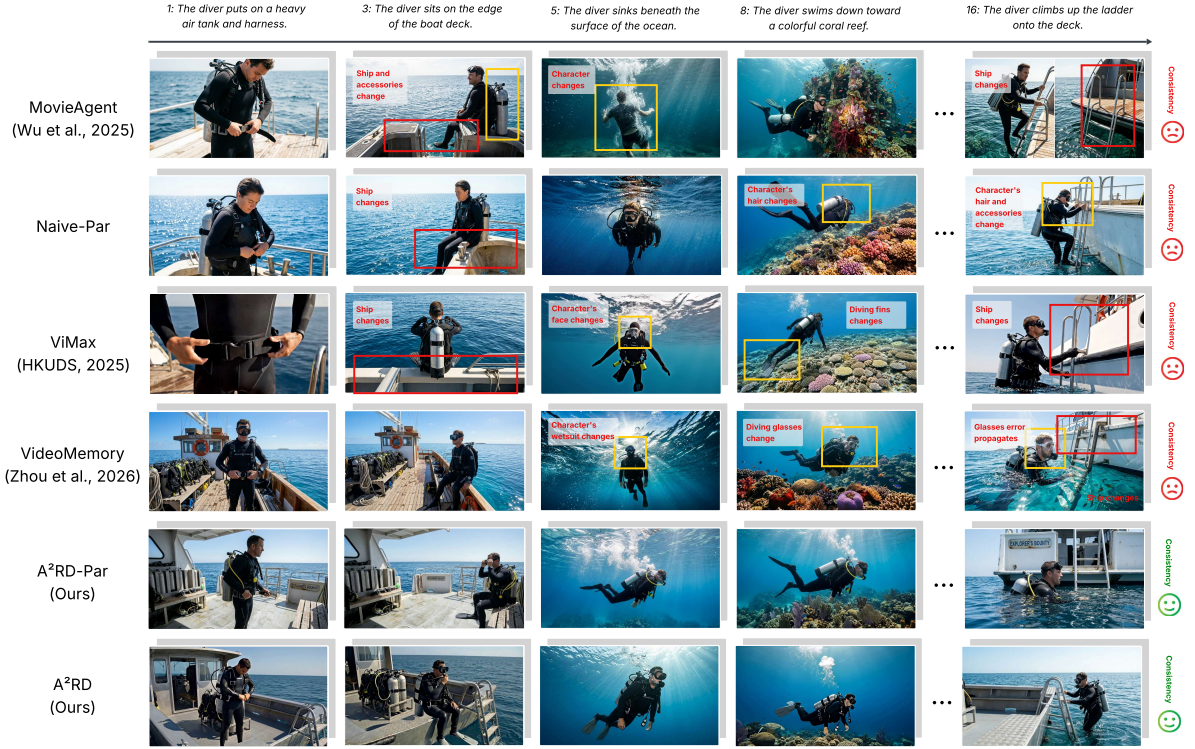} 
\caption{Long-horizon consistency: a \dataset{} sample depicting a diver preparing, diving, and returning to the ship. \model{} maintains rigorous consistency while baselines suffer from unintended drifts in environments (red) and entities (yellow), such as changing character's hair, face, accessories and ship layouts.} 
% \kmy{The red is too hard to read in certain cases with low contrast.  Consider adding an semi-opaque background for the text to be readable.}}  
\label{fig:baseline-examples-consistency}
\end{figure*}

State-of-the-art long video synthesis approaches follow the dominant \emph{passive, open-loop} paradigm, yet they have limitations. \emph{Frame-based autoregressive (FAR)} models synthesize videos frame-by-frame or chunk-by-chunk \citep{huang2025self,yang2025longlive,chen2026context}, {naturally preserving local temporal continuity.} However, once a frame is generated, it is frozen as fixed conditioning for all subsequent generation, {causing errors to propagate uncorrected and limiting narrative controllability}. This often leads to \emph{semantic drift} and \emph{narrative repetition} over long horizons \citep{zhao2026ultravico}. \emph{Segment-based} methods {synthesize and concatenate short segments, either autoregressively (SAR) \citep{zhou2026videomemory,an2025onestory,zhang2025storymem} or in parallel \citep{wu2025automated,wang2025mavis}, offering stronger narrative control}. However, they struggle to maintain \emph{inter-segment consistency and continuity} \citep{elmoghany2025survey}. {While recent SAR methods employ frame-based memory to bridge segments, visual-only conditioning proves insufficient for reliably tracking entities and narratives, causing persistent consistency and coherence errors across segments} (\Cref{fig:baseline-examples-progression}, \ref{fig:baseline-examples-consistency}, \ref{sec:storyboard-examples}). 

We reframe long video synthesis as a \emph{closed-loop, agentic} process to address these limitations. Our resulting architecture, {\model{}} (/\textit{a:rd}/, {A}gentic {A}uto-{R}egressive {D}iffusion), enables video diffusion models to synthesize and self-improve long videos autoregressively, enforcing temporal consistency and narrative coherence over long horizons. See \Cref{fig:teaser-figure} for example videos. \model{} is training-free and built upon three pillars: a {Multimodal Video Memory}, an {Adaptive Segment Generation}, and {{Hierarchical Test-Time Self-Improvement}} algorithms. These components form a {Retrieve--Synthesize--Refine--Update} cycle: for each segment, the agent retrieves relevant video world contexts from memory, determines the segment generation mode (e.g., extrapolation or interpolation) adaptively, synthesizes boundary frames then video segment with hierarchical self-improvements applied at both frame and video levels, and finally updates the memory for subsequent generation. 

Moreover, existing benchmarks lack the realistic complexity of long-horizon narratives, where entities and environments undergo non-linear transitions. We contribute \dataset{}, a benchmark designed to challenge long-horizon consistency where entities and environments appear, disappear, and reappear {(\emph{``cyclic''})} with optional state changes. Extensive evaluations on \dataset{} and public benchmarks show that \model{} achieves state-of-the-art consistency and narrative coherence in just two self-improved iterations, corroborated by human studies. In summary, this paper contributes: 

\begin{itemize}
\item We introduce \model{}, the first agentic autoregressive architecture for long video synthesis that integrates multimodal memory, adaptive segment generation, and self-improvement to enforce temporal consistency and narrative coherence. \model{} significantly outperforms existing baselines, scaling to ultra-long video while substantially mitigating semantic drift and content collapse.
\item We contribute {\dataset{}}, a challenging benchmark evaluating long-horizon video consistency through cyclical entity and environment appearances with optional state evolutions.
% \yale{Not sure the ``cyclical'' concept is clear yet.}
\item We conduct extensive experiments to provide insights into \model{} and its key components.
\end{itemize}

%%%%%%%%%%%%%%%%%%%%%%%%%%%%%
%%%%%%%%%%%%%%%%%%%%%%%%%%%%%

%%%%%%%%%%%%%%%%%%%%%%%%%%%%%
%%%%%%%%%%%%%%%%%%%%%%%%%%%%%

\section{Related Work}
% \ll {I suggest to put after introduction. Is there any specific reason that you put here?}

\paragraph{Long-Form Video Synthesis.} 
{State-of-the-art (SOTA) long video synthesis approaches are passive, following two main paradigms. {Frame-based autoregressive} models generate frames or chunks autoregressively, conditioning each on prior content via rolling KV caches~\citep{huang2025self}, short-window attention with frame-level sinks~\citep{yang2025longlive}, or initial-frame anchoring~\citep{liu2025rolling}. While preserving local visual fidelity, they remain prone to semantic drift and content repetition, and offer limited controllability~\citep{zhao2026ultravico}. Segment-based methods synthesize segments in parallel, with \citep{wang2025multishotmaster,meng2025holocine} or without shared denoising \citep{yin2023nuwa,wu2025automated,wang2025mavis}, or autoregressively (SAR) \citep{an2025onestory,zhang2025storymem,zhou2026videomemory}. These offer finer narrative control but struggle with inter-segment consistency \citep{elmoghany2025survey}. Segments are typically synthesized via \emph{extrapolation} from a begin frame~\citep{wang2025mavis,an2025onestory,zhou2026videomemory} or \emph{interpolation} between planned boundaries~\citep{yin2023nuwa}. Yet, each has limitations: extrapolation often causes inconsistencies for details absent from the begin frame, while interpolation can yield unnatural progression from poorly planned boundary frames. Existing methods also lack mechanisms to correct such errors, causing them to propagate across segments (\Cref{fig:baseline-examples-consistency}, \Cref{sec:storyboard-examples}). \model{} addresses these limitations by coupling SAR with an adaptive generation strategy, multimodal memory for richer conditioning, and closed-loop self-improvement, achieving strong temporal consistency, and narrative controllability.}

\paragraph{Test-Time Scaling for Generative Models.} 
Test-time scaling (TTS) improves generation quality by investing additional computation during inference \citep{snell2024scaling}. For image, this includes best-of-N sampling \citep{zhang2025let}, iterative refinement \citep{zhuo2025reflection,qu2026scale}, prompt optimization \citep{wan2025maestro,wang2024discrete,manas2024improving}, and evolutionary search \citep{he2025scaling}. Video TTS methods have recently emerged \citep{ gao2025devil,long2025vista,huang2025planning,zhu2026brandfusion,yang2026spiral,hong2026comic}, primarily focusing on prompt optimization: RAPO \citep{gao2025devil} enriches prompts through retrieval-augmented refinement, VISTA \citep{long2025vista} employs multi-agent iterative planning and critique, and VQQA \citep{song2026vqqa} uses VLM-generated questions for closed-loop optimization. However, these methods operate on single-segment quality only and do not address inter-segment consistency or narrative progression across segments. \model{} introduces efficient test-time algorithms specifically targeting consistency and narrative coherence in multi-segment long video synthesis.

\paragraph{Memory for LLM Agents.} 
% \kmy{You should make it clear why memory is needed -- it is needed to address long-range consistency problems.  This is a natural reason why this comes third in your related work section}
Memory has become an important component in modern agentic systems, enabling agents to maintain long-range dependencies across sequential decisions \citep{ hu2025memory,zhang2025survey}. Current memories for LLM agents save information in diverse formats including text \citep{ packer2023memgpt,zhong2024memorybank}, hidden representations \citep{wang2025m+}, and graphs \citep{ chhikara2025mem0,xu2025mem}, and typically incorporate retrieval mechanisms (e.g., semantic search) alongside management strategies (e.g., updating). Memory construction for image and video synthesis has also been increasingly studied, where the memory is typically composed of images \citep{parmar2018image,zhang2025storymem,yu2025context,zhou2026videomemory}, image--text pairs \citep{ zhu2019dm}, and hidden representations \citep{zhu2025kv,cai2026mixture}. While image-based memories provide visual references, relying on the generative models to implicitly infer entity identity and narrative state is unreliable over long horizons. Representation-based memories offer seamless conditioning but lack interpretability for explicit consistency control. \model{} addresses both limitations with a multimodal memory that explicitly tracks fine-grained visual and narrative world progression across modalities, enabling targeted control over consistency and coherence.
% \kmy{Say more about what is good / bad about each of these approaches with respect to fidelity, use-cases and contrast this with what you need to do in this paper.}
%%%%%%%%%%%%%%%%%%%%%%%%%%%%%
\begin{figure*}[t!]
    \centering
\includegraphics[width=1\linewidth]{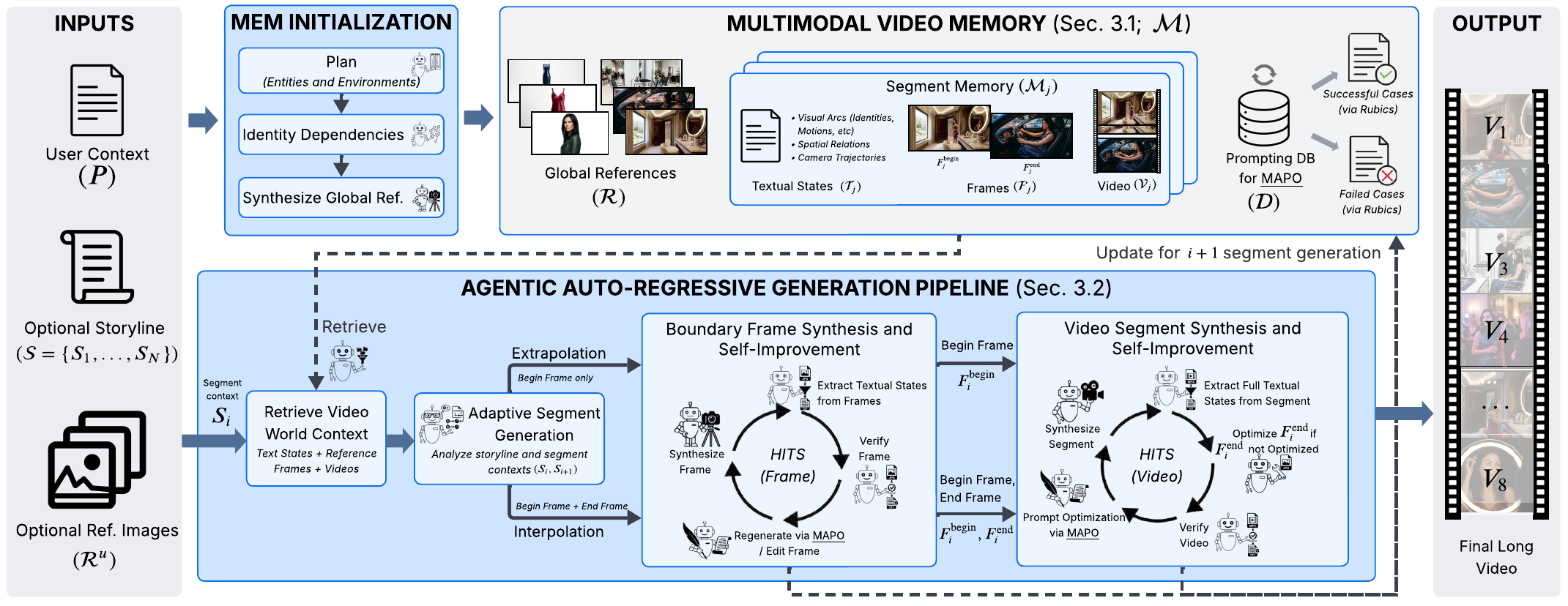} % Change to your file name
\caption{Overview of \model{} architecture. For each segment, \model{} retrieves relevant context from memory, adaptively determines the generation mode (extrap. or interp.), and synthesizes boundary frames followed by the video segment, with hierarchical self-improvements. Blue boxes denote methods implemented in \model{}.}
    \label{fig:method-overview}
\end{figure*}
%%%%%%%%%%%%%%%%%%%%%%%%%%%%%

%%%%%%%%%%%%%%%%%%%%%%%%%%%%%%%%%%
\section{\model{}: {A}gentic {A}uto-{R}egressive {D}iffusion} \label{sec:methods}

We present \textbf{\model{}} (\Cref{fig:method-overview}), an agentic segment-based autoregressive architecture for long video synthesis. We term our basic generation unit as \textbf{``segment''} (equivalent to ``clip''), {a flexible unit that can span one or more scenes or shots}. \model{} takes as input a user context $P$, a storyline $\mathcal{S} = \{S_1, \ldots, S_N\}$ (provided or planned from $P$) with $S_i$ being $i$-th segment context, and optional reference images $\mathcal{R}^u$. The agent supports any Text-Image-to-Video (TI2V) model via incorporating a Multimodal Large Language Model (MLLM) and a Text-Image-to-Image (TI2I) model. It begins by initializing a {M}ultimodal {V}ideo {MEM}ory (\emph{\memory{}}) via synthesizing global entity and environment references  $\mathcal{R}$, then synthesizing video segments autoregressively, continuously retrieving and updating the \memory{} for context-aware synthesis and self-improvement. 

\subsection{\memory{} Design and Initialization} \label{subsec:agentic-memory}
\memory{} enables \model{} to explicitly track evolving video world states and events, thus enforcing long-range dependencies for temporal consistency and narrative coherence across segments. 
% We detail its schema below, and see \Cref{fig:method-overview} for its architectural mapping.

\paragraph{Memory Schema.} 
Unlike existing studies that store only visual references, \memory{} stores structured contexts from synthesized segments, denoted as $\mathcal{M} := \{\mathcal{M}_1, \ldots, \mathcal{M}_{N}\} \cup \mathcal{R} \cup \mathcal{D}$. Here, $\mathcal{R}$ is the set of global reference frames (including user-provided $\mathcal{R}^u$), and $\mathcal{D}$ is the prompt database. Each segment memory $\mathcal{M}_j := \{T_j, \mathcal{F}_j, V_j\}$ disentangles the video segment into three complementary modalities:

\begin{itemize}
    \item \emph{Textual States ($T_j$).} To capture the evolving narrative for consistency and coherence, we model the video's underlying state as a structured, fine-grained representation, inspired by \citep{johnson2015image}. $T_j$ consists of: \emph{(1) Visual Arcs} that track entity and environment features and their temporal evolution, recording elements' Identity, Identity Changes, and Motion; \emph{(2) Spatial Relations} that capture subject-relation-object triplets from the begin frame to ground geometric layouts; and \emph{(3) Camera} states that record viewport trajectories for visual continuity. We extract $T_j$ hierarchically: first deriving elements' Identity and Spatial Relations from the begin frame (\textcolor{red}{$T^F_j$}), then supplementing missing elements, Identity Changes, Motions, and Camera dynamics from the full segment to form \textcolor{red}{$T_j$}. This decouples frame-level from video-level extraction for \model{}'s pipeline.
    \item \emph{Frames ($\mathcal{F}_j$ or $\mathcal{R}$).} To anchor the concrete visual details that text cannot fully articulate,  \memory{} stores global reference frames $\mathcal{R}$ (both synthesized and user-provided, $\mathcal{R}^{u} \subseteq \mathcal{R}$), each indexed by a generated caption, and segment keyframes $\mathcal{F}_j := \{F^{\text{begin}}_ j, F^{\text{end}}_j\}$, indexed by $S_j$. Our framework can accommodate more advanced frame extraction and indexing methods.
    \item \emph{Videos ($V_j$).} To capture temporal motion dynamics for cross-segment smooth transitions and motion continuity, \memory{} saves the synthesized segments for segment verification and refinement. Like keyframes, $V_j$ is simply indexed by $S_j$.
\end{itemize}
\memory{} enables two core online operations: \emph{Retrieve} fetches relevant past states and \emph{Update} writes the newly synthesized $\mathcal{M}_j$ for subsequent generation, see below. $\mathcal{D}$ is described in \Cref{subsec:memory-augmented-algorithms}. 

\paragraph{\memory{} Initialization.} 
Before synthesizing segments, inspired by identity-reference approaches \citep{zheng2024videogen,liu2025rolling} for consistency, \model{} initializes \memory{} by establishing global reference backgrounds and entities, $\mathcal{R} := \mathcal{R}^{\text{bg}} \cup \mathcal{R}^{\text{ent}} \cup \mathcal{R}^{u}$ as a form of long-term memory:

\noindent\emph{\text{(}i\text{)} Planning.} The agent first reasons over $\mathcal{S}$ (and $\mathcal{R}^{\text{u}}$ if available) to identify the environments and entities, their required appearances (both explicitly specified and implicitly implied from $\mathcal{S}$), and generates prompts for synthesizing these references, using the MLLM:
\begin{equation}\label{longveo-eq:memory-initializations}
\small
\mathcal{P}_{\mathcal{R}}^{\text{bg}} := \text{MLLM}^{\text{bg}}_{\text{plan}}\text{(}P, \mathcal{S},
\mathcal{R}^u\text{)}, \quad \mathcal{P}_{\mathcal{R}}^{\text{ent}} := \text{MLLM}^{\text{ent}}_{\text
{plan}}\text{(}P, \mathcal{S}, \mathcal{R}^u\text{)}, \quad \mathcal{P}_{\mathcal{R}}^{u} := \text{MLLM}_{\text{cap}}\text{(}P, \mathcal{S}, \mathcal{R}^u\text{)}
\end{equation}
where $\mathcal{P}_{\mathcal{R}} := \mathcal{P}_{\mathcal{R}}^{\text{bg}} \cup \mathcal{P}_{\mathcal{R}}^{\text{ent}} \cup \mathcal{P}^{\text{u}}_{\mathcal{R}}$ represents identified entities and environments' prompts, and $\mathcal{P}^{\text{u}}_{\mathcal{R}}$ denotes the captions of user-provided reference frames $\mathcal{R}^{\text{u}}$.

\noindent\emph{\text{(}ii\text{)} Identifying Dependencies.} 
The agent constructs a dependency Directed Acyclic Graph $\mathcal{G}$ over $\mathcal{P}_{\mathcal{R}}$: $\mathcal{G} := \text{MLLM}_{\text{dep}}\text{(}\mathcal{P}_{\mathcal{R}}\text{)}$, to identify which references depend on others (e.g., an entity depends on its environment). $\mathcal{G}$ is then decomposed into weakly connected components. Within each component, topological sorting is applied to determine the synthesis order to respect the dependencies. 

\noindent\emph{\text{(}iii\text{)} Synthesizing References.} 
The agent synthesizes a reference frame for each prompt in $\mathcal{P}_{\mathcal{R}} \setminus \mathcal{P}^{\text{u}}_{\mathcal{R}}$ using the TI2I, conditioned on its dependent references following topological order. All components are synthesized in parallel, yielding $\mathcal{R}$. See \Cref{appdx:init-prompts} for our prompts.

\subsection{Agentic Auto-Regressive Generation Pipeline} 
\label{subsec:autoregressive-generation}

After establishing global references, \model{} synthesizes long videos autoregressively, segment-by-segment. For each segment context $S_i$, the agent first determines the generation mode, then operates in a \textbf{Retrieve--Synthesize--Refine--Update} closed-loop, where Synthesize--Refine is applied first to boundary frames and then to the video segment. For convenience, we duplicate $S_{N\text{+}1} := S_N$ in $\mathcal{S} := \{S_1,\dots,S_{N\text{+}1}\}$, and synthesize segments for $i\leq N$. See \Cref{appdx:method-prompts} for our prompts.

\noindent\emph{(i) Adaptive Segment Generation.} 
A key challenge for segment-based generation is balancing narrative progression with consistency. Prior works adopt either {extrapolation}
% ---extrapolating each segment from a begin frame 
\citep{an2025onestory,zhou2026videomemory} or {interpolation}
% ---interpolating it from begin and end frames 
\citep{yin2023nuwa}. Extrapolation allows natural video world progression but risks semantic drift, particularly for entities and environments not visible in the begin frame. Interpolation enforces stronger consistency, but {risks unnatural progression especially when TI2I models lack the temporal reasoning to reliably synthesize how environments evolve over a predefined duration, given static references} (\Cref{fig:baseline-examples-progression}). \model{} instead adaptively selects the mode per segment:
\begin{equation}
\label{longveo-eq:generation-mode}
\small
\text{mode}_i := \begin{cases} \text{extrapolation} & \text{if }\mathcal{C}\text{(}S_i\text{)} \wedge S_{i\text{+}1} \text{ is spatio-temporally continuous with } S_i \\ \text{interpolation} & \text{otherwise} \end{cases}
\end{equation}
\noindent where  $\mathcal{C}\text{(}S_i\text{)}$ indicates that $S_i$ does not transition to a new established environment, and both conditions are inferred by $\text{MLLM}_{\text{mode}}\text{(}\mathcal{S}\text{)}$. Interpolation applies when $S_i$ is a multi-shot context whose shots span different environments, or when $S_{i\text{+}1}$ jumps to a new environment. The second condition is omitted for $i = N$. See \Cref{fig:adaptive-segment-generation-examples} for an example of our mode selection.

\noindent\emph{(ii) Retrieve.} 
After determining the mode, \model{} retrieves text, image, and video contexts for synthesis. For any $j$-th segment context, to mitigate false-positive conditioning, the agent employs an MLLM to identify the top-$k$ most narratively relevant previous segments. It acquires the textual states $\mathcal{T}^{\text{rel}}_j$, visual references $\mathcal{F}^{\text{rel}}_j$, and the immediate temporally contiguous segment $\mathcal{V}^{\text{rel}}_j$ (if any):
\begin{equation}\label{longveo-eq:retrieve}
\small
\mathcal{S}^{\text{rel}}_j := \text{MLLM}^{\text{scenes}}_{\text{retr}}\text{(}\mathcal{S}, k, S_j\text{)}, \quad \mathcal{V}^{\text{rel}}_j := \text{MLLM}^{\text{cont}}_{\text{retr}}\text{(}\{V_{< j}\}, S_j\text{)}, \quad \mathcal{R}^{\text{rel}}_j := \text{MLLM}^{\text{ref}}_{\text{retr}}\text{(}\mathcal{R}, \mathcal{P}_{\mathcal{R}}, S_j\text{)}
\end{equation}
\begin{equation} \label{longveo-eq:retrieve-global-refs}
\small
\mathcal{T}^{\text{rel}}_j := \{T_n \mid S_n \in \mathcal{S}^{\text{rel}}_j\}, \quad \mathcal{F}^{\text{rel}}_j := \{F^{\text{begin}}_n, F^{\text{end}}_n \mid S_n \in \mathcal{S}^{\text{rel}}_j\} \cup \mathcal{R}^{\text{rel}}_j
\end{equation}
When available, $\mathcal{F}^{\text{rel}}_j$ and $\mathcal{T}^{\text{rel}}_j$ are extended with $F^{\text{begin}}_i$ and $T^F_i$ respectively, ensuring subsequent synthesis for the current segment is conditioned on the begin frame's context.

\noindent\emph{(iii) Synthesize and Self-Improve--Boundary Frame(s).} 
Based on \Cref{longveo-eq:generation-mode}, \model{} synthesizes boundary frames $\mathcal{F}_i := \{F^{\text{begin}}_i, F^{\text{end}}_i\}$. It assigns $F^{\text{begin}}_{i} := F^{\text{end}}_{i-1}$ for $i > 1$ for continuity, and synthesizes $F^{\text{begin}}_1$ via generating its frame prompt first, and then the frame $F^{\text{begin}}_1 := \text{TI2I}\text{(}\text{MLLM}^{\text{img}}_{\text{pgen}}\text{(}S_1\text{)}, \mathcal{F}^{\text{rel}}_1\text{)}$. For $i > 1$, denote $\mathcal{V}^{\text{rel}}_i := \{V_t\}$, the end frame $F^{\text{end}}_i$ is determined using the lookahead context of the subsequent segment $i\text{+}1$:
\begin{equation}
\label{longveo-eq:interpolate-extrapolate}
\small
F^{\text{end}}_i := \begin{cases} \text{TI2I}\text{(}P_i^{\text{end}}, \mathcal{F}^{\text{rel}}_{i\text{+}1}\text{)} & \text{mode}_i = \text{interp},\ V_t \text{ does not exist} \lor t = i{-}1 \\ \text{MLLM}^{\text{img}}_{\text{retr}}\text{(}\mathcal{K}_t, S_t, S_{i\text{+}1}\text{)} & \text{mode}_i = \text{interp},\ V_t \text{ exists},\ t \neq i{-}1 \\ \emptyset & \text{mode}_i = \text{extrap}
\end{cases}
\end{equation}
where $P_i^{\text{end}} := \text{MLLM}^{\text{img}}_{\text{pgen}}\text{(}S_{i\text{+}1}, \mathcal{T}^{\text{rel}}_{i\text{+}1}\text{)}$ is the generated frame prompt. The case $t \neq i{-}1$ is particularly challenging. It arises when segment $i$ resumes some events from the middle of a non-adjacent segment $t$. To obtain the end frame of the relevant shot in segment $t$ for resumption, we extract all shot end frames $\mathcal{K}_t$ from $V_t$ \citep{Castellano_PySceneDetect}, then the $\text{MLLM}$ selects the one that best continues into $S_{i\text{+}1}$ (\Cref{appdx-how-aard-maintains-long-distance-continuity} for an example). All synthesized frames by the TI2I model then undergo a frame-level self-improvement process, described in \Cref{subsec:memory-augmented-algorithms}. 

\noindent\emph{(iv) Synthesize and Self-Improve--Video Segment.} After obtaining $\mathcal{F}_i$, \model{} synthesizes the segment:
\begin{equation} \label{longveo-eq:video-syn}
\small
V_i := \text{TI2V}\text{(}P_i, \mathcal{F}_i, \mathcal{F}^{\text{rel}}_i\text{)}
\end{equation}
where $P_i := \text{MLLM}^{\text{vid}}_{\text{pgen}}\text{(}S_i, \mathcal{T}^{\text{rel}}_i, \mathcal{F}_i\text{)}$ is the video segment prompt. Once synthesized, $V_i$ undergoes a video-level self-improvement process, described in \Cref{subsec:memory-augmented-algorithms}. 

\noindent\emph{(v) Update.} After self-improvements, \memory{} saves $\text{(}\mathcal{F}_i, V_i, T_i, T^F_{i\text{+}1}\text{)}$ if $i < N$ for reference in subsequent generation, where $T_i$ and $T^F_{i\text{+}1}$ are textual states extracted during refinement processes.

\subsection{Hierarchical Boundary Frame and Video Self-Improvement} \label{subsec:memory-augmented-algorithms}

To mitigate the risk of cascading temporal errors, where a single inconsistent frame can propagate artifacts across the entire horizon, \model{} introduces \textbf{HIerarchical Test-time Self-improvement (HITS)} to self-improve synthesized frames and video segments hierarchically. Unlike existing works \citep{liu2025video,long2025vista} that apply search and closed-loop refinements to short clips, \model{} extends this paradigm to self-improve both intra- and inter-segment coherence. 

\paragraph{Boundary Frame Self-Improvement.} This step self-improves $F^{*}_i$ ($* \in \{\text{begin}, \text{end}\}$) interactively. At each iteration, \model{} extracts frame textual states from the synthesized frame: $T^{F}_i$ (\Cref{subsec:agentic-memory}) for the begin frame $F^\text{begin}_i$, and $T^{F}_{i\text{+}1}$ for the optional end frame $F^\text{end}_i$, via $T^{F}_{i\text{+}\delta} := \text{MLLM}^{\text{img}}_{\text{ext}}\text{(}S_{i\text{+}\delta}, F^{*}_i\text{)}$ with $\delta \in \{0,1\}$. ($F^{*}_i$, $T^{F}_{i\text{+}\delta}$) is then verified via a 8-metric rubric focusing on consistency and basic image quality on a scale of 1--10 in 3 groups: (i) \emph{Entity Consistency}, \emph{Environment Consistency}, \emph{Narrative Progression}, and \emph{Spatial Logicalness}; (ii) \emph{Entity State}, \emph{Environment State}; (iii) \emph{Instruction Following} and \emph{Physical Plausibility}: 
\begin{equation}\label{longveo-eq:frame-verifications}
\small
\begin{aligned}
\mathcal{Q}_i^F(F^*_i) := \underbrace{\text{MLLM}^{\text{img-imgs}}_{\text{vfy}}\text{(}P^{*}_i, F^{*}_i, \mathcal{F}^{\text{rel}}_{i\text{+}\delta}\text{)}}_{\text{(i) Inter Consistency over Images}} \cup \underbrace{\text{MLLM}^{\text{img-st}}_{\text{vfy}}\text{(}P^{*}_i, T^{F}_{i\text{+}\delta}, \mathcal{T}^{\text{rel}}_{i\text{+}\delta}\text{)}}_{\text{(ii) Inter Consistency over States}} \cup \underbrace{\text{MLLM}^{\text{img-qa}}_{\text{vfy}}\text{(}S_{i\text{+}\delta}, P^{*}_i, F^{*}_i, T^{F}_{i\text{+}\delta}\text{)}}_{\text{(iii) Basic Quality}}
\end{aligned}
\end{equation}
The agent then refines $F^{*}_i$ via \emph{Edit} or \emph{Regenerate}: the $\mathcal{Q}_i^F$ is input into the MLLM to decide the mode and, if \emph{Edit} is chosen, to suggest the edit prompt. The Edit mode targets a single issue only, as {it is challenging to fix multiple errors simultaneously}. If {Regenerate} is chosen, $P^{*}_i$ is optimized through our \emph{Memory-Augmented Prompt Optimization (MAPO)} algorithm, see below, and $F^{*}_i$ is re-synthesized for next iteration. The final $F^{*}_i$ is:
\begin{equation}\label{longveo-eq:refine-eq}
\small
F^{*}_i := \arg\max_{F \in \mathcal{F}_{\text{cand}}} \text{Agg}\text{(}\mathcal{Q}^F_i\text{(}F\text{)}\text{)}
\end{equation}
where $\mathcal{F}_{\text{cand}}$ is the set of candidate frames generated across all refinement iterations.

\paragraph{Video Segment Self-Improvement.} 
This step self-improves $V_i$ interactively. Similar to above, \model{} first extracts full video states: $T_i := \text{MLLM}^{\text{vid}}_{\text{ext}}\text{(}S_i, T^{F}_i, V_i\text{)}$ (\Cref{subsec:agentic-memory}). It then verifies ($V_i$, $T_i$) via a 10-metric rubric focusing on inter-segment consistency, intra-segment consistency, and basic video quality, divided into three groups, each scored on a scale of 1--10: (i) \emph{Inter Entity Consistency}, \emph{Inter Environment Consistency}, \emph{Inter Motion Consistency}, \emph{Camera Consistency}; (ii) \emph{Intra Entity Consistency}, \emph{Intra Environment Consistency}; and (iii) \emph{Instruction Following}, \emph{Physical Plausibility}, \emph{Narrative Progression}, and \emph{Frame Fit} (only when $F^{\text{end}}_i$ is available):
\begin{equation}\label{longveo-eq:vid-verifications}
\small
\begin{aligned}
\mathcal{Q}_i^V(V_i) := \underbrace{\text{MLLM}^{\text{vid-vid}}_{\text{vfy}}\text{(}\mathcal{S}^{\text{rel}}_i, S_i, S_t, T_i, T_t, V_i, V_t\text{)}}_{\text{(i) Inter Consistency}} \cup \underbrace{\text{MLLM}^{\text{vid-qa}}_{\text{vfy}}\text{(}S_i, P_i, V_i, T_i, F^{\text{end}}_i\text{)}}_{\text{(ii), (iii): Intra Consistency and Basic Quality}}
\end{aligned}
\end{equation}
The agent refines $V_i$ depending on the availability of $F^{\text{end}}_i$. If $F^{\text{end}}_i$ is available, the agent optimizes the text prompt $P_i$ only. If $F^{\text{end}}_i$ is unavailable, prompt-only optimization is insufficient, as entities and environments absent from $F^{\text{begin}}_i$ or transformed during the segment can drift from references. In this case, \model{} co-optimizes both $F^{\text{end}}_i$ and $P_i$ sequentially: it first extracts $F^{\text{end}}_i$ from $V_i$, self-improves it following the frame self-improvement process above with \emph{Edit} mode only (to preserve any natural layout progression), re-optimizes $P_i$ via MAPO conditioned on the updated boundary frames $F_i := \{F^{\text{begin}}_i, F^{\text{end}}_i\}$, and then re-synthesizes $V_i$ for the next iteration. The final $V_i$ is:
\begin{equation}
\small
V_i := \arg\max_{V \in \mathcal{V}_{\text{cand}}} \text{Agg}\text{(}\mathcal{Q}^V_i\text{(}V\text{)}\text{)}
\end{equation}
where $\mathcal{V}_{\text{cand}}$ is the set of candidate videos generated across refinement iterations.

\paragraph{Memory-Augmented Prompt Optimization (MAPO).}
To improve the refinement efficacy, we introduce \textbf{MAPO}, which leverages the history of successful and failed cases indexed by rubric scores. Specifically, \memory{} maintains a prompt database $\mathcal{D} := \{\text{(}P, P^{*}, \mathcal{Q}, \ell\text{)}\}$ where each entry stores an original prompt $P$, its refined version $P^{*}$, rubric scores $\mathcal{Q}$, and a {hard label} $\ell \in \{\text{pos}, \text{neg}\}$ indicating positive and negative refinements. Each entry is indexed by a semantic embedding $\text{Emb}\text{(}P, \mathcal{Q}\text{)}$ for efficient retrieval. $\mathcal{D}$ is seeded with a few prior cases and updated online: a case is assigned as `$\text{pos}$' if all rubric scores improve, and `$\text{neg}$' if all scores worsen. Given $\mathcal{Q}^F_i$/$\mathcal{Q}^V_i$ and the $P^{*}_i$/$P_i$, MAPO retrieves the top-$k$ relevant positive and negative cases via {cosine similarity over ($\mathcal{Q}$, ${P}$) embeddings} from $\mathcal{D}$. Inspire by \citet{zhao2024expel}, MAPO then contrasts the positive and negative cases to derive refinement guidelines, reasons over $\mathcal{Q}$ to identify root causes of failures, and applies targeted edits with guidelines to produce $P^{*}$. {After each refinement, the new case is labeled and added to $\mathcal{D}$ online.}

\subsection{\model{}-Parallel (\model{}-Par)} \label{subsec:longveo-par}

% \kmy{How is ``refinement'' different from ``self-improvement''?  Please use consistent terminology.}
We introduce \model{}-Par, a parallel version of \model{}, for efficiency. \model{}-Par synthesizes both boundary frames $\mathcal{F}_i := \text{\{}F^{\text{begin}}_i, F^{\text{end}}_i\text{\}}, \forall i$ autoregressively and performs the same frame self-improvement process. All video segments are then synthesized in parallel, with no video self-improvement applied. 
% \kmy{Start with the original runtime and explain all variables; then in a separate sentence, describe the savings.  Don't mix and match, it is confusing.} 
In \model{}, video synthesis latency takes $N k_v L_V$, where $N$ is the \#segments, $k_v$ is the \#self-improvement iterations, and $L_V$ is the latency of a single video synthesis call. \model{}-Par removes the sequential dependency, reducing this to $L_V$ under ideal hardware (\Cref{subsec:latency-analysis}). While this sacrifices some spatial consistency for evolving environments, \model{}-Par still enforces strict character consistency and produces coherent stories for scenes with static environments (\Cref{fig:baseline-examples-progression,fig:baseline-examples-consistency} for examples). 
% \kmy{Make clear where you can find out the performance tradeoffs in your report, cross reference.  Why keep character consistency?}
% \yale{Need to be precise (yet concise!); can we add some complexity analysis/numbers here?}. 
% \yale{Run-on sentence; can we split this into smaller sentences.}

% %%%%%%%%%%%%%%%%%%%%%%%%%%%%%%%%%
\section{Long Video Bench-Challenge (\dataset{})}

%%%%%%%%%%%%%%%%%%%%%%%%%%%%%%%%%
\begin{table*}[t!]
\centering
\resizebox{\textwidth}{!}{%
\begin{tabular}{l|ccc|ccc|cccc}
\toprule
& \multicolumn{3}{c|}{\textbf{Basic Statistics}}
& \multicolumn{3}{c|}{\textbf{Single-Scene}} 
& \multicolumn{4}{c}{\textbf{Multi-Scene}} \\

\textbf{Benchmark} 
& \textbf{\#Samples / }
& \textbf{Avg. Prompt}
& \textbf{Avg.}
& \textbf{Identity} 
& \textbf{Temporal} 
& \textbf{Spatial} 
& \textbf{Character State} 
& \textbf{Object State} 
& \textbf{Environment State}
& \textbf{Cyclical} \\

& \textbf{\#Prompts}
& \textbf{Length}
& \textbf{Dur.}
& \textbf{Consistency} 
& \textbf{Consistency}
& \textbf{Consistency}
& \textbf{Evolving}
& \textbf{Evolving}
& \textbf{Evolving}
& \textbf{Appearance} \\
\midrule

VBench \citep{huang2025vbench} & 946 / 946 & 7.64 & N/A & \cmark & \cmark & \cmark & \xmark & \xmark & \xmark & \xmark \\
VBench-Long \citep{huang2025vbench} & 40 / 40 & 48.25 & N/A & \cmark & \cmark & \cmark & \xmark & \xmark & \xmark & \xmark \\
T2V-CompBench \citep{sun2024t2v} & 1400 / 1400 & 10.42 & N/A & \cmark & \cmark & \cmark & \xmark & \xmark & \xmark & \xmark \\            
VBench-2.0 \citep{zheng2025vbench} & 1013 / 1013 & 21.83 & N/A & \cmark & \cmark & \cmark & \xmark & \xmark & \xmark & \xmark \\
MovieGenBench \citep{polyak2025moviegencastmedia} & 1003 / 1003 & 18.21 & N/A & \cmark & \cmark & \cmark & \xmark & \xmark & \xmark & \xmark \\
\midrule
MovieBench (test) \citep{wu2025moviebench} & 6 / 2875 & 91.5 & 32.0m & \cmark & \cmark & \cmark & N/A & N/A & N/A & N/A \\
MoviePrompts \citep{wu2025automated} & 10 / 10 & 95.70 & N/A & \cmark & \cmark & \cmark & \cmark & \xmark & \xmark & \xmark \\
ST-Bench \citep{zhang2025storymem} & 30 / 95 & 40.69 & 1m & \cmark & \cmark & \cmark & \cmark & \cmark & \cmark & \xmark \\
\midrule
\rowcolor[HTML]{E8E8E8} \textbf{\dataset{} (Ours)} & 125 / 5200 & 18.25 & 3m, 5m, 10m & \cmark & \cmark & \cmark & \cmark & \cmark & \cmark & \cmark \\
\bottomrule
\end{tabular}
}
\caption{Comparison of text-to-video generation benchmarks.
Existing benchmarks focus on single-scene consistency but lack evaluation of the challenging cyclical state tracking where entities and environments appear, disappear for extended periods, then reappear with non-trivial evolved states. } 
\label{tab:dataset-statistics}
\end{table*}
%%%%%%%%%%%%%%%%%%%%%%%%%%%%%%%%%

Existing single- or multi-scene benchmarks neglect real-world scenarios where entities and environments undergo non-linear transitions---appearing, disappearing, and reappearing with optional state changes across scenes (\Cref{tab:dataset-statistics}). We introduce \textbf{\dataset{}}, a benchmark that stress-tests models' ability to maintain consistent and coherent world states in such scenarios. 
% Our core insight is that we force entities and environments to change through cyclical ``appearance-disappearance-reappearance'' patterns.
\dataset{} features three challenge types: (i) \emph{Evolving Character States}, where characters reappear with evolved states (e.g., clothing, appearance, physical condition); (ii) \emph{Evolving Object States}, where objects reappear with changed states (e.g., position, orientation, condition); and (iii) \emph{Evolving Environment States} (e.g., evolution, progressive revelation of details). We instantiate \dataset{} with 120 text-only scenarios: 50 samples each for 3- and 5-minute videos (30 character, 10 object, 10 environment), and 25 samples for 10-minute videos (10 character, 5 object, 5 environment). Each scenario consists of concise scene descriptions that either continue from previous scenes or transition to new ones, forming coherent narratives, see Appx.-\Cref{fig:example_3m_scenario,fig:example_5m_scenario,fig:example_10m_scenario} for examples. 

\paragraph{Human-In-The-Loop Dataset Construction.} 
We instantiate \dataset{} with 3-, 5-, and 10-minute scenarios using a common time-independent construction pipeline. For each challenge type, we first carefully craft a professional screenwriter persona prompt, incorporating one human-designed demonstration and generate scenarios using a state-of-the-art MLLM \citep{googledeepmind2025gemini3flash}. 
% \kmy{Here and elsewhere: Please edit this into a coherent English sentence, not a choppy noun phrase.  If important, add numberings (e.g., (3) State Change.} 
This generation is enforced with constraints: \emph{(i) Content}, where story flow must be meaningful and natural, each segment fits a pre-defined duration, clear cause-and-effect relationships, no random events; \emph{(ii) Gap Rules}, where main entities and environments must be absent for at least
% \kmy{Added variable to imply the possibility of tuning.}  
$n=10$ segments before reappearing with or without state changes; and \emph{(iii) State Change}, ensuring natural state changes through realistic activities, specific visual markers (positions, appearances, conditions). During generation, we manually review generated scenarios and update constraints or demonstrations to improve diversity.

\paragraph{Data Refinement and Deduplication.} 
% \kmy{The refinement steps are actually better for motivating the dataset but it doesn't read forcefully enough to really suggest that there is a clear need for this.  You should make it clear that the dataset consists of the prompts and descriptions first (as opposed to the videos) so that you can make it an objective evaluation dataset that others can use and benchmark against.  Make it clear how others can use this dataset (either in this section or in the discussion section later.} \\ 
% \kmy{Worries about self-preference here as you use the same model, can you use at least a different critique agent model?} 
Since raw MLLM-generated scenarios often contain duplicates, logical gaps, and low-quality content, we apply systematic refinement. The generated scenarios are deduplicated using the same MLLM: we summarize these scenarios one by one, and feed summaries into the MLLM to {identify and remove similar scenarios}. We then customize Self-Refine \citep{madaan2023self} for self-refinement: selected scenarios undergo rigorous MLLM-Judge validation against six criteria: \emph{(i) Specificity Verification} ensuring each scene must be specific enough with clear revelations; \emph{(ii) Logic Verification} where each scene reveals new details not mentioned before that must have logically existed from beginning; \emph{(iii) Natural Verification} where entity actions must be natural without forced or contrived scenarios; \emph{(iv) Realism Verification} ensuring details must be realistic and appropriate with everyday believable activities; \emph{(v) Repetition Verification} ensuring activities and details must be varied without repetitive content; and \emph{(vi) Contradiction Verification} with no contradictions across segments regarding entity and environment states. Failed scenarios in any criterion undergo refinement until successful or up to a limited number of iterations. To avoid self-preference bias, a separate MLLM \citep{anthropic2025sonnet45} re-verifies all scenarios using the same criteria, and refines any minor issues. We manually review a subset of samples to confirm quality.

% \paragraph{Descriptive Statistics.} 
% % \kmy{This should have come at the beginning.}
% \dataset{} contains 120 text-only scenarios: 50 samples each for 3- and 5-minute videos (30 character, 10 object, 10 environment), and 20 samples for 10-minute videos (10 character, 5 object, 5 environment). Each scenario consists of concise scene descriptions that either continue from previous scenes or transition to new ones, forming coherent narratives (see \Cref{tab:dataset-statistics}). 

%%%%%%%%%%%%%%%%%%%%%%%%%%%%%%%%%
\section{Experiments}\label{sec:experiments}

\subsection{Settings}\label{subsec:experiment-setups}

\paragraph{Benchmarks.} 
We conduct experiments on both {single-scene} and {multi-scene} long video generation. For single-scene, following \citet{yang2025longlive,yi2025deep}, we use \emph{VBench-Long} \citep{huang2025vbench} with 40 prompts, each decomposed into 8 continuous segments using Gemini \citep{googledeepmind2025gemini3flash}, yielding approximately 1-min videos. For multi-scene, we use our \emph{\dataset{}} benchmark featuring challenging environment transitions and entity evolutions across 3-min videos (24 scenes) and 5-min videos (40 scenes), with 20 samples each, and 10-min videos in \Cref{subsec:experiments-10m-dataset}.

\paragraph{Models and Baselines.} 
We instantiate \model{} and baselines with Gemini 3 Flash \citep{googledeepmind2025gemini3flash} as the MLLM, Nano Banana 2 \citep{ raisinghani2026nanobanana2} as the TI2I model, and Veo 3.1 \citep{deepmind2025veo3} as the TI2V model. We compare \model{} against {SOTA autoregressive and parallel segment-based baselines}: \emph{(i) Direct Prompting}, which generates each video segment directly from its scene description without any conditioning; \emph{(ii) Naive Autoregressive (Naive-AR)}, which generates each video segment via extrapolation, conditioned only on the last frame of the previous segment; {\emph{(iii) Naive-Par}}, which autoregressively synthesizes the end frame of each segment ($F^{\text{end}}_i := \text{TI2I}(S_i | F^{\text{begin}}_1, F^{\text{end}}_1, \dots, F^{\text{end}}_{i-1})$) and subsequently uses the begin and end frames to interpolate the video segments in parallel; \emph{(iv) MovieAgent} \citep{wu2025automated}, a hierarchical multi-agent framework that automates long-form movie generation; \emph{(v) ViMax} \citep{vimax2025}, a multi-agent framework that automates end-to-end long-form video generation through script planning, storyboarding, character design, and reference-guided shot synthesis; and \emph{(vi) VideoMemory} \citep{zhou2026videomemory}, {which synthesizes long video autoregressively} via maintaining a dynnamic memory bank of entity references for visual consistency.

\paragraph{Implementation Details.} 
We run \model{} with two iterations for frame and two for video refinements. At each iteration, we synthesize three frames and three videos via batch inference, run all the judges in \Cref{longveo-eq:frame-verifications,longveo-eq:vid-verifications} in parallel and the best one will be selected for next refinement iteration. We implement early stopping when average frame/video scores $\geq$ 9/10. In total, a video segment requires at most 6 videos and 6 images. For efficiency, we pre-compute \Cref{longveo-eq:generation-mode,longveo-eq:retrieve} and \Cref{longveo-eq:retrieve-global-refs} ($\text{MLLM}_\text{rel-imgs}$) over all segments at once, see \emph{All Scenes} prompts in \Cref{appdx:method-prompts}. For scaling to longer horizons, we limit each \memory{} schema to available hardwares, and cap the maximum window size at 100 for operations involving $\mathcal{S}$ and image lists as MLLM inputs (e.g., \Cref{longveo-eq:memory-initializations,longveo-eq:generation-mode,longveo-eq:retrieve}). 

% More advanced windowing strategies are left for future work.

% For each $S_i$, we identify relevant previous segments via semantic similarity with $S_{< i}$, retrieving the top-$r$ candidates with $r \gg k$. We then feed all these segments to the MLLM to select the top-$k$ most relevant ones. Similarly, relevant reference frames for each segment are selected by feeding all $\mathcal{R}$'s captions to the MLLM. Additionally, the contiguous previous segment is also pre-identified using the MLLM: given any segment, it identifies the contiguous previous segment (which is not necessarily segment $i-1$), if one exists. We use MLLM for these retrieval steps because the MLLM and TI2V models can be highly sensitive to the selected information, thus requiring only the most relevant segments. We set $k=5$ and $r=20$. See Appendix for details.

%%%%%%%%%%%%%%%%%%%%%%%%%%%
\begin{table*}[t!]
\centering
\resizebox{\linewidth}{!}{
\begin{tabular}{l|c|ccccccc}
\toprule
\multirow{2}{*}{\textbf{Method}} & \multirow{2}{*}{\textbf{Type}} & \multirow{2}{*}{\begin{tabular}[c]{@{}c@{}}\textbf{Semantic}\\\textbf{Alignment}\end{tabular}} & \multirow{2}{*}{\begin{tabular}[c]{@{}c@{}}\textbf{Narrative}\\\textbf{Coherence}\end{tabular}} & \multicolumn{3}{c}{\textbf{Inter-shot Consistency}} & \multicolumn{2}{c}{\textbf{Intra-shot Consistency}} \\ \cmidrule(lr){5-7} \cmidrule(lr){8-9}
 & & & & \textbf{Character} & \textbf{Environment} & \textbf{Motions} & \textbf{Subject} & \textbf{Environment} \\ \midrule
Direct Prompting  & Par & 0.2012 & 0.3875 & 0.2695 & 0.3115 & 0.9710 & 0.7713 & 0.8929 \\
Naive-AR           & AR & 0.1828 & 0.7466 & 0.5125 & 0.7148 & {0.9903} & 0.9062 & 0.9363  \\
{Naive-Par}           & Par & 0.1814 & 0.6796 & 0.5500 & 0.6779 & 0.9836 & 0.8852 & 0.9148 \\
MovieAgent \citep{wu2025automated}            & Par & 0.1783 & 0.6071 & 0.5002 & 0.4944 & 0.9723 & 0.9150 & 0.9386 \\
ViMax \citep{vimax2025}         & Par & 0.1961 & 0.6912 & 0.5552 & 0.7015 & 0.9754 & \underline{0.9215} & 0.9488 \\
VideoMemory \citep{zhou2026videomemory}         & AR & 0.1926 & 0.6717 & 0.5729 & 0.7273 & 0.9777 & 0.9207 & 0.9492 \\
% \midrule
% Scaled Naive-AR           & AR & 0.2051 & 0.6875 & 0.6204 & 0.7251 & \underline{0.9906} & 0.8881 & 0.9131 \\
% Scaled VideoMemory        & AR & - & 0.7311 & 0.6595 & 0.7655 & - & - & - \\ 
\midrule
\rowcolor[HTML]{E8E8E8} \textbf{\model{}-Par (Ours)} & Par & \underline{0.2094} & \underline{0.8082} & \underline{0.6817} & \underline{0.8180} & 0.9843 & 0.9108 & \textbf{0.9578} \\ 
\rowcolor[HTML]{E8E8E8} \textbf{\model{} (Ours)} & AR & \textbf{0.2111} & \textbf{0.8987} & \textbf{0.7353} & \textbf{0.8368} & \textbf{0.9935} &  \textbf{0.9231} & \underline{0.9514} \\ 
\bottomrule
\end{tabular}
}
\caption{Experiments on VBench-Long \citep{huang2025vbench} expanded to 8 continuous scenes for 1-min videos.}
\label{tab:vbench-long-results}
\vspace{-3mm}
\end{table*}
%%%%%%%%%%%%%%%%%%%%%%%%%%%

\subsection{Automatic Evaluations} \label{subsec:automatic-evals}

\paragraph{Automatic Metrics.} 
We evaluate generated videos across nine metrics: \emph{(i) Semantic Alignment} measuring ViCLIP-based \citep{wang2023internvid} text-video similarity between each video segment and its description; \emph{(ii) Narrative Coherence} following \citet{wang2025multishotmaster}, where we employ Gemini 3 Pro \citep{googledeepmind2026gemini31pro} to score story, entity and environment progression, and causal logic on a scale of 0--1 with Self-Consistency \citep{wang2022self}, with penalties for repetitive or incoherent content. For inter consistency, we measure \emph{(iii) Character Consistency}, \emph{(iv) Environment Consistency} following \citet{an2025onestory,meng2025holocine}, and \emph{(v) Motion Consistency} customized from VBench~\citep{Huang_2024_CVPR} for contiguous segments. For intra consistency, we report \emph{(vi) Subject Consistency}, \emph{(vii) Environment Consistency} from VBench. We assess general video quality in \Cref{appdx:general-video-quality}, and provide implementation details in \Cref{appdx:implemntation-details} and qualitative analysis in \Cref{appdx:method-design-analysis}.

\paragraph{Single-Scene Results.} 
\Cref{tab:vbench-long-results} presents results on 1-min generation on VBench-Long. \model{} achieves substantial improvements across all metrics in both narrative quality and consistency. For narrative coherence, existing segment-based methods perform poorly (0.69 for ViMax, 0.67 for VideoMemory), as they force shot changes at every segment and produce repetitive or inconsistent environments (\Cref{fig:baseline-examples-progression}, \Cref{sec:storyboard-examples}). Meanwhile, \model{} reaches 0.9, outperforming the best baseline (Naive-AR at 0.75) by 20\%. For consistency, \model{} attains significant gains in both characters (0.74 vs. 0.57/0.56, a 30\% improvement) and environments (0.84 vs. VideoMemory's 0.73). {Remarkably, it obtains a motion consistency of 0.9935, substantially surpassing baselines and indicating that transitions between consecutive segments are nearly as smooth as a single diffusion generation.} \model{}-Par also performs competitively with 0.81 narratives while enabling parallel generation, offering a practical efficiency-consistency trade-off. Both \model{} versions achieve the highest semantic alignment, suggesting that they better satisfy user prompts than baselines.

%%%%%%%%%%%%%%%%%%%%%%%%%%%
\begin{table*}[t!]
\centering
\label{tab:results_modified}
\resizebox{\linewidth}{!}{
\begin{tabular}{l|c|ccccc|ccccc}
\toprule
\multirow{2}{*}{\textbf{Method}} 
& \multirow{2}{*}{\textbf{Type}} & \multicolumn{5}{c|}{\textbf{3-min (24 scenes)}} 
& \multicolumn{5}{c}{\textbf{5-min (40 scenes)}} \\ 
\cmidrule(lr){3-7} \cmidrule(lr){8-12}
& & \begin{tabular}[c]{@{}c@{}}\textbf{Semantic}\\\textbf{Alignment}\end{tabular}
& \begin{tabular}[c]{@{}c@{}}\textbf{Narrative}\\\textbf{Coherence}\end{tabular}
& \begin{tabular}[c]{@{}c@{}}\textbf{Inter-shot}\\\textbf{Character}\end{tabular}
& \begin{tabular}[c]{@{}c@{}}\textbf{Inter-shot}\\\textbf{Background}\end{tabular}
& \begin{tabular}[c]{@{}c@{}}\textbf{Inter-shot}\\\textbf{Motions}\end{tabular}
& \begin{tabular}[c]{@{}c@{}}\textbf{Semantic}\\\textbf{Alignment}\end{tabular}
& \begin{tabular}[c]{@{}c@{}}\textbf{Narrative}\\ \textbf{Coherence}\end{tabular}
& \begin{tabular}[c]{@{}c@{}}\textbf{Inter-shot}\\\textbf{Character}\end{tabular}
& \begin{tabular}[c]{@{}c@{}}\textbf{Inter-shot}\\\textbf{Environment}\end{tabular}
& \begin{tabular}[c]{@{}c@{}}\textbf{Inter-shot}\\\textbf{Motions}\end{tabular} \\ 
\midrule
Direct Prompting & Par & 0.2116 & 0.3525 & 0.3123 & 0.2819 & 0.9856 & 0.1700 & 0.3026 & 0.3102 & 0.3083 & 0.9760 \\
Naive-AR                       & AR & 0.2044 & 0.6325 & 0.3519 & 0.3224 & 0.9791 & 0.1409 & 0.6088 & 0.3310 & 0.3950 & 0.9823 \\
Naive-Par     & Par & 0.1941 & 0.8175 & 0.3637 & 0.3298 & 0.9813 & 0.1501 & 0.7132 & 0.3785 & 0.3869 & 0.9760\\
MovieAgent  & Par & 0.1886 & 0.5733 & 0.2922 & 0.2713 & 0.9513 & 0.1643 & 0.6112 & 0.3010 & 0.2569 & 0.9771 \\
ViMax    & Par & 0.2108 & 0.8600 & 0.3557 & 0.3231 & 0.9756 & 0.1559 & 0.7750 & 0.3570 & 0.3838 & 0.9799 \\
VideoMemory    & AR & 0.2062 & 0.8475 & 0.3687 & 0.3554 & 0.9759 & 0.1545 & 0.8292 & 0.3697 & 0.3830 & 0.9778 \\
% \midrule
% Scaled Naive-AR                       & AR & 0.1895 & 0.6250 & 0.3447 & 0.3295 & 0.9806 & 0.1796 & 0.6510 & 0.3257 & 0.3181 & 0.9790 \\
% Scaled VideoMemory        & AR & - & - & 0.3007 & 0.3573 & - & - & 0.3615 & 0.3615 & - & - \\
\midrule
\rowcolor[HTML]{E8E8E8} \textbf{\model{}-Par (Ours)}    & Par & \underline{0.2153} & \underline{0.8641} & \underline{0.4082} & \textbf{0.4270} & \underline{0.9918} & \textbf{0.1976} & \underline{0.8813} & \underline{0.3930} & \textbf{0.4567} & \underline{0.9926} \\ 
\rowcolor[HTML]{E8E8E8} \textbf{\model{} (Ours)}  & AR & \textbf{0.2215} & \textbf{0.9000} & \textbf{0.4262} & \underline{0.4148} & \textbf{0.9941} & \underline{0.1806} & \textbf{0.9500} & \textbf{0.4395} & \underline{0.4119} & \textbf{0.9927} \\ 
\bottomrule
\end{tabular}
}
\caption{Experiments on \dataset{} (Ours) for multi-scene 3-min and 5-min video synthesis.}
\label{tab:ourbench-results}
\vspace{-3mm}
\end{table*}
%%%%%%%%%%%%%%%%%%%%%%%%%%%

\paragraph{Multi-Scene Results.} 
\Cref{tab:ourbench-results} reveals the critical challenge of long-horizon consistency, with all baselines degrading notably compared to \Cref{tab:vbench-long-results}. In both 3-min and 5-min settings, baseline character consistency reaches only up to 0.38, while environment peaks at 0.40. Baselines also show significantly lower semantic alignment at 5-min than at 3-min or 1-min, highlighting the difficulty of maintaining prompt fidelity over extended horizons. \model{} achieves superior consistency, outperforming VideoMemory by 16\% on average at 3-min and 13\% at 5-min. It also produces notably more coherent narratives, scoring 10\% higher than VideoMemory while maintaining a motion smoothness of above 0.99. Additionally, baseline narrative scores are higher on \dataset{} than VBench-Long. This is because \dataset{} focuses on multi-scene scenarios which better suit baselines that enforce regular shot transitions.

\paragraph{Scaling Baselines.} 
For fair comparisons, \Cref{tab:scaling-results} experiments with \emph{best-of-N} (\Cref{appdx:bon-prompts}) scaling autoregressive baselines to match the same \#videos and frames sampled per segment as \model{} on VBench-Long, where the best-of-N video is selected by Gemini 3 Pro. We find that consistency indeed improves remarkably for all baselines: Naive-AR improves from 0.61 $\rightarrow$ 0.67 avg. consistency, while VideoMemory improves from 0.65 $\rightarrow$ 0.71. However, narrative coherence is not always the case, with Naive-AR decreases. Meanwhile, \model{} shows significantly more promising test-time scaling potential thanks to its multi-dimensional judges, which can more reliably distinguish quality among candidates, improving from 0.73 $\rightarrow$ 0.78 for avg. consistency and 0.74 $\rightarrow$ 0.9 for coherence.

%%%%%%%%%%%%%%%%%%%%%%%%%%%%%%
\subsection{Human Evaluations} \label{subsec:human-evals}

%%%%%%%%%%%%%%%%%%%%%%%%%%%%%%
\begin{table}[t!]
\centering
\resizebox{1\linewidth}{!}{
\begin{tabular}{l|cccccc|c}
\toprule
\textbf{Method} & \textbf{Character} & \textbf{Object} & \textbf{Environment} & \textbf{Transition} & \textbf{Narrative} & \textbf{Reference} & \textbf{Average} \\
& \textbf{Consistency} & \textbf{Consistency} & \textbf{Consistency} & \textbf{Smoothness} & \textbf{Coherence} & \textbf{Consistency} & \\
\midrule
Naive-AR    & 3.95 & 4.05 & 3.85 & 3.63 & 4.11 & N/A & 3.92 \\
Naive-Par    & 4.19 & 3.72 & 3.66 & 3.34 & 3.84 & N/A & 3.75 \\
MovieAgent \citep{wu2025automated} & 3.40 & 2.80 & 2.77 & 2.20 & 2.39 & 3.20 & 2.79  \\
ViMax \citep{vimax2025} & 4.23 & 3.95 & 3.41 & 2.64 & 3.59 & 4.50 & 3.72 \\
VideoMemory \citep{zhou2026videomemory} & 4.40 & 4.05 & 3.95 & 3.10 & 3.65 & 4.40 & 3.93 \\
\midrule
\rowcolor[HTML]{E8E8E8} \textbf{\model{}-Par (Ours)} & 4.74 & 4.26 & 4.17 & 3.71 & 4.17 & 4.71 & 4.29 \\
\rowcolor[HTML]{E8E8E8} \textbf{\model{} (Ours)}  & \textbf{4.89} & \textbf{4.67} & \textbf{4.52} & \textbf{4.34} & \textbf{4.75} & \textbf{4.91} & \textbf{4.68} \\
\bottomrule
\end{tabular}
}
\vspace{1mm}
\caption{Human evaluation results on 40 VBench-Long samples (1--5 scale).}
\vspace{-5mm}
\label{tab:human-eval}
\end{table}
%%%%%%%%%%%%%%%%%%%%%%%%%%%%%%

\paragraph{Human Metrics.} 
To understand user satisfactions, we conduct a human study on 40 VBench-Long single-scene samples (approx. 40-min per baseline), following a similar scale to \citet{yu2025videossm}. We recruit 7 highly qualified evaluators, each presented with a random subset of generated videos from all methods in randomized, anonymized order. Evaluators rate each video on six criteria on a 5-point Likert scale (1: very poor, 3: acceptable, 5: excellent): \emph{(i) Character Consistency} and \emph{(ii) Object Consistency} (whether characters and objects maintain consistent appearance across shots), \emph{(iii) Environment Consistency} (whether environments and environments remain coherent across scene transitions), \emph{(iv) Transition Smoothness} (whether cuts between segments are visually and temporally natural), \emph{(v) Narrative Coherence} (whether the story progresses logically with meaningful causal relationships), and \emph{(vi) Reference Consistency} (how faithfully the generated video adheres to the provided reference images). Each sample is rated by at least 2 evaluators and scores are averaged.

\paragraph{Human Results.} 
\Cref{tab:human-eval} presents the human results, corroborating our automatic metrics. \model{} achieves the highest scores across all six criteria with an average of 4.68 over 5.00, substantially outperforming the best baseline, VideoMemory (3.93). It excels in character consistency (4.89) and narrative coherence (4.75), confirming strong identity preservation with strong story progression in 1-min. It also scores 4.34 in transition smoothness, significantly higher than the best baseline of 3.34. Reference consistency reaches 4.91, showing potentially strong alignment when users provide reference images. While \model{}-Par maintains good character consistency, it shows notable drops in environment consistency and transition smoothness due to parallel generation from predefined frames. This confirms the benefits of autoregressive generation for both visual and temporal coherence.

\section{Analysis} \label{sec:analysis}

We present our main analyses in this section while additional analyses including qualitative, test-time scaling and latency analyses are provided in \Cref{appdx:additional-analysis}.

%%%%%%%%%%%%%%%%%%%%%%%%%%%%%
\subsection{Ablation Studies} \label{subsec:ablation-studies}

%%%%%%%%%%%%%%%%%%%%%%%%%%%%%
\begin{figure}[t!]
\centering
\begin{minipage}{0.65\linewidth}
\centering
\resizebox{\linewidth}{!}{
\begin{tabular}{l|ccccc}
\toprule
\multirow{2}{*}{\textbf{Method}} & \multirow{2}{*}{\begin{tabular}[c]{@{}c@{}}\textbf{
Semantic}\\\textbf{Alignment}\end{tabular}} & \multirow{2}{*}{\begin{tabular}[c]{@{}c@{}}
\textbf{Narrative}\\\textbf{Coherence}\end{tabular}} & \multirow{2}{*}{\begin{tabular}[c]{@{}c@{}}
\textbf{Inter-shot}\\\textbf{Character}\end{tabular}} & \multirow{2}{*}{\begin{tabular}[c]{@{}c@{}}
\textbf{Inter-shot}\\\textbf{Environment}\end{tabular}} & \multirow{2}{*}{\begin{tabular}[c]{@{}c@{}}
\textbf{Inter-shot}\\\textbf{Motions}\end{tabular}} \\ 
\\
\midrule
\rowcolor[HTML]{E8E8E8} \textbf{\model{}-Par (Ours)}      & 0.2094 & 0.8082 & 0.6817 & 0.8180 &
0.9843 \\
\rowcolor[HTML]{E8E8E8} \textbf{\model{} (Ours)}  & \textbf{0.2111}  & \textbf{0.8987} & {0.7353} & \underline{0.8368} & \textbf{0.9935} \\
\midrule
\model{} w/o \memory{} &  0.1912      & 0.7235      & 0.5061 & 0.7392 & 0.9890\\
\model{} w/o \memory{}'s Text States & 0.2032     & 0.7794   &  0.6791 & 0.7645 & 0.9903 \\
\model{} w/o \memory{}'s Videos & 0.2064      & \underline{0.8647}  & 0.7034      & 0.8153 & 0.9881 \\
\midrule
\model{} w/o HITS & 0.2054      & 0.7395      & 0.6774 & 0.7871 & 0.9911 \\
\model{} w/o MAPO & \underline{0.2095}  & 0.7900      & 0.6843 &  0.8248 & 0.9904 \\
\model{} w/o Global References & 0.2080      & 0.8136      & {0.7107} & 0.7923 & 0.9913 \\
\midrule
\model{} always extrapolates & 0.2054      &  0.8278 & \underline{0.7112} & 0.8136 & 0.9912 \\
\model{} always interpolates & 0.2084     & 0.7094 & \textbf{0.7400} & \textbf{0.8499} & \underline{0.9924} \\
\bottomrule
\end{tabular}
}
\caption{Ablation studies over \model{}'s \memory{}, TTS algorithms, and adaptive segment generation strategies on Vbench-Long.}
\label{tab:ablation-studies}
\end{minipage}
\hspace{2.5mm}
\begin{minipage}{0.27\linewidth}
\centering
\includegraphics[width=\linewidth]{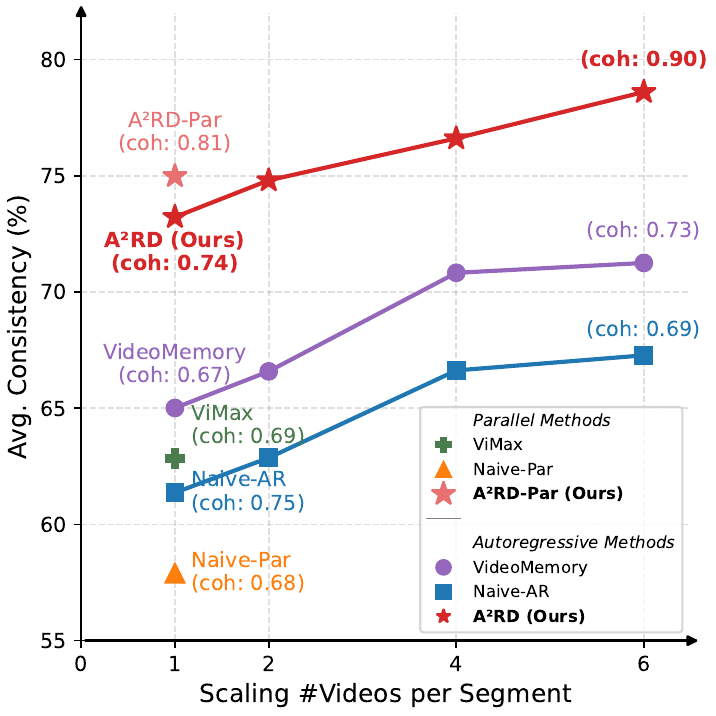}
\vspace{-5mm}
\caption{Consistency versus scaling \#videos per segment.}
\label{tab:scaling-results}
\end{minipage}
\label{tab:ablation}
\vspace{-3mm}
\end{figure}
%%%%%%%%%%%%%%%%%%%%%%%%%%%%%

\textbf{Ablation Setups.} 
We conduct ablation studies on VBench-Long to assess each \model{} component's contribution across three groups: (i) \memory{}'s components including the complete \memory{} system, its Textual States, and its Videos; (ii) Test-time scaling components including global references, HITS, and MAPO (replaced by Self-Refine \citep{madaan2023self}); and (iii) adaptive segment generation strategy when we instead always extrapolate, or interpolate. We omit components that have been comprehensively studied in prior work, such as \memory{}'s Frames.

\paragraph{Ablation Results.} 
\Cref{tab:ablation-studies} shows the critical role of each component. First, \memory{} is the backbone of \model{}; removing it severely degrades performance to near Naive-AR levels, as the system loses long-range dependency conditioning, consistency validation, and HITS. Ablating \memory{}'s individual modalities reveals their contributions: without Textual States, narrative and consistency drop notably, while removing Videos causes less impact, as they are primarily for motion continuity.  Second, test-time scaling components are also critical: we find that removing HITS causes considerable drops (0.90 $\rightarrow$ 0.74 for narratives, 0.74 $\rightarrow$ 0.68 for characters), while without MAPO, the prompt refinements are less effective. Interestingly, we see that removing global references minimally affects narrative and character consistency but notably drops environment consistency (0.84 $\rightarrow$ 0.79), suggesting that environments are harder to maintain and benefit from global references. Finally, the adaptive mechanism is also crucial: always extrapolating maintains reasonable narrative coherence (0.83) but reduces consistency. Meanwhile, always interpolating achieves the highest consistency but at the cost of reduced narrative coherence. In this case, each frame is conditioned on richer context from previous segments, and HITS (Video) further enforce intra-shot consistency, which together can over-constrain the generation and tend to produce limited visual progression.

\subsection{Generalization to Other Video Diffusion Models}\label{subsec:generalization-to-other-models}

%%%%%%%%%%%%%%%%%%%%%%%%%%%%%
\begin{table}[h!]
\centering
\resizebox{.85\linewidth}{!}{
\begin{tabular}{l|ccccc}
\toprule
\multirow{2}{*}{\textbf{Model} / \textbf{Method}} & \multirow{2}{*}{\begin{tabular}[c]{@{}c@{}}\textbf{
Semantic}\\\textbf{Alignment}\end{tabular}} & \multirow{2}{*}{\begin{tabular}[c]{@{}c@{}}
\textbf{Narrative}\\\textbf{Coherence}\end{tabular}} & \multirow{2}{*}{\begin{tabular}[c]{@{}c@{}}
\textbf{Inter-shot}\\\textbf{Character}\end{tabular}} & \multirow{2}{*}{\begin{tabular}[c]{@{}c@{}}
\textbf{Inter-shot}\\\textbf{Environment}\end{tabular}} & \multirow{2}{*}{\begin{tabular}[c]{@{}c@{}}
\textbf{Inter-shot}\\\textbf{Motions}\end{tabular}} \\
\\
\midrule
\textbf{LTX 0.9.8 (13B)} \citep{HaCohen2024LTXVideo} \\
\quad\quad Naive-AR & 0.1884 &  0.5903 & 0.5040 & 0.5788 & 0.9558 \\
\rowcolor[HTML]{E8E8E8}\quad\quad \model{} (Ours) & 0.1978 &  0.7922 & 0.7011 & 0.7719 & 0.9852 \\
\midrule
\textbf{Wan 2.2 (5B)} \citep{wan2025wan} \\
\quad\quad Naive-AR & 0.1584    &  0.6719 & 0.5104 & 0.5898 & 0.9752 \\
\rowcolor[HTML]{E8E8E8}\quad\quad \model{} (Ours) & 0.2219     &  0.8000 & 0.6912 & 0.7824 & 0.9901 \\
\bottomrule
\end{tabular}
}
\vspace{3mm}
\caption{Experimental results with two open-source TI2V models.}
\label{tab:experinemts-with-open-sources}
\end{table}
%%%%%%%%%%%%%%%%%%%%%%%%%%%%%

\paragraph{Setups.} 
To study whether \model{} generalizes across diffusion backbones, we evaluate it with two strong open-source TI2V models: LTX-Video 0.9.8 (13B)~\citep{HaCohen2024LTXVideo} and Wan 2.2 (5B)~\citep{wan2025wan}. For both models, we use $30$ denoising steps with resolution $704 \times 480$ and follow the same evaluation protocol on VBench-Long in Section~\ref{sec:experiments}. Since Wan 2.2 does not support interpolation, we run all \model{} experiments with this backbone using the always-extrapolate mode.

\paragraph{Results.} As shown in Table~\ref{tab:experinemts-with-open-sources}, \model{} improves over Naive-AR across both models. On LTX-Video, it yields notable gains in narrative coherence (0.59 $\rightarrow$ 0.79) and character consistency (0.50 $\rightarrow$ 0.70). Wan 2.2 follows a similar trend, with narrative coherence rising from 0.67 $\rightarrow$ 0.80, character consistency from 0.51 $\rightarrow$ 0.69, and environment consistency from 0.59 $\rightarrow$ 0.78. These results confirm that \model{} generalizes across different video diffusion backbones without costly retraining.

\subsection{Consistency Analysis over Extended Horizons} \label{subsec:consistency-versus-horizons}

\begin{figure}[t!]
    \centering
\includegraphics[width=\textwidth]{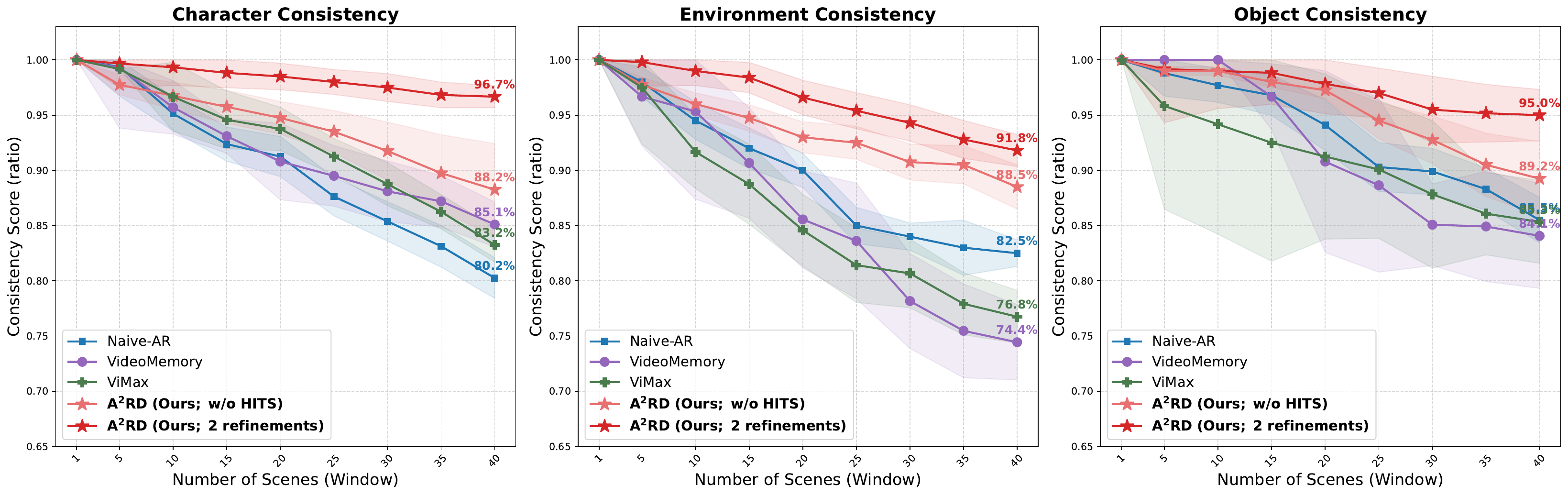}
    \caption{Consistency scores as a function of scene window size (1--40). All methods are evaluated on \dataset{}, 5-minute using LLM-Judge. \model{} (Ours) consistently maintains higher consistency over extended horizons compared to baselines.}
    \label{fig:drigt-analysis}
\end{figure}

\paragraph{Setups.} 
We analyze how consistency degrades as generation extends over longer horizons. Since the automatic metrics in \Cref{subsec:automatic-evals} cannot fully capture the evolving states of entities and environments, we develop an MLLM-Judge method to evaluate consistency in these evolving scenarios. Specifically, after grouping segments that share the same characters, objects, or environments for automatic metrics as in \Cref{appx:metrics-impl}, a carefully calibrated MLLM-Judge with a state-of-the-art MLLM \citep{googledeepmind2026gemini31pro} (\Cref{subsec:dfirt-prompts} for prompts) evaluates each dimension $d \in \{\text{char}, \text{obj}, \text{env}\}$ per segment through 3 steps: \emph{(i) Identify Visual Differences} by comparing with relevant segments' frames and references; \emph{(ii) Classify Inconsistencies} as expected (justified by the story) or unexplained; and \emph{(iii) Flag Inconsistencies} that are unexplained and obvious. For each sample $i$-th, with the same notation introduced in \Cref{sec:methods} and $\mathds{1}$ being the indicator function, the $d$'s consistency ratio is: 
\begin{equation} \label{longveo-eq:drift-consistency-verifications}
\text{Consistency}^d = \frac{\Sigma_{i=1}^{\lvert \mathcal{S} \rvert} \mathds{1}\bigl[\text{MLLM}^{\text{const}}_{\text{vfy}}\text{(}S_i, F^{\text{begin}}_i, \mathcal{F}_{\text{
ref}}, \mathcal{S}_{\text{ref}}, d\text{)}\bigr]}{\lvert \mathcal{S} \rvert}
\end{equation}
\paragraph{Results.} 
Figure~\ref{fig:drigt-analysis} reveals interesting insights. For baselines, we see that they exhibit a monotonic decline in consistency as \#scenes grows, confirming that consistency in long horizons is a non-trivial challenge. For characters, Naive-Par performs worse than VideoMemory and ViMax, which is expected given their dedicated character memory mechanisms. Environment consistency proves the most challenging dimension overall. Interestingly, Naive-Par performs much better than ViMax and VideoMemory here; we attribute this to these baselines forcing frequent shot changes, which causes more often environment hallucinations. Finally, all baselines perform similarly on object consistency, the least challenging dimension, likely because objects do not exhibit complex identities compared to characters and environments. Our method, \model{}, outperforms baselines across all three axes by large margins: it retains {96.7\%} character consistency, {91.8\%} environment consistency, and {95.0\%} object consistency. These results further validate the superiority of \model{} architecture in combating long-horizon consistency degradation.

We manually verified a subset of the MLLM-Judge outputs and found its flags to be effective (80\% agreement) for detecting obvious errors. \Cref{fig:error-analysis} shows representatives. We want to note that subtle inconsistencies, which are very common in synthesized frames, may be missed---by design, as it is calibrated to flag only clear violations. Nevertheless, the method effectively captures the overall trend that \model{} significantly outperforms baselines, aligning with other automatic and human evaluations.

\subsection{Scaling to Longer-Horizon Video Generation} \label{subsec:experiments-10m-dataset}

We experiment with \model{} on ten 10-min scenarios from \dataset{}, approaching the frontier of current long-form video generation. By using the MLLM-Judge method developed above, on average, \model{} attains average consistency of 90.5\% for characters, 84.0\% for environments, and 91.5\% for objects, confirming its strong capability in generating coherent ultra-long videos.

\section{Conclusions}
We presented \model{}, an agentic autoregressive architecture for long video synthesis. By decoupling creative synthesis from consistency, \model{} addresses two fundamental challenges: temporal consistency and narrative coherence. This is achieved through three key components: a multimodal video memory for cross-modal context tracking; an adaptive generation mechanism enabling natural narrative progression with consistency enforcement; and hierarchical test-time self-improving algorithms that self-refine frames and segments to prevent error propagation. We further introduced \dataset{}, a benchmark designed to stress-test long-horizon consistency via cyclical appearance with optional state evolutions. Experiments across 1-10 minute videos show that \model{} sets a new state-of-the-art for narrative coherence and visual consistency compared to existing baselines.

\section*{Limitations}
We acknowledge that \model{} incurs more computational overhead than the baselines experimented in \Cref{sec:experiments}. However, since all baselines follow the passive generation paradigm, direct comparison is therefore the meaningful way to evaluate their capabilities. To ensure fairness, we also provide compute-matched variants of all autoregressive baselines in the Scaling Baselines paragraph. In addition, it is worth noting that \model{}'s active reasoning incurs insignificant costs. Our analysis following \Cref{subsec:latency-analysis} reveals that its cost overhead with using Gemini 3 Flash can be estimated only about less than 0.5\$ per segment (with less than 10K tokens per call on average; excluding costs associated with generating 5 additional videos and 5 additional frames per segment).

In addition, we did not report human agreement scores in our study. This is because several videos are rated by exactly two reviewers, making the agreement not very informative. Additionally, evaluating long-form video generation is inherently subjective, particularly for complex criteria such as transition smoothness and narrative coherence. We decided to average scores over raters, as they are all highly qualified.

Finally, while \model{} made solid progress in long-form video synthesis, several limitations remain. It requires component models (MLLM, TI2I, TI2V models) with strong instruction-following and visual reasoning capabilities. Additionally, the verification rubrics reflect implicit assumptions about consistency and quality that may not generalize to diverse creative styles, cultural contexts, or domain-specific preferences. We encourage adapting them for domain-specific settings. Finally, our user study reveals that transition smoothness and physical environment consistency remain the most challenging aspects for segment-based methods, presenting promising directions for future work.

\section*{Acknowledgements}
We thank Jinsung Yoon, Bhavana Dalvi Mishra, and our colleagues at Google Cloud AI Research for their helpful feedback. We also want to thank Xingchen Wan, Sercan \"O. Ar\i k for several useful discussions, and Nancy F. Chen and Kenji Kawaguchi for supporting Do Xuan Long's internship.

\bibliographystyle{abbrvnat}
\bibliography{main}

\newpage
\appendix

\section{Terminologies and Summary of Major Implemented Functions} \label{appdx:terminologies-and-methods}

\begin{table}[h!]
\centering
\renewcommand{\arraystretch}{1.1}
\footnotesize
\begin{tabularx}{\textwidth}{@{} l X @{}}
\toprule
\textbf{Terminology} & \textbf{Description} \\
\midrule
Shot & A continuous sequence of frames captured from a single camera angle without cuts. \\
Scene & A narrative unit representing continuous action within a single physical environment or location. \\
Segment (Clip) & The fundamental generation unit in $A^{2}RD$, which is flexible and can span one or multiple shots or scenes. \\
Segment Context ($S_i$) & The textual description dictating the narrative, actions, and settings for the $i$-th segment. \\
Storyline ($\mathcal{S}$) & The complete sequential collection of segment contexts $\{S_1, \dots, S_N\}$ defining the full video narrative. \\
Extrapolation & A generation mode that synthesizes a video segment moving forward from only a beginning frame. \\
Interpolation & A generation mode that synthesizes a video segment to seamlessly connect a fixed beginning and ending frame. \\
\bottomrule
\end{tabularx}
\caption{Glossary of key terms used throughout the paper.}
\label{tab:glossary_short}
\end{table}

\begin{table}[h!]
\centering
\renewcommand{\arraystretch}{1.1}
\footnotesize
\begin{tabularx}{\textwidth}{@{} p{2cm} p{4cm} p{5cm} p{2cm} p{2cm} @{}}
\toprule
\textbf{Function} & \textbf{Implemented Variants} & \textbf{Description} & \textbf{Reference} & \textbf{Prompt(s)}  \\
\midrule
$\text{MLLM}_{\text{plan}}$ & $\text{MLLM}^{\text{bg}}_{\text{plan}}$, $\text{MLLM}^{\text{ent}}_{\text{plan}}$ & Plans prompts for establishing global reference backgrounds and entities. & \Cref{longveo-eq:memory-initializations} & \Cref{appdx:init-prompts} \\
\midrule
$\text{MLLM}_{\text{cap}}$ & $\text{MLLM}_{\text{cap}}$ & Generates captions for user-provided reference frames. & \Cref{longveo-eq:memory-initializations} & \Cref{appdx:init-prompts} \\
\midrule
$\text{MLLM}_{\text{dep}}$ & $\text{MLLM}_{\text{dep}}$ & Constructs a dependency graph to determine the synthesis order of global references. & Init (ii) & \Cref{appdx:init-prompts}\\
\midrule
$\text{MLLM}_{\text{mode}}$ & $\text{MLLM}_{\text{mode}}$ & Adaptively determines the segment generation mode. & \Cref{longveo-eq:generation-mode} & \Cref{appdx:mode-prompt} \\
\midrule
$\text{MLLM}_{\text{retr}}$ & $\text{MLLM}^{\text{scenes}}_{\text{retr}}$, $\text{MLLM}^{\text{cont}}_{\text{retr}}$, $\text{MLLM}^{\text{ref}}_{\text{retr}}$, $\text{MLLM}^{\text{img}}_{\text{retr}}$ & Retrieves narratively relevant previous segments, contiguous contexts, and global references. & \Cref{longveo-eq:retrieve,longveo-eq:retrieve-global-refs,longveo-eq:interpolate-extrapolate} & \Cref{appdx:contiguous-scenes,appdx:retrieve-relevant-scenes,appdx:retrieve-global-references,appdx:retrieve-previous-shot} \\
\midrule
$\text{MLLM}_{\text{pgen}}$ & $\text{MLLM}^{\text{img}}_{\text{pgen}}$, $\text{MLLM}^{\text{vid}}_{\text{pgen}}$ & Generates detailed narrative prompts for boundary frames and video segments. & Pipepline (iii) and (iv) & \Cref{appdx:gen-frame-prompt,appdx:gen-vid-prompt} \\
\midrule
$\text{MLLM}_{\text{ext}}$ & $\text{MLLM}^{\text{img}}_{\text{ext}}$, $\text{MLLM}^{\text{vid}}_{\text{ext}}$ & Extracts fine-grained textual states from synthesized frames and videos. & HITS & \Cref{appdx:extract-states-from-frame-prompt,appdx:extract-states-from-video-prompt}\\
\midrule
$\text{MLLM}^{\text{img}}_{\text{vfy}}$ & $\text{MLLM}^{\text{img-imgs}}_{\text{vfy}}$, $\text{MLLM}^{\text{img-st}}_{\text{vfy}}$, $\text{MLLM}^{\text{img-qa}}_{\text{vfy}}$ & Verifies intra-consistency and basic quality for frames. & \Cref{longveo-eq:frame-verifications} & \Cref{appdx:frame-judge-prompt,appdx:frame-judge-prompt-2,appdx:frame-judge-prompt-3,appdx:frame-judge-prompt-4} \\
\midrule
$\text{MLLM}^{\text{vid}}_{\text{vfy}}$ & $\text{MLLM}^{\text{vid-vid}}_{\text{vfy}}$, $\text{MLLM}^{\text{vid-qa}}_{\text{vfy}}$ & Verifies inter- and intra-consistency and basic quality for videos. & \Cref{longveo-eq:vid-verifications} & \Cref{appdx:vid-judge-prompt-1,appdx:vid-judge-prompt-2} \\
\midrule
$\text{MLLM}^{\text{const}}_{\text{vfy}}$ & $\text{MLLM}^{\text{const}}_{\text{vfy}}$ & Evaluates consistency drift over extended horizons. & \Cref{longveo-eq:drift-consistency-verifications} & \Cref{subsec:dfirt-prompts} \\
\midrule
MAPO & Feedback Reasoning, Lesson Synthesis, Prompt Refinement & Reason over judge outcomes,  derive guidelines from past positive and negative refinement cases, and refine prompts targetedly & HITS & \Cref{appdx:mapo-prompt} \\
\bottomrule
\end{tabularx}
\caption{Summary of main MLLM and synthesis functions implemented in \model{}.}
\label{tab:implemented_functions}
\end{table}

\newpage

\section{Additional Analysis} \label{appdx:additional-analysis}

\subsection{General Video Quality Assessments}\label{appdx:general-video-quality}

%%%%%%%%%%%%%%%%%%%%%%%%%%%%%%
\begin{table*}[h!]
\centering
\resizebox{.75\textwidth}{!}{%
\begin{tabular}{l|cccccc}
\toprule
\textbf{Method} & \textbf{Subject} & \textbf{Background} & \textbf{Aesthetic} & \textbf{Imaging} & \textbf{Temporal} & \textbf{Motion} \\
 & \textbf{Consistency} & \textbf{Consistency} & \textbf{Quality} & \textbf{Quality} & \textbf{Flickering} & \textbf{Smoothness} \\
\midrule
Direct Prompting & 0.7713 & 0.8929 & 0.6180 & 0.7160 & 0.9638 & 0.9915 \\
Naive-AR         & 0.9062 & 0.9363 & 0.6546 & 0.7424 & 0.9800 & 0.9898 \\
Naive-Par        & 0.8852 & 0.9148 & \textbf{0.6692} & 0.6417 & 0.9873 & \textbf{0.9924} \\
MovieAgent       & 0.9150 & 0.9386 & 0.6532 & \textbf{0.7450} & 0.9641 & 0.9914 \\
ViMax & \underline{0.9215} & 0.9488 & 0.6181 & 0.7040 & 0.9668 & 0.9914 \\
VideoMemory      & 0.9207 & 0.9492 & 0.6435 & 0.7029 & \underline{0.9907} & 0.9901 \\
\midrule
\rowcolor[HTML]{E8E8E8} \textbf{\model{}-Par (Ours)} & 0.9108 & \textbf{0.9578} & 0.6209 & 0.7257 & \textbf{0.9910} & \underline{0.9917} \\
\rowcolor[HTML]{E8E8E8} \textbf{\model{} (Ours)} & \textbf{0.9231} & \underline{0.9514} & \underline{0.6632} & \underline{0.7431} & 0.9578 & 0.9901 \\
\bottomrule
\end{tabular}%
}
\caption{General video quality over six VBench metrics~\citep{Huang_2024_CVPR}.}
\label{tab:vbench-general}
\end{table*}
%%%%%%%%%%%%%%%%%%%%%%%%%%%%%%

\noindent\Cref{tab:vbench-general} presents the evaluation results on six VBench metrics that assess general video quality. Overall, \model{} performs competitively across all metrics, achieving the best score in Subject Consistency and the second-best in Background Consistency, Aesthetic Quality, and Imaging Quality. The parallel variant, \model{}-Par, obtains the highest scores in Background Consistency and Temporal Flickering. These results confirm that our method does not sacrifice visual fidelity for temporal consistency and narrative coherence.

\subsection{Methodology Design Analysis: Why Does \model{} Work?} \label{appdx:method-design-analysis}

\begin{figure}[h!]
    \centering
\includegraphics[width=\textwidth]{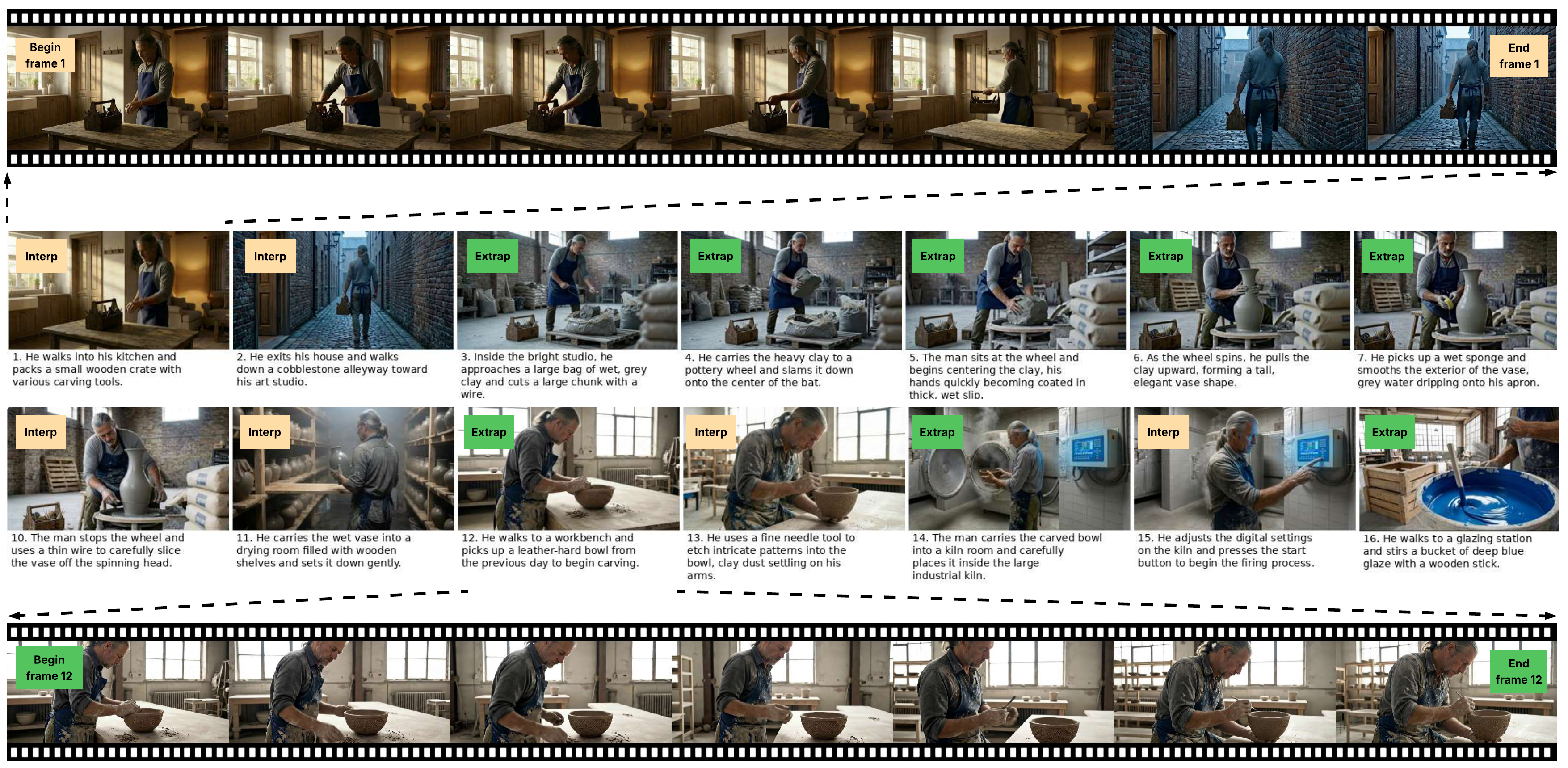}
\caption{An example from \dataset{} about the story of a ceramist illustrating how our Adaptive Segment Generation (\Cref{longveo-eq:generation-mode}) determines the appropriate generation mode.}
\label{fig:adaptive-segment-generation-examples}
\end{figure}

We justify why \model{} works by analyzing its novel components, including Adaptive Segment Generation, Frame-Video Co-Optimization, \memory{}'s Textual States, some criteria in MLLM-Judges, and MAPO. Components such as \memory{}'s keyframes, and other MLLM-Judge criteria have been widely studied in existing works, so we omit them in this section. We denote \textcolor{red}{red} boxes as erroneous and \textcolor{darkgreen}{green} boxes as positive.

\subsubsection{Adaptive Segment Generation Mechanism} 

Adaptive Segment Generation is a core mechanism of \model{} that balances natural narrative progression with consistency enforcement. Here we illustrate an example from \dataset{} where \model{} determines the generation mode (\Cref{longveo-eq:generation-mode}) and why it is effective. In scene context 1, the character packs small items; since the next scene context 2 takes place in a Cobblestone Alleyway, which is a known, established environment during memory initialization, \model{} selects Interpolation mode to enforce strict consistency with that setting. In contrast, in scene context 12, the character begins carving, and in scene 13 he continues the same action; since the next environment is unknown, \model{} selects Extrapolation mode to allow the scene to evolve naturally from the current context, which preserves the natural progression of the room as shown in the keyframes from segment 12.

In practice, we find that state-of-the-art MLLM \citep{googledeepmind2025gemini3flash} performs this task reliably, achieving $>$ 85\% agreement with human experts; misdeterminations mostly involve determining extrapolation as interpolation. We note that such misdeterminations are largely benign, as the image model can sometimes successfully depict highly consistent environments in these cases.

\begin{figure}[t!]
    \centering
\includegraphics[width=\textwidth]{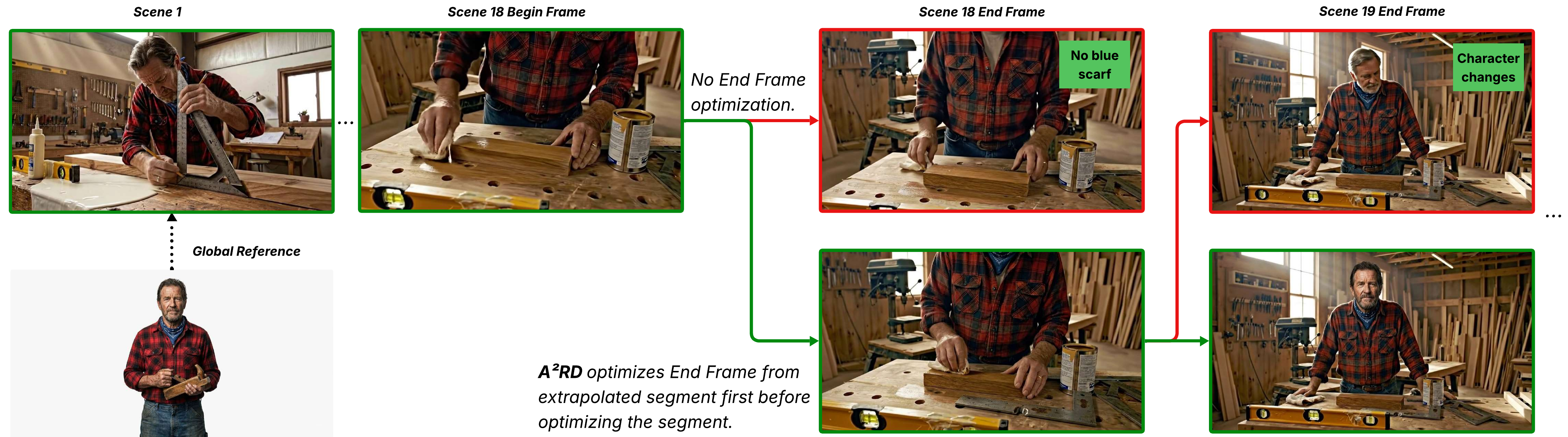}
\caption{Qualitative comparison of with and without \model{}'s end frame co-optimization in extrapolation mode. Without co-optimization, correcting inconsistencies such as a missing blue scarf (Scene 18) and incorrect character identity (Scene 19) persist can be challenging. With \model{}'s co-optimization, these issues are corrected naturally.}
\label{fig:why-co-optimizations}
\end{figure}

\subsubsection{Why Co-Optimization in Extrapolation Mode Is Crucial (\Cref{subsec:memory-augmented-algorithms})}

We present two examples illustrating why co-optimization in HITS is important when $F^{\text{end}}_i$ is unavailable, as shown in \Cref{fig:why-co-optimizations}. The scenario depicts a carpenter working in a woodworking workshop as the camera gradually zooms in and out. We observe that without end frame optimization, the Scene 18 end frame is used directly and the prompt is only optimized to bridge the two frames which miss the blue scarf as an inconsistency. While this blue scarf can sometimes be sufficiently specified in the prompt to produce a similar one during video prompt optimization phase, the situation is more challenging for the Scene 19 end frame, where the character's face in the erroneous frame closely resembles but does not match the person in the reference frame. In that case, \model{} demonstrates its superiority by correcting the character's face to maintain identity consistency.

\begin{figure}[t!]
    \centering
\includegraphics[width=\textwidth]{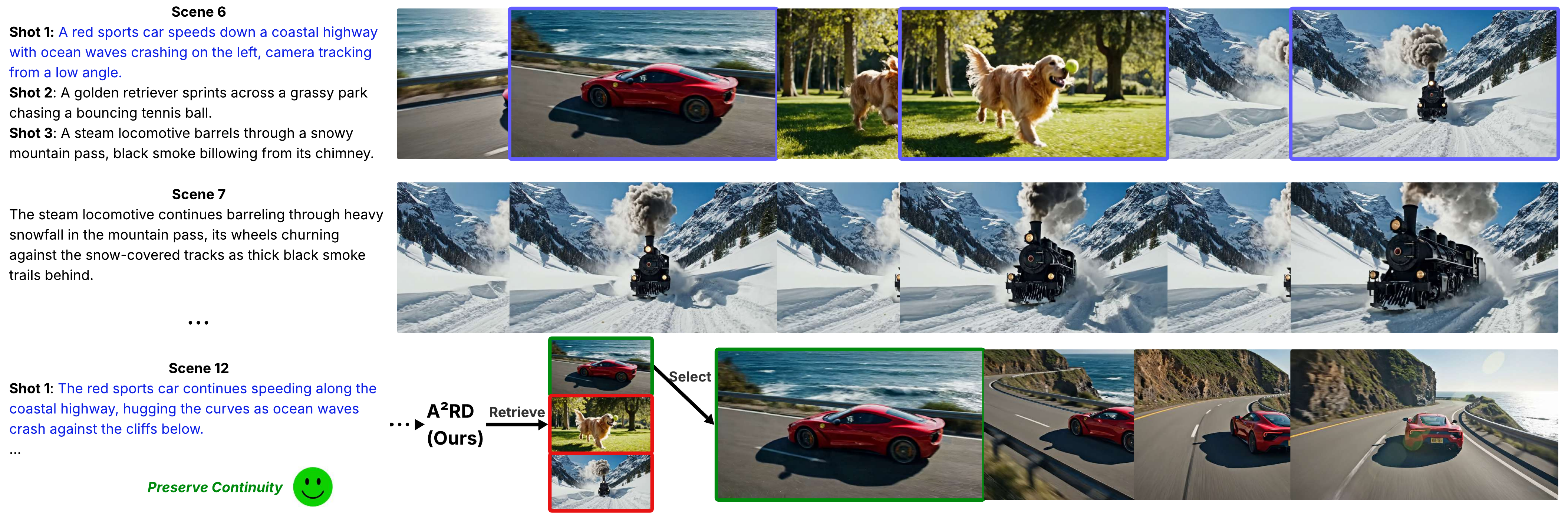}
\caption{Qualitative comparison how \model{} is able to maintain distanced continuity.}
\label{fig:distance-contuinity-example}
\end{figure}

\subsubsection{How \model{} Maintains Long-Distance Motion Continuity ($\text{MLLM}^{\text{img}}_{\text{retr}}$ in \Cref{longveo-eq:interpolate-extrapolate})} \label{appdx-how-aard-maintains-long-distance-continuity}

$\text{MLLM}^{\text{img}}_{\text{retr}}$ is one of our key contributions for bridging long-distance motion continuity, which, to the best of our knowledge, has not been addressed by prior works. Here, we explain how it works.

Consider the example in \Cref{fig:distance-contuinity-example} where Scene 6 contains three shots: a red sports car speeding down a coastal highway, a golden retriever sprinting across a grassy park, and a steam locomotive barreling through a snowy mountain pass. In Scene 7, we continue the locomotive's journey through the snow. Since \model{} only saves the begin and end frames of each segment, the car's end-of-shot frame from Scene 6 is not directly available. Later, in Scene 12, we return to the red sports car, now continuing drifting around a sharp cliffside bend. To handle this, \model{} extracts all end-of- shot frames from the stored video segment of Scene 6, then uses $\text{MLLM}^{\text{img}}_{\text{retr}} $ to retrieve the correct one (Shot 1 of Scene 6) and uses it as the begin frame of Scene 12, thereby preserving visual and motion continuity across the large temporal gap.

\begin{figure}[t!]
    \centering
\includegraphics[width=\textwidth]{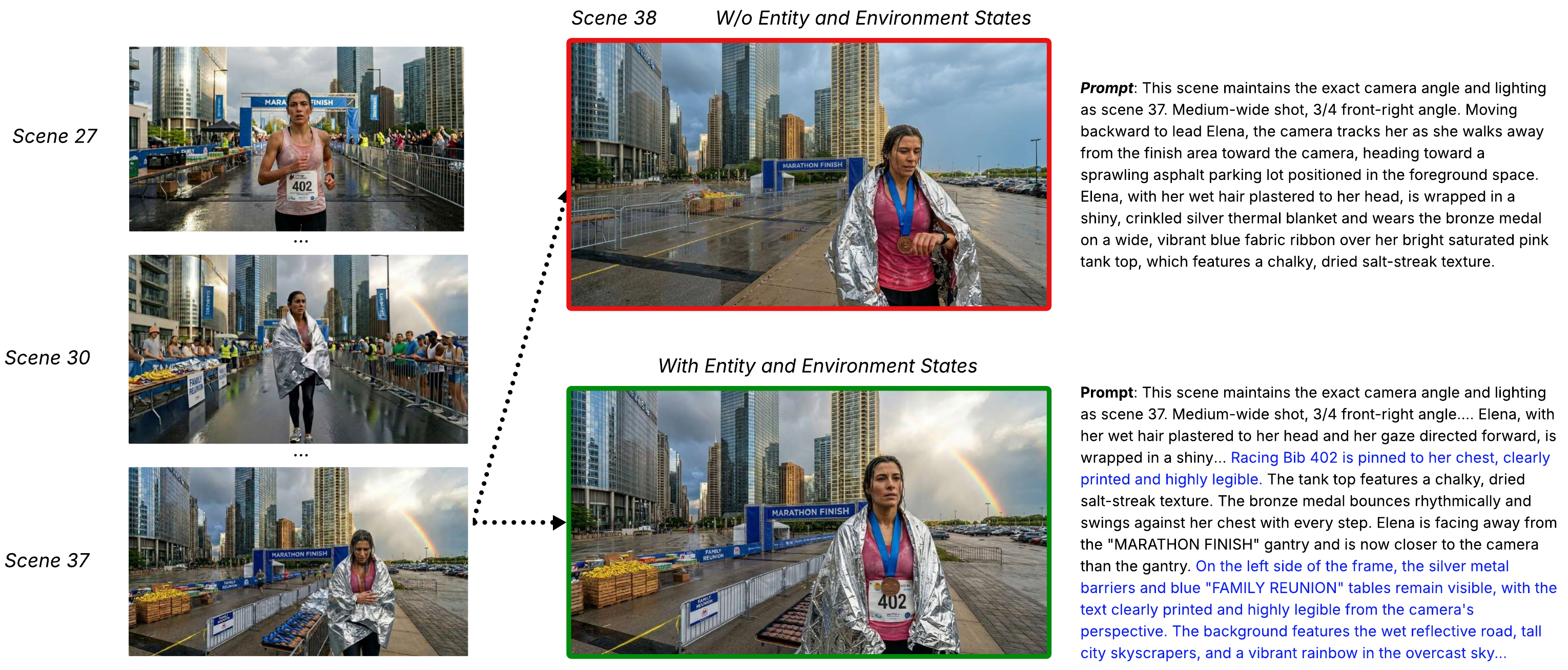}
\caption{An example about the marathon of Elena  showing the importance of \memory{}'s Textual States. With Textual States, key details are explicitly enforced into the prompt, leading to highly consistent outcomes.}
\label{fig:entity-env-necessary}
\end{figure}

\subsubsection{The Role of \memory{}'s Textual States}

Textual states are important for explicitly enforcing necessary details into frame and video prompts for consistency. \Cref{fig:entity-env-necessary} shows an example for frame synthesis. We observe that while the prompt for Scene 38 without entity and environment states is quite comprehensive, several important details were not explicitly specified, leading to the synthesized frame having several inconsistencies. In particular, \texttt{Racing Bib 402} disappears after about 9 scenes, yet it needs to reappear when Elena's lower abdomen is visible. Other minor details such as the \texttt{FAMILY REUNION tables} and the \texttt{rainbow} are also lost. With \memory{}'s Textual States, these components are explicitly retrieved and enforced into the prompt, together with reference frames from previous scenes and global anchors, leading to highly consistent outcome preserving these details.

\begin{figure}[t!]
    \centering
\includegraphics[width=\textwidth]{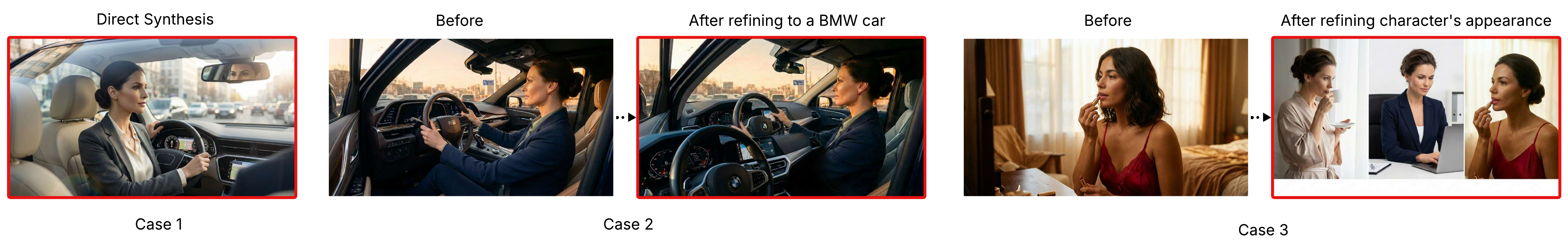}
\caption{An example showing the importance of the Physical Plausibility criterion during frame synthesis and self-improvement. A synthesized frame with physical implausibilities can cause the resulting video to hallucinate at least at the beginning.}
\label{fig:physical-plausibility-necessary}
\end{figure}

\begin{figure}[t!]
\centering
\includegraphics[width=\textwidth]{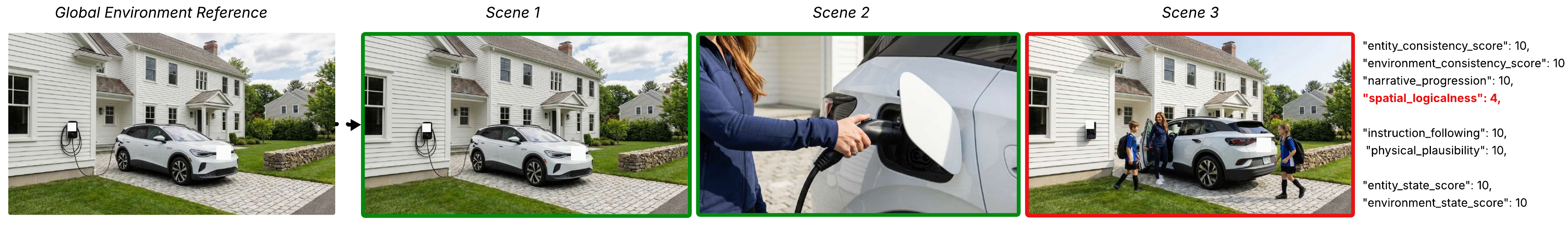}
\vspace{0.5em}
\parbox{\textwidth}{\tiny\textit{%
\textbf{\model{}'s MLLM-Judge}: ``The image captures all elements of the prompt with high fidelity\ldots\ The scene is very coherent and realistic, representing a single unified moment\ldots\ {Progression Discussion:} The frame shows a clear and logical narrative progression from Scene 1 to Scene 2\ldots\ {Spatial Analysis:} \textcolor{blue}{The current scene contains a significant spatial contradiction regarding the orientation of the vehicle when compared to the established environment in Scene 0. In Scene 0, the white car is parked in the driveway facing to the road, exposing its left (driver's) side to the camera. However, in the current scene, the vehicle has been flipped 180 degrees and is now facing to the left of the frame, \textbf{a change not justified by the scene descriptions}}\ldots''%
}}
\caption{An example showing the importance of the Spatial Logicness criterion during frame self-improvement. Scene 3 passes all other criteria yet introduces a spatial discontinuity across segments.}
\label{fig:spatial-logicness-necessary}
\end{figure}

\begin{figure}[t!]
    \centering
\includegraphics[width=\textwidth]{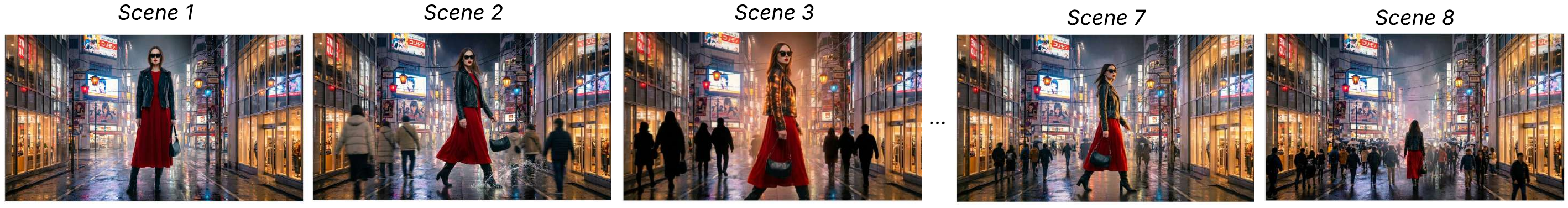}
\caption{An example of a woman walking on a Japanese street, showing that without the Narrative Progression criterion during judging and self-improvement, the resulting storyboard can be highly consistent yet lack meaningful story progression.}
\label{fig:narrative-progression-necessary}
\end{figure}

\subsubsection{Why Do We Need Those MLLM-Judge Criteria?}

From \Cref{subsec:memory-augmented-algorithms}, we see that \model{} requires 8 metrics to verify frames and 10 metrics to verify videos. While some are intuitively important, such as Instruction Following and Inter/Intra Consistency, the importance of others is less obvious, including Physical Plausibility, Spatial Logicalness, and Narrative Progression. Here, we dive into these metrics:

\paragraph{Physical Plausibility (Frame).} 
Since we enforce consistency, synthesized and refined frames can become awkward, where the image model attempts to satisfy the consistency requirement without regard for image quality, as shown in \Cref{fig:physical-plausibility-necessary}. Using these frames as the beginning frame for video synthesis leads to visual hallucinations, at least at the start of the video.

\paragraph{Spatial Logicalness (Frame).} 
Spatial logicalness is an important criterion for maintaining continuity across segments. As shown in \Cref{fig:spatial-logicness-necessary}, Scene 3 achieves perfect scores in all consistency and basic quality metrics. However, synthesizing a segment from Scene 2 and Scene 3 introduces a spatial discontinuity, as the car appears rotated 90 degrees. This breaks the natural continuity between the two scenes, potentially causing hallucination artifacts during video synthesis when interpolating between these frames.

\paragraph{Narrative Progression (Frame and Video).} 
Balancing narrative progression and consistency requires the Narrative Progression criterion. Without it, we can end up with a storyboard that is highly consistent in both entity and environment, as shown in \Cref{fig:narrative-progression-necessary}, yet the story is not meaningful.

\subsubsection{Memory-Augmented Prompt Optimization (MAPO)}

\begin{figure}[t!]
    \centering
\includegraphics[width=\textwidth]{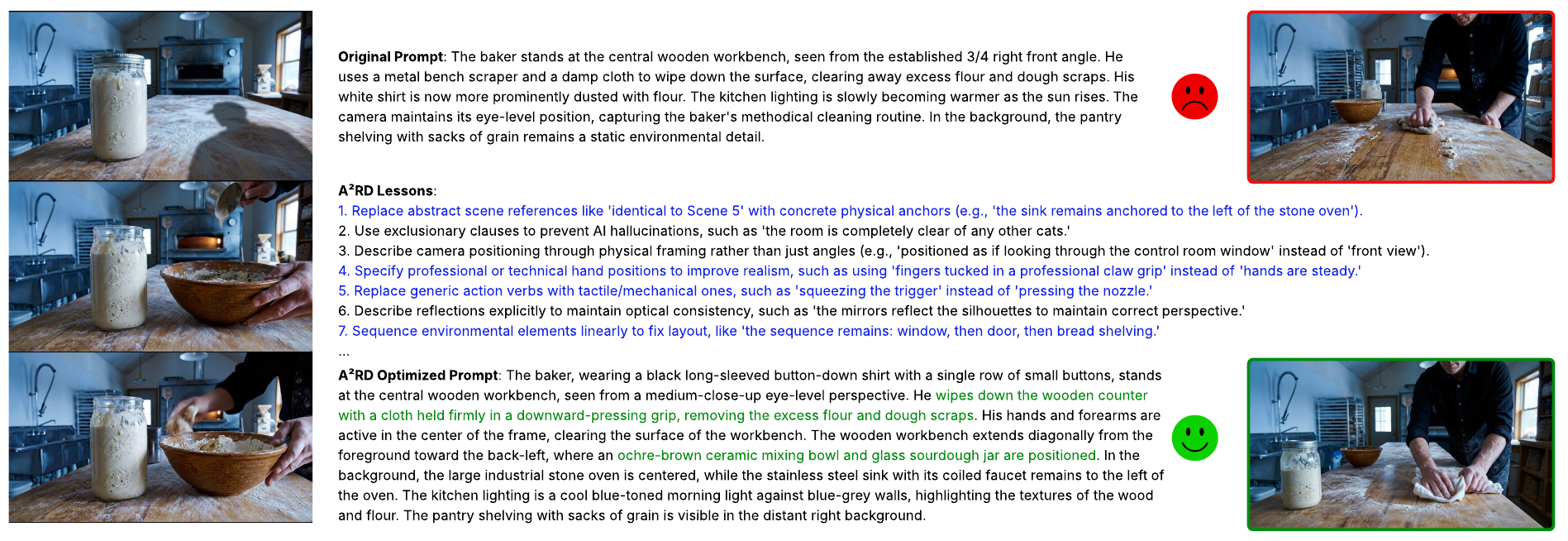}
\caption{An example of MAPO in action. Given the original prompt and validation feedback, MAPO retrieves similar refinement cases from memory, synthesizes actionable lessons, and applies them to produce a refined prompt that directly addresses the failure modes, improving the average score from 6.4 to 8.3.}
\label{fig:narrative-progression-necessary}
\end{figure}

Here, we present an example showing how MAPO works. The original prompt describes a baker who uses a metal bench scraper and a damp cloth to wipe down the surface seen from the established 3/4 right front angle. This prompt contains several points to be improved, such as the camera angle references an abstract established view and not a concrete physical description, and crucially background elements lack spatial ordering.

\model{}'s MAPO retrieves 10 positive and 5 negative cases and synthesizes 12 applicable lessons, notably: (1) \textit{Replace abstract references with concrete physical anchors}, (4) \textit{Specify professional hand positions}, (5) \textit{Replace generic action verbs with tactile/mechanical ones}, and (7) \textit{Sequence environmental elements linearly}. The refined prompt applies lesson (1) to anchor the foreground with a glass sourdough jar on the front-left corner of the workbench and an ochre-brown ceramic mixing bowl in the back-left as concrete physical anchors, and to replace the established 3/4 right front angle with a medium-close-up eye- level perspective; applies lessons (4) and (5) to change the action to wipes down the wooden counter with a cloth held firmly in a downward-pressing grip with hands and forearms active in the center of the frame. The refined frame scores 8.3 on average, with entity reference consistency, environment reference consistency, and spatial logicalness all reaching 10/10, demonstrating that the retrieved lessons directly addressed the failure modes.

\subsection{Error Analysis}

\begin{figure}[h!]
    \centering
\includegraphics[width=\textwidth]{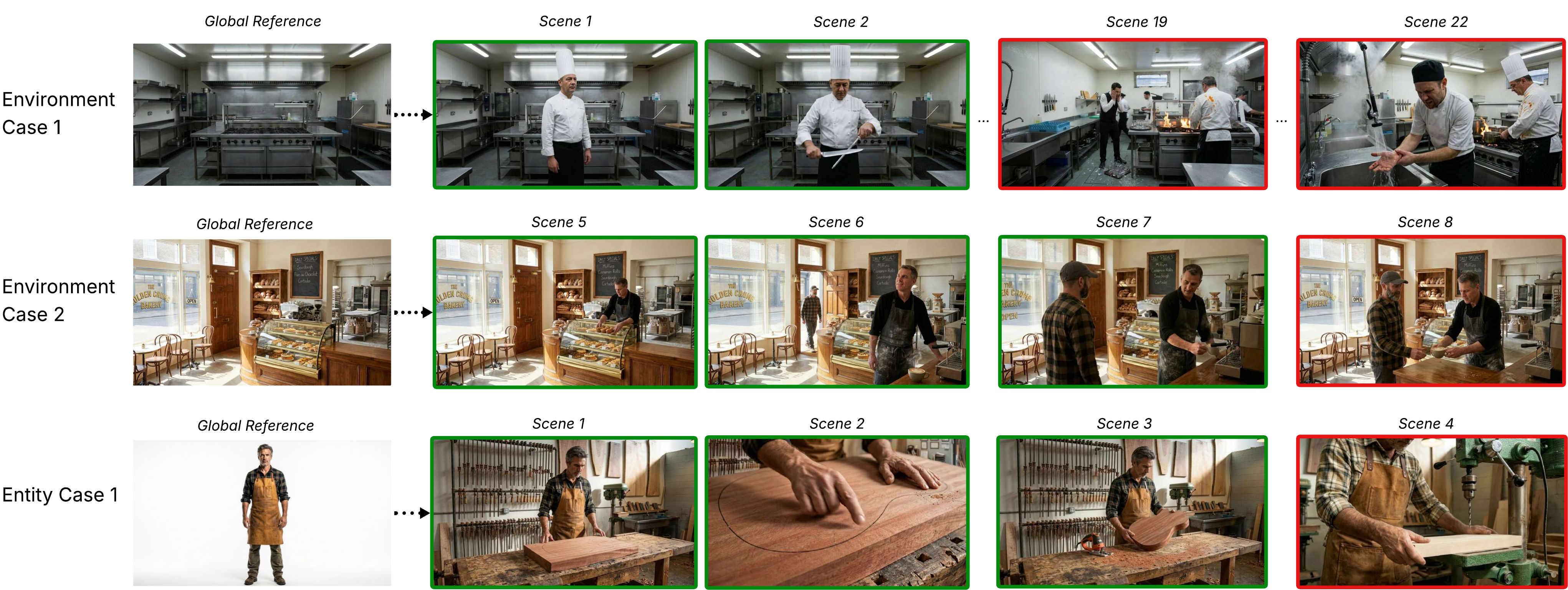}
    \caption{Error analysis of \model{}. Green boxes highlight temporally consistent frames, while red boxes indicate inconsistencies detected by our MLLM-Judge (\Cref{subsec:consistency-versus-horizons}). We present representative failure cases spanning environment and entity consistency, suggesting that advancing foundation model reasoning and physics-aware control are promising directions for future work.}
\label{fig:error-analysis}
\end{figure}

\paragraph{Environment Inconsistencies.} 
Environment consistency remains the most challenging dimension, as layout and physical arrangements are challenging to control during image synthesis. We manually select two representative failure cases flagged by our MLLM-Judge (\Cref{subsec:consistency-versus-horizons}), as shown in \Cref{fig:error-analysis}: in Environment Case 1, the generator produces scenes visually similar but not physically identical to the reference; in Environment Case 2 (Scene 8), a wooden table abruptly appears when the server brings coffee to the customer. Both errors involve either (1) approximating rather than faithfully reproducing the physical layout, or (2) hallucinating physical elements. In addition, we observe that while the MLLM-Judge can effectively identify these issues, the test-time self-refinement loop with our current budget of two refinement iterations is not always sufficient to resolve them.  Addressing these limitations via advancing foundation models with more precise physics-aware control will be a promising direction for future work.

\paragraph{Entity Inconsistencies.} 
We provide a common entity error in \Cref{fig:error-analysis}, where the image generator reproduces the same checkered shirt but with a slightly different color than the original reference. We found that the MLLM-Judge also occasionally fails to detect such minor inconsistencies, suggesting that similar errors could go unnoticed during self-refinement. Addressing this limitation will likely require both more accurate (MLLM-)judges and stronger reasoning foundation models.

\subsection{Test-Time Self-Improvement Analysis} \label{subsec:tts-analysis}

%%%%%%%%%%%%%%%%%%%%%%%%%%
\begin{table*}[h!]
\centering
\resizebox{\textwidth}{!}{
\begin{tabular}{l|c|c|c|c|c|c|c|c|c|c}
\toprule
Iteration & \multicolumn{4}{c|}{Group 1: Consistency over Images} & \multicolumn{3}{c|}{Group 2: Consistency over Texts} & \multicolumn{2}{c|}{Group 3: Basic Quality} & Avg. \\
\midrule
& Entity Ref. & Env. Ref. & Narrative & Spatial & Character & Object & Environment & Instruction & Physical & \\
& Consistency & Consistency & Progression & Logicalness & State & State & State & Following & Plausibility & \\
\midrule
Init & 9.250 & 9.000 & 9.125 & 7.083 & 9.667 & 8.375 & 8.708 & 8.875 & 8.583 & 8.741\\
1 & 9.667 & 9.958 & 9.708 & 8.375 & {9.717} & 9.083 & 9.208 & 8.833 & 8.583 & 9.259\\
2 & \textbf{9.750} & \textbf{9.958} & \textbf{9.792} & \textbf{8.708} & \textbf{9.875} & \textbf{9.375} & \textbf{9.542} & \textbf{8.875} & \textbf{8.792} & \textbf{9.407}\\
\bottomrule
\end{tabular}
}
\caption{Frame-level MLLM-Judge scores across HITS refinement iterations averaged over all benchmarks.}
\end{table*}
%%%%%%%%%%%%%%%%%%%%%%%%%%

% %%%%%%%%%%%%%%%%%%%%%%%%%%
\begin{table*}[t!]
\centering
\resizebox{\textwidth}{!}{
\begin{tabular}{l|c|c|c|c|c|c|c|c|c|c|c|c}
\toprule
Iteration & \multicolumn{5}{c|}{Group 1: Inter Consistency} & \multicolumn{3}{c|}{Group 2: Intra Consistency} & \multicolumn{3}{c}{Group 3: Basic Quality} & Avg. \\
\midrule
 & Inter Entity & Inter Environment & Inter Motion & Camera & Narrative & Character & Object & Environment & Instruction & Physical & Frame &  \\
 & Consistency & Consistency & Consistency & Consistency & Progression & State & State & State & Following & Plausibility & Fit &  \\
\midrule
Init & 7.250 & 7.430 & 7.425 & 7.225 & 8.700 & 9.875 & 9.725 & 9.450 & 8.325 & 8.175 & 9.625 & 7.928 \\
1 & 7.525 & 8.112 & 7.775 & 7.475 & 9.000 & 9.850 & 9.825 & 9.600 & 8.575 & 8.600 & 9.800 & 8.194 \\
2 & \textbf{7.725} & \textbf{8.795} & \textbf{8.050} & \textbf{7.950} & \textbf{9.450} & \textbf{9.800} & \textbf{9.950} & \textbf{9.775} & \textbf{8.950} & \textbf{9.000} & \textbf{9.975} & \textbf{8.493} \\
\bottomrule
\end{tabular}
}
\caption{Video-level MLLM-Judge scores across HITS refinement iterations averaged over all benchmarks.}
\end{table*}
% %%%%%%%%%%%%%%%%%%%%%%%%%%

\paragraph{Setups.} 
We examine the effectiveness of our HITS algorithms in refining frames and videos. We record the MLLM-Judge scores during experiments across two refinement iterations on all benchmarks. These scoring criteria are carefully calibrated by our experts during development.

\paragraph{Results.} At the frame level, we find that most dimensions start off strong, with spatial logicalness (7.083) and physical plausibility (8.583) as the weakest dimensions. After two HITS iterations, both improve noticeably with spatial logicalness jumps from 7.083 $\to$ 8.708, while physical plausibility also sees a steady gain (8.583 $\to$ 8.792). Other dimensions such as entity reference consistency and narrative progression also benefit, pushing the overall average from 8.741 $\to$ 9.407. At the video level, inter-consistency metrics start relatively lower but improve steadily across iterations, with environment consistency showing the largest gain (7.430 $\to$ 8.795). Narrative progression and physical plausibility also benefit noticeably (8.700 $\to$ 9.450 and 8.175 $\to$ 9.000). Intra-consistency dimensions such as character, object, and environment state already start strong and converge near perfect scores by iteration 2. {Overall, the average video score improves from 7.928 $\to$ 8.493, showing that HITS indeed enhances both temporal coherence and consistency. While MLLM scoring could be noisy, the relative trends align with both human judgments (\Cref{tab:human-eval}) and automatic metrics (\Cref{tab:dataset-statistics}): motion and environmental consistency emerge as the weakest dimensions across all evaluations, and removing HITS notably drops coherence and consistency scores. }

\subsection{Estimated Latency Analysis} \label{subsec:latency-analysis} 

Let $L_M$, $L_I$, and $L_V$ denote the latency of a single MLLM call, TI2I call, and TI2V call, respectively. Let $N_{\text{interp}}$ ($N_{\text{interp}} \leq N$) denote the \#interpolation segments. Our estimation analysis is under ideal hardwares, where the batch inference latency is expectedly equal to the one call. Naive-AR, as defined in \Cref{sec:experiments}, incurs approximately $NL_M \text{+} NL_V$, while Naive-Par incurs $NL_M \text{+} \text{(}N\text{+} 1\text{)}L_I \text{+} L_V$.

\subsubsection{Baseline Latency}

\paragraph{MovieAgent \citep{wu2025automated}.} MovieAgent is a passive parallel segment-based method that generates a movie by decomposing a script into scenes and shots, then synthesizing images and videos per shot. It decomposes a script hierarchically across three LLM-driven planning stages: Script breakdown, scene planning, and shot creation, followed by per-shot image and video synthesis. Each stage issues one LLM call per unit (sub-script, scene, and shot, respectively). Let $N_s$, $N_c$, and $N$ denote the number of sub-scripts, scenes, and shots, and assuming all frames and videos can be synthesized in parallel (even though in the official implementation the authors process them sequentially), the total ideal latency is estimated as:
\begin{equation}\label{longveo-eq:movieagent-latency}
L_{\text{MovieAgent}} \approx \text{(}4 \text{+} N_s \text{+} N_c\text{)}L_M \text{+} L_I \text{+} L_V
\end{equation}

\paragraph{VideoMemory \citep{zhou2026videomemory}.} VideoMemory is an autoregressive segment-based method with a dynamic memory bank. It first calls StoryboardAgent (once) with 1 MLLM call to decompose synopsis into $N$ shots. For each shot $i$ sequentially:
\begin{itemize}
    \item MemoryAgentAnalyze (per shot): 1 MLLM call ($L_M$) to decide REUSE or CREATE for each entity. For each entity that needs CREATE: 1 TI2I call ($L_I$). Let $E_i$ = number of new entities in shot $i$, so this costs $L_M \text{+} E_i L_I$ per shot.
    \item VisualizeAgentKeyframe (per shot): 1 MLLM call ($L_M$) to build prompts, then 1 TI2I call ($L_I$) for the keyframe, then 1 TI2V call ($L_V$) for the video. Cost: $L_M \text{+} L_I \text{+} L_V$ per shot.
\end{itemize}

The sum of $E_i$ terms is bounded by the total number of unique entities across all shots. Assume it is $|\mathcal{R}|$ for compatibility. The total ideal latency is estimated as,

\begin{equation}\label{longveo-eq:videomemory-latency}
L_{\text{VideoMemory}} \approx \text{(}1 \text{+} 2N\text{)}L_M \text{+} \text{(}N \text{+} |\mathcal{R}|\text{)}L_I \text{+} NL_V    
\end{equation}

\subsubsection{\model{} Latency}

We omit the case $t \neq i - 1$ (\Cref{longveo-eq:interpolate-extrapolate}) in this analysis for simplicity. 

\paragraph{\memory{} Init.} This step incurs $2L_M$ from 2 batch MLLM calls: one for batch inference obtaining $\mathcal{P}_\mathcal{R}^{\text{bg}}$, $\mathcal{P}_\mathcal{R}^{\text{ent}}$, $\mathcal{P}_\mathcal{R}^{\text{u}}$ in \Cref{longveo-eq:memory-initializations} and one for identifying dependencies (step ($ii$)). It incurs at most $|\mathcal{R}| \times L_I$ for synthesizing global references, thus the maximum init latency is $L_{\text{init}} \approx 2L_M \text{+} |\mathcal{R}| L_I$. 

\paragraph{Precomputations.} As mentioned in \Cref{subsec:experiment-setups}, we precompute all retrieval steps in \Cref{longveo-eq:retrieve,longveo-eq:retrieve-global-refs}, and the segment generation mode in \Cref{longveo-eq:generation-mode}, which incur 1 batch MLLM call $L_M$.

\paragraph{Synthesize and Self-Improve--Boundary Frame(s).} 
\model{} incurs an initial latency of $\text{(}N_{\text{interp}} \text{+} 1\text{)}L_M$ for generating frame prompts ($P_0^{\text{begin}}$/$P_i^{\text{end}}$ in \Cref{longveo-eq:interpolate-extrapolate}). Each prompt is subjected to a maximum of $k_f$ refinement iterations. During each, \model{} first synthesizes a batch of candidate frames via a TI2I call ($L_I$), extracts the textual states $T^F_i$ via a batched MLLM call ($L_M$), and performs verification using another batched MLLM call ($L_M$). The best candidate then undergoes at most three subsequent MLLM calls ($3L_M$) for test-time optimization (allocated for feedback reasoning, lesson generation, and prompt refinement, respectively). In the final iteration, the best candidate frame is returned without optimization. Consequently, the approximated maximum refinement latency per segment is $k_f \text{(}L_I \text{+} 2L_M\text{)} \text{+} \text{(}k_f - 1\text{)}3L_M$, which simplifies to $k_f \text{(}5L_M \text{+} L_I\text{)} - 3L_M$. Scaling this across all interpolated segments and including the initial prompt generation, the total latency for boundary frame processing is approximated at most as $L_{\text{frame}} \approx \text{(}N_{\text{interp}} \text{+} 1\text{)} \left[k_f \text{(}5L_M \text{+} L_I\text{)} - 2L_M\right]$.

\paragraph{Synthesize and Self-Improve--Video Segment.} 
Similarly, we can derive the latency of video segment synthesis. \model{} incurs an initial latency of $N L_M$ for generating the video prompts ($P_i$ in \Cref{longveo-eq:video-syn}). Each prompt is subjected to a maximum of $k_v$ refinement iterations. During each, \model{} first synthesizes a batch of candidate video segments via a TI2V call ($L_V$), extracts the video states $T_i$ via a batched MLLM call ($L_M$), and performs verification using another batched MLLM call ($L_M$). The best candidate is then selected to undergo a refinement process contingent upon the generation mode. For interpolation, the agent utilizes one subsequent MLLM calls ($3L_M$) for test-time prompt optimization (allocated for feedback reasoning, lesson generation, and prompt refinement, respectively). For extrapolation, \model{} incurs an additional computational cost to extract and \emph{edit} the end frame. Because editing solely invokes feedback reasoning without lesson generation and prompt refinement, this frame edit iteration requires only $\text{(}3L_M \text{+} L_I\text{)}$. Skipping the reasoning step in the final edit iteration, the extrapolation edit overhead becomes $k_f \text{(}3L_M \text{+} L_I\text{)} - L_M$. In the final iteration, the best candidate video is returned directly without the optimization process. Aggregating these across all segments and incorporating the extrapolation overhead, the latency is approximated as $L_{\text{video}} \approx N \left[k_v\text{(}5L_M \text{+} L_V\text{)} - 2L_M\right] \text{+} \text{(}N - N_{\text{interp}}\text{)} \left[k_f \text{(}3L_M \text{+} L_I\text{)} - L_M\right]$.

\paragraph{\model{} Latency.} In total, $L_{\text{\model{}}} = L_{\text{init}} \text{+} L_{\text{pre}} \text{+} L_{\text{frame}} \text{+} L_{\text{video}}$, which can be simplified as:
\begin{equation}\label{longveo-eq:aard-latency}
L_{\text{\model{}}} \approx \left[N\text{(}5k_v \text{+} 3k_f - 3\text{)} \text{+} N_{\text{interp}}\text{(}2k_f - 1\text{)} \text{+} 5k_f \text{+} 1\right] L_M \text{+} \text{(}|\mathcal{R}| \text{+} N k_f \text{+} k_f\text{)} L_I \text{+} \text{(}N k_v\text{)} L_V
\end{equation}
\paragraph{\model{}-Par Latency.} In \model{}-Par, we set $N_{\text{interp}} = N$ and $k_v = 1$. Because video synthesis is fully parallelized, its latency does not scale with $N$ and instead incurs $L_M \text{+} L_V$. Thus, 
\begin{equation}\label{longveo-eq:aard-parallel-latency}
L_{\text{\model{}-Par}} \approx \left[N\text{(}5k_f - 2\text{)} \text{+} 5k_f \text{+} 2\right] L_M \text{+} \text{(}|\mathcal{R}| \text{+} N k_f \text{+} k_f\text{)} L_I \text{+} L_V
\end{equation}

Compared to \Cref{longveo-eq:videomemory-latency}, \model{} incurs at most $\frac{5k_v \text{+} 5k_f - 6}{2}$ additional MLLM calls, $k_f - 1$ additional TI2I calls, and $k_v - 1$ additional TI2V calls per segment due to iterative self-improvement. In practice, we set $k_v = k_f = 3$ (one initial generation followed by two refinement iterations), resulting in roughly 12 additional MLLM calls, 2 additional TI2I calls, and 2 additional TI2V calls per segment compared to VideoMemory. With observed latencies of under 10 seconds per MLLM call, 30 seconds per TI2I call (Nano Banana), and under 120 seconds per TI2V call (Veo 3.1), \model{} incurs a total computational overhead of under 7.2 minutes per segment. 

\begin{figure}[h!]
    \centering
\includegraphics[width=.9\textwidth]{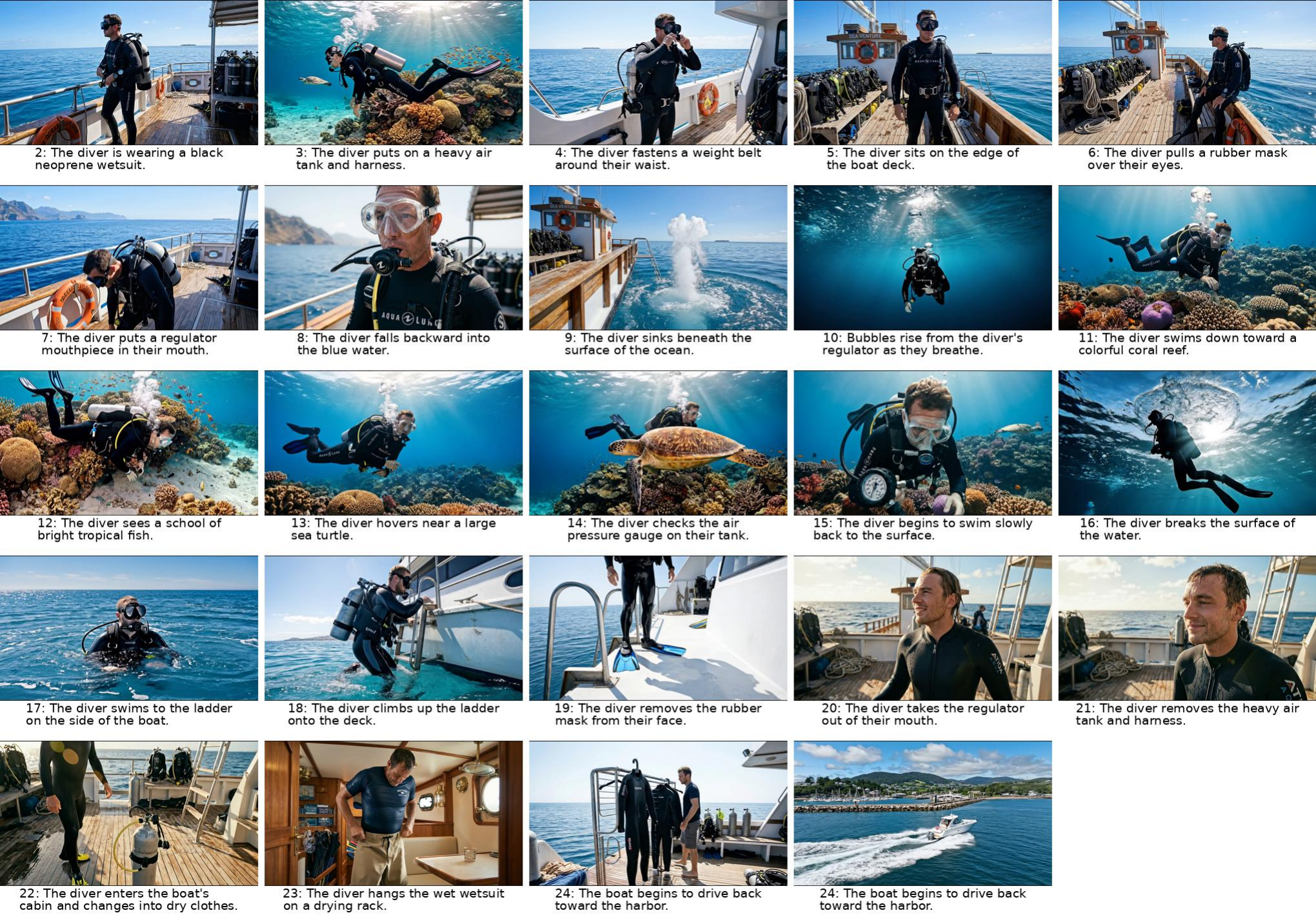}
\caption{Storyboard synthesized by VideoMemory of a 3-minute video from \dataset{}: \textit{The Scuba Diver’s Reef
Exploration} (24 scenes). Frequent inconsistencies in characters, objects, and environments accumulate throughout the video, severely degrading narrative coherence and visual continuity.}
\label{fig:videomemory-3m-example}
\end{figure}

\begin{figure}[h!]
    \centering
\includegraphics[width=.9\textwidth]{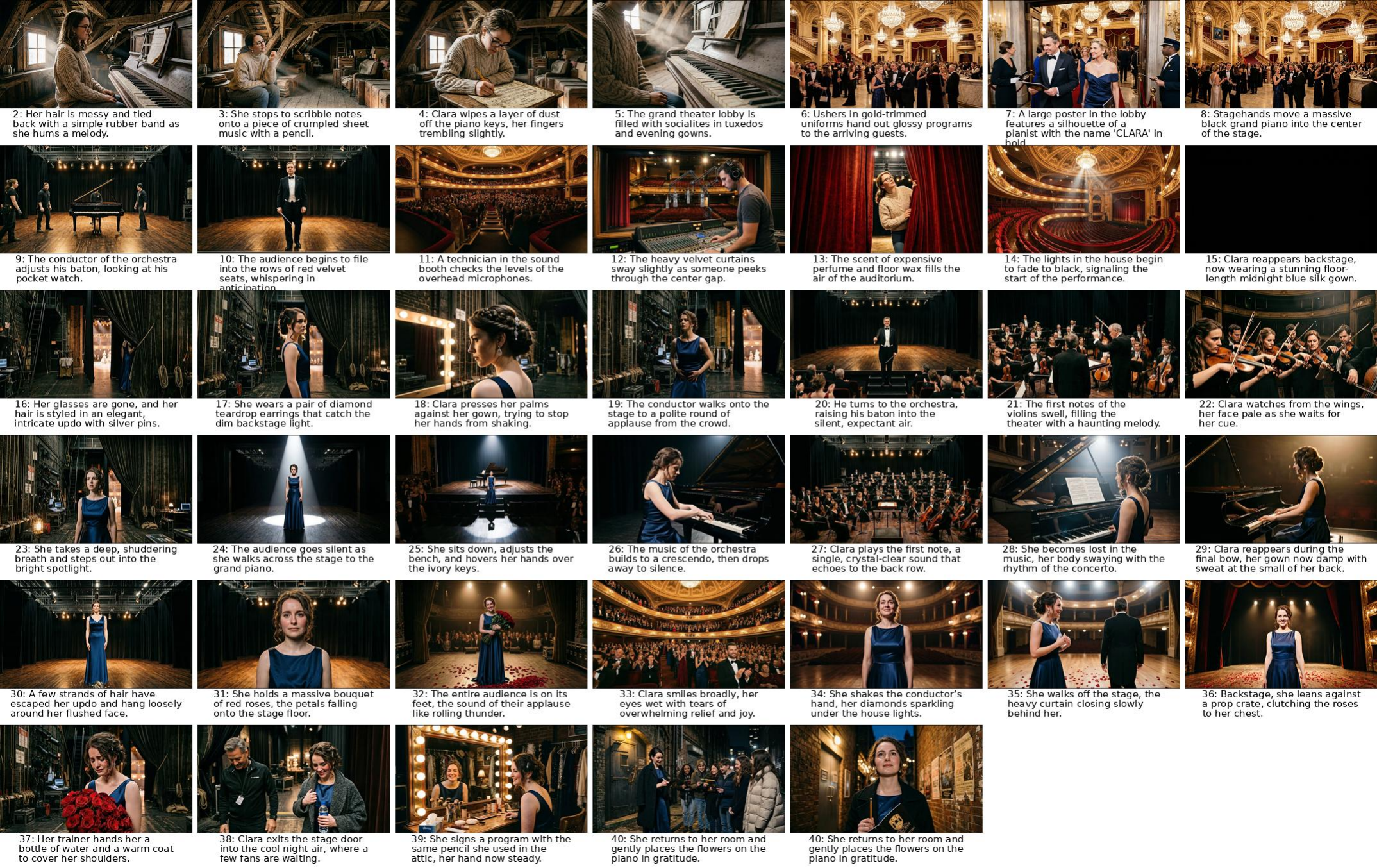}
\caption{Storyboard synthesized by VideoMemory of a 5-minute video from \dataset{}: \textit{The Stage Fright (Clara)} (40 scenes). Frequent inconsistencies in characters, objects, and environments accumulate throughout the video, severely degrading narrative coherence and visual continuity.}
\label{fig:videomemory-5m-example}
\end{figure}

Although this added latency can be non-trivial for standard real-time LLM applications (e.g., question answering), within the computationally intensive domain of video synthesis, we argue that it is highly tractable for several reasons. First, none of the baselines, including the strongest one, VideoMemory, produce strong consistent videos when scaled to longer horizons; see \Cref{fig:videomemory-3m-example,fig:videomemory-5m-example} for examples. \model{} introduces self-improvement phases specifically to fix those highly frequent inconsistencies. Second, to achieve comparable consistency with \model{} conventionally, a human operator would need to manually inspect segments, iteratively refine prompts, and sequentially regenerate video segments, a process that typically requires tens of minutes per segment. Furthermore, this overhead is also cost-effective: MLLM calls act as a lightweight gating mechanism that catch inconsistencies early, preventing the significant computational and financial waste associated with fully rendering and discarding failed videos. Finally, when contextualized within traditional film-making production pipelines, where manual scripting, storyboarding, and editing can consume hours or days for a multi-minute video, an 7.2-minute automated per segment is justifiable. In summary, the latency introduced by \model{} represents a justified trade-off for substantially improved generative yield and autonomous scalability.

\subsection{Use of AI Assistants}
The authors used AI-assisted tools (ChatGPT, Gemini, and Claude) for coding and writing support. All substantive content, analysis, and conclusions remain solely the authors' own work.

\section{Implementation Details}\label{appdx:implemntation-details}

\subsection{Automatic Metrics' Implementations} \label{appx:metrics-impl}

\begin{itemize}
    \item \emph{Semantic Alignment} is computed using ViCLIP~\citep{wang2023internvid}, measuring cosine similarity between video and scene description embeddings.
    \item \emph{Narrative Coherence} is evaluated by an MLLM prompted with the original story and the generated video. See the prompt below.
    \item \emph{Inter-Shot Character and Environment}: we first group $N$ scenes into two types of groups via the MLLM: groups having the same characters with the same states, groups having the same environments with the same states. Once these groups are established, we evaluate the \emph{Inter-Shot Character} and \emph{Inter-Shot Environment} consistency. We use YOLOv8~\citep{varghese2024yolov8} to segment character and background regions within the keyframes of each scene within each group, extracting their corresponding DINOv3~\citep{simeoni2025dinov3} embeddings. We then compute the pairwise cosine similarities of these embeddings across scenes within each group, and average the results across all groups to obtain the final metrics.
    \item \emph{Inter-Shot Motions}: For each scene, we first identify its previous continuous scene (if exists), see \Cref{appdx:contiguous-scenes}. For each couple, we then evaluate the transitions between the two synthesized video segments by extracting the last 10 frames of the preceding video segment and the first 10 frames of the subsequent video segment. We then compute the VBench \texttt{motion\_smoothness} score across these boundary frames.
    \item \emph{Intra-shot Consistency}: We use those metrics from VBench.
\end{itemize}

\subsubsection{Best-of-N Baseline's Prompts}\label{appdx:bon-prompts}

\begin{lstlisting}
You are an expert image quality evaluator. Given {n} candidate keyframe images generated from the same scene description and reference entities, select the single best candidate.

Scene Description: {scene_description}

Reference entities:
[reference_image_parts]

Candidates:
[candidate_image_parts]

Evaluate each candidate on the following criteria:
1. Scene faithfulness – how well the image matches the described scene
2. Visual quality – sharpness, composition, absence of artifacts
3. Entity consistency – how closely characters and objects match the reference images

Select the candidate that best satisfies all criteria holistically.
Respond ONLY with JSON: {"best": <1-{n}>, "reason": "<brief justification>"}
\end{lstlisting}
\vspace{3mm}
\begin{lstlisting}
You are an expert video quality evaluator. Given {n} candidate video clips generated from the same prompt and previous clip, select the single best candidate.

Scene Description: {scene_description}

Previous clip:
[previous_video_part]

Candidates:
[candidate_video_parts]

Evaluate each candidate on the following criteria:
1. Prompt faithfulness – how well the video matches the described scene
2. Visual quality – sharpness, color accuracy, absence of artifacts
3. Motion naturalness – smooth, physically plausible continuation from previous clip

Select the candidate that best satisfies all criteria holistically.
Respond ONLY with JSON: {"best": <1-{n}>, "reason": "<brief justification>"}
\end{lstlisting}

\subsubsection{Prompt for Narrative Coherence Evaluation}
\begin{lstlisting}
You are evaluating the narrative coherence of a video story.

Context (For Reference Only): {story}

Watch the video and evaluate its narrative coherence on a scale of 0 to 1, where:
- 0.9-1.0 = Almost perfect: consistent story, character, object, and environment progression
- 0.6-0.8 = Good: story events follow a logical progression with clear cause-and-effect, but minor inconsistencies in character appearance, object continuity, or environment identity are present
- 0.3-0.5 = Moderate: story flow is partially coherent but noticeable inconsistencies in character appearance, object continuity, or environment break immersion, or scenes feel loosely connected without clear cause-and-effect
- 0.0-0.2 = Poor: story flow is largely incoherent with major inconsistencies across character, object, or environment, or the narrative is broken, incomprehensible, or highly repetitive without story justification

Consider and PENALIZE heavily for:
1. Story progression - What's wrong with the story flow from scene to scene? Do events follow logically from prior events? Are cause-and-effect relationships between scenes clear and believable? Penalize if scenes feel disconnected or outcomes appear without plausible causes.
2. Character progression - What's wrong with character appearance or identity progression? Does the character's state or condition change causally as a result of story events?
3. Object progression - What's wrong with object progression across scenes? Do objects appear, change, or disappear in ways that are causally justified by the story?
4. Environment progression - What's wrong with the setting progression? Are environment changes causally motivated by the story rather than arbitrary?
5. Repetitive penalties - If repetitive activities or environments appear in the video but are NOT present in the Context, the score MUST NOT exceed 0.6. If the Context itself specifies repetitive actions or settings, do not penalize for repetition.

IMPORTANT: Heavily penalize any character appearance progression, object progression issues, or environment shifts that break immersion. If the video, character, or environment as a whole makes no sense, is imperceptible, or is incomprehensible, the score MUST NOT exceed 0.2.

Output ONLY a JSON object wrapped in ```json and ```:
```json
{
  "narrative_coherence": <score>,
  "reasoning": "details of the reasoning"
}
```
\end{lstlisting}

\subsubsection{Prompt for Grouping Scenes for Inter-Shot Metrics}

\begin{lstlisting}
Analyze this video scenario and its scenes to identify which scenes share the same background/location and which scenes need the same character or object appearance (visual consistency).

Scenario: {scenario}

Scenes:
{chr(10).join(f"{i}. {scene}" for i, scene in enumerate(scenes))}

Output JSON with:
- "background_groups": array of arrays, each inner array contains scene indices (0-based) that share the same background/location
- "character_groups": array of arrays, each inner array contains scene indices (0-based) where the same character should have consistent appearance (same face, body type, cloths, core features)
- "object_groups": array of arrays, each inner array contains scene indices (0-based) where the same object should have consistent appearance (same color, shape, material)

Example: {{"background_groups": [[0,1,2], [3,4]], "character_groups": [[0,1,3], [2,4,5]], "object_groups": [[0,2,4]]}}

Output only the JSON object:
\end{lstlisting}

\subsection{Human Metrics' Scoring Guidelines}

Thank you for participating in our human evaluation study! This human evaluation will take approximately 30 minutes. You will be presented with a randomly sampled subset of generated videos from all methods in \textbf{randomized, anonymized order}. Each video consists of multiple \textbf{8-second segments}. For each video, please watch it, review the reference images and scene descriptions, then submit your ratings. Please rate each video on \textbf{six criteria} using a \textbf{5-point Likert scale}:

\begin{table}[h]
\centering
\begin{tabular}{|c|l|}
\hline
\textbf{Score} & \textbf{Meaning} \\ \hline
1 & Very poor \\ \hline
2 & Poor \\ \hline
3 & Acceptable/Ok to Watch \\ \hline
4 & Good \\ \hline
5 & Excellent \\ \hline
\end{tabular}
\end{table}

\subsection*{Evaluation Criteria}
\begin{enumerate}
    \item \textbf{Character Consistency}: Do characters maintain consistent appearance across video?
    \item \textbf{Object Consistency}: Do objects maintain consistent appearance across video?
    \item \textbf{Environment Consistency}: Do backgrounds and environments remain consistent across transitions?
    \item \textbf{Transition Smoothness}: Are the cuts between segments visually and temporally natural?
    \item \textbf{Narrative Coherence}: Does the story progress logically with meaningful causal relationships?
    \item \textbf{Reference Consistency}: How faithfully does the generated video adhere to the provided reference images? N/A if no reference images are provided.
\end{enumerate}

\section{\dataset{}: Examples}

%%%%%%%%%%%%%%%%%%%%%%%%%%%%%%
\begin{figure}[ht!]
\centering
\begin{tcolorbox}[colback=gray!5, colframe=black!75]
\footnotesize
\textbf{Scenes 1-5 (Dutch oven \& Chef appear - Initial cooking):}
\begin{enumerate}[leftmargin=*, itemsep=0pt, topsep=2pt]
\item A heavy cast-iron Dutch oven sits empty and cold on a gas stove.
\item A chef pours golden olive oil into the Dutch oven as the flame ignites below.
\item Chopped onions and garlic are tossed into the Dutch oven, sizzling in the hot oil.
\item Slabs of raw beef are added to the Dutch oven, browning quickly against the metal.
\item A splash of red wine is poured into the Dutch oven, deglazing the bottom as steam rises.
\end{enumerate}

\textbf{Scenes 6-15 (Chef transitions to prep - Dutch oven absent):}
\begin{enumerate}[leftmargin=*, itemsep=0pt, topsep=2pt, resume]
\item The chef walks to the pantry to grab a bag of fresh organic carrots.
\item He peels the carrots over a compost bin with quick, rhythmic strokes.
\item The carrots are sliced into thick medallions on a heavy wooden cutting board.
\item A bundle of fresh thyme and rosemary is tied together with kitchen twine.
\item The chef cleans his professional knife carefully under a stream of warm water.
\item He sets the dining table with linen napkins and polished silver cutlery.
\item Two crystal wine glasses are placed precisely next to the dinner plates.
\item A crusty baguette is sliced and placed into a decorative wicker bread basket.
\item The chef checks his watch, noting the time remaining for the slow-cooking process.
\item He wipes down the marble countertop until it shines under the bright kitchen lights.
\end{enumerate}

\textbf{Scenes 16-20 (Return to Dutch oven - Serving):}
\begin{enumerate}[leftmargin=*, itemsep=0pt, topsep=2pt, resume]
\item The Dutch oven is now filled with a thick, bubbling beef stew and tender vegetables.
\item The chef lifts the lid of the Dutch oven, releasing a dense cloud of savory steam.
\item He ladles the rich stew from the Dutch oven into a large ceramic serving bowl.
\item The Dutch oven is moved to a heat-proof mat, its exterior now stained with dried drips.
\item He sprinkles fresh parsley over the stew inside the Dutch oven before serving.
\end{enumerate}

\textbf{Scenes 21-24 (Dining room - Final scene):}
\begin{enumerate}[leftmargin=*, itemsep=0pt, topsep=2pt, resume]
\item Guests enter the dining room, reacting to the rich aroma of the cooked meal.
\item The chef carries the serving bowl to the table as guests take their seats.
\item Everyone begins to eat, enjoying the deep flavors developed over several hours.
\item The chef smiles as he watches his friends enjoy the hearty homemade dinner.
\end{enumerate}
\end{tcolorbox}
\caption{\textbf{Example 3 minute (24 scenes) scenario from \dataset{}, Object State Evolving}: The Dutch oven appears in (1-5), disappears in (6-15), then reappears in (16-20).}\label{fig:example_3m_scenario}
\end{figure}
%%%%%%%%%%%%%%%%%%%%%%%%%%%%%%

%%%%%%%%%%%%%%%%%%%%%%%%%%%%%%
\begin{figure}[ht!]
\centering
\begin{tcolorbox}[colback=gray!5, colframe=black!75]
\footnotesize
\textbf{Scenes 1-4 (Elias \& Sing appear - Initial state):}
\begin{enumerate}[leftmargin=*, itemsep=0pt, topsep=2pt]
\item Elias and Sing lounge on a stained sofa wearing torn undershirts and mismatched flip-flops.
\item Sing slams the table, shouting that they are destined for greatness, not noodles.
\item Elias looks down at his empty bowl, a spark of sudden, desperate greed in his eyes.
\item Elias grabs Sing's collar and yells that they must find the Magic Master to change their lives.
\end{enumerate}

\textbf{Scenes 5-14 (Characters absent - 10 scenes):}
\begin{enumerate}[leftmargin=*, itemsep=0pt, topsep=2pt, start=5]
\item A wide shot reveals a room thick with expensive cigar smoke where gamblers shout and shove chips.
\item The Rich Street Boy walks in, slamming a stack of heavy gold bars onto the green felt table.
\item The boy screams a challenge at the empty dealer's chair, his voice echoing through the hall.
\item The camera pans to the top of the grand stairs, revealing the Master with a cigarette dangling from his lip.
\item The Master descends the staircase slowly, the smoke trailing behind him like a silk ribbon.
\item He stops halfway, leaning over the gold-leaf railing to stare down at the Street Boy.
\item The Master reaches the table and sits, the leather chair creaking under his weight of authority.
\item The Master spreads a card deck in a perfect, lightning-fast rainbow arc across the felt.
\item The Rich Street Boy bluffs, sweat dripping off his chin as the Master stares him down.
\item With a flick of his wrist, the Master reveals the winning card, ending the game instantly.
\end{enumerate}

\textbf{Scenes 15-40 (Characters reappear - State changed):}
\begin{enumerate}[leftmargin=*, itemsep=0pt, topsep=2pt, start=15]
\item Elias and Sing stand by a pillar in the room, now wearing oversized, poorly-fitted tuxedos with crooked ties.
\item Sing tries to look dignified but accidentally trips over his own overly-long trouser hem.
\item Elias whispers urgently, his face pale and eyes twitching with desperate hope.
\item The duo walks toward the Master's table, bowing so low their foreheads nearly hit the floor.
\item The Master looks at the duo and flicks his cigarette ash directly onto Elias's shoe.
\item Sing opens his mouth to speak but the Master raises one finger, silence falls instantly.
\item The Master deals three cards face-down, then looks at them with complete disinterest.
\item Elias reaches for a card but the Master slaps his hand away without even looking.
\item The Rich Street Boy snickers and tosses a single coin at Sing's feet mockingly.
\item Sing's face flushes red with shame, his fists clenching at his sides.
\item The coin rolls across the floor, everyone's eyes following it in tense silence.
\item It stops at the Master's foot. He crushes it flat.
\item Sing suddenly drops to both knees, forehead touching the floor in a full kowtow.
\item Elias hesitates, then joins him, both men prostrated before the Master's chair.
\item The entire casino goes silent, even the roulette wheel stops spinning.
\item The Master stands up slowly, his chair scraping loudly against the marble floor.
\item He walks around them in a circle, examining them like livestock at a market.
\item The Master stops, picks up the flattened coin from under his shoe.
\item He flips it high into the air without warning.
\item Sing's eyes track the coin. He lunges and catches it mid-air with desperate speed.
\item The Master's expression doesn't change, but he nods once barely perceptible.
\item He drops a business card on Sing's back: 'Kitchen. Tomorrow. 5 AM. Don't be late.'
\item The Bodyguard opens the door as the Master walks away without another word.
\item The crowd erupts in confused chatter as Elias and Sing remain frozen on the floor.
\item Outside in the rain, Sing takes the card from his back and stares at the card, then at Elias both soaking wet and shivering.
\item Elias grins stupidly and Sing nods slowly as they argue about who gets to hold the card.
\end{enumerate}
\end{tcolorbox}
\caption{\textbf{Example 5 minute (40 scenes) scenario from \dataset{}, Character State Evolving}: Characters appear (scenes 1-4), disappear for 10 scenes (5-14), then reappear with evolved states (15-40).}
\label{fig:example_5m_scenario}
\end{figure}
%%%%%%%%%%%%%%%%%%%%%%%%%%%%%%

%%%%%%%%%%%%%%%%%%%%%%%%%%%%%%
\clearpage
\begin{tcolorbox}[colback=gray!5, colframe=black!75, breakable]
\footnotesize
\textbf{Scenes 1-5 (Lantern room - Morning routine):}
\begin{enumerate}[leftmargin=*, itemsep=0pt, topsep=2pt]
\item The lantern room features crystal-clear windows and polished brass gears under a bright, cloudless morning sky.
\item The lighthouse keeper wipes a stray smudge off the massive glass lens.
\item A seagull perches on the exterior railing, visible through the lantern room glass.
\item The keeper checks his pocket watch and notes the time in a small leather logbook.
\item Sunlight reflects off the brass machinery, casting bright spots across the lantern room floor.
\end{enumerate}

\textbf{Scenes 6-15 (Lighthouse base - Supply delivery \& storm warning):}
\begin{enumerate}[leftmargin=*, itemsep=0pt, topsep=2pt, resume]
\item Down at the lighthouse base, the heavy iron door stands open to the salty breeze.
\item A supply boat docks at the stone pier nearby, tossing ropes to the keeper's assistant.
\item The assistant carries crates of oil and food toward the lighthouse base.
\item The keeper's orange tabby cat darts into the lighthouse base, searching for shadows.
\item The assistant stacks the heavy crates against the curved stone wall of the lighthouse base.
\item Waves begin to chop more aggressively against the rocks surrounding the lighthouse base.
\item The sky turns a hazy gray as the assistant looks up from the lighthouse base entrance.
\item He begins to haul the first oil canister into the center of the lighthouse base.
\item The assistant closes the heavy iron door of the lighthouse base as the wind picks up.
\item A radio on a small table in the lighthouse base crackles with a storm warning.
\end{enumerate}

\textbf{Scenes 16-20 (Lantern room - Storm preparation):}
\begin{enumerate}[leftmargin=*, itemsep=0pt, topsep=2pt, resume]
\item The lantern room now has a slight layer of condensation on the glass and the sky outside is dark gray with gathering clouds.
\item The keeper pours fresh oil into the lighting mechanism of the lantern room.
\item The wind whistles through the small vents at the top of the lantern room.
\item The keeper strikes a long match, his face illuminated in the dim lantern room.
\item A first drop of rain strikes the exterior of the lantern room window.
\end{enumerate}

\textbf{Scenes 21-30 (Spiral staircase - Assistant's climb):}
\begin{enumerate}[leftmargin=*, itemsep=0pt, topsep=2pt, resume]
\item On the spiral staircase, the assistant slowly climbs the narrow stone steps with a flashlight.
\item The flashlight beam bounces off the damp walls of the spiral staircase.
\item He pauses to catch his breath on a small landing of the spiral staircase.
\item The sound of the assistant's boots echoes rhythmically up the spiral staircase.
\item Rainwater begins to seep through a tiny crack in the masonry of the spiral staircase.
\item The assistant continues his ascent, the spiral staircase feeling tighter and steeper.
\item He wipes sweat from his forehead as he navigates the curve of the spiral staircase.
\item The metal handrail of the spiral staircase feels cold and slippery under his grip.
\item A distant roll of thunder vibrates through the stones of the spiral staircase.
\item The assistant reaches the top hatch leading away from the spiral staircase.
\end{enumerate}

\textbf{Scenes 31-35 (Lantern room - Storm intensifies):}
\begin{enumerate}[leftmargin=*, itemsep=0pt, topsep=2pt, resume]
\item In the lantern room, the rotating beam is now active and rain is lashing violently against the windows in the darkening afternoon.
\item The keeper watches the beam sweep across the turbulent white-capped waves from the lantern room.
\item A sudden gust of wind makes the glass panes of the lantern room rattle in their frames.
\item The keeper adjusts the rotation speed, his shadow dancing across the lantern room walls.
\item Lightning flashes, momentarily blinding the keeper inside the lantern room.
\end{enumerate}

\textbf{Scenes 36-45 (Lighthouse base - Storm preparation):}
\begin{enumerate}[leftmargin=*, itemsep=0pt, topsep=2pt, resume]
\item Back at the lighthouse base, the floor is now wet with tracked-in mud and the room is lit by a single flickering electric bulb.
\item The assistant searches a cabinet in the lighthouse base for emergency flares.
\item Water begins to wash over the stone pier outside the lighthouse base.
\item The assistant secures the internal bolts on the iron door of the lighthouse base.
\item The radio in the lighthouse base is now emitting only static.
\item The orange tabby cat huddles under the supply crates in the messy lighthouse base.
\item The assistant grabs a heavy raincoat from a hook in the lighthouse base.
\item The sound of the ocean crashing against the lighthouse base becomes a deafening roar.
\item The assistant prepares a thermos of hot coffee in the dim lighthouse base.
\item He looks up at the ceiling of the lighthouse base, listening to the storm's fury above.
\end{enumerate}

\textbf{Scenes 46-50 (Lantern room - Electrical failure):}
\begin{enumerate}[leftmargin=*, itemsep=0pt, topsep=2pt, resume]
\item The lantern room is now engulfed in a full storm at night, with the main beam cutting through sheets of rain and the air thick with the smell of ozone.
\item The keeper struggles to keep his balance in the lantern room as the tower sways slightly.
\item A massive wave sends spray high enough to coat the lantern room windows in salt and foam.
\item Sparks suddenly fly from a control panel in the corner of the lantern room.
\item The main light flickers and dies, plunging the lantern room into darkness before emergency red lamps activate.
\end{enumerate}

\textbf{Scenes 51-60 (Spiral staircase - Emergency ascent):}
\begin{enumerate}[leftmargin=*, itemsep=0pt, topsep=2pt, resume]
\item The spiral staircase is now pitch black except for the assistant's swaying flashlight beam.
\item The assistant hurries up the spiral staircase, his breathing heavy and panicked.
\item A small stream of water is now flowing down the steps of the spiral staircase.
\item The assistant trips on a slick step but catches himself on the spiral staircase railing.
\item He shouts the keeper's name, his voice echoing hollowly up the spiral staircase.
\item The flashlight beam reveals the rising water at the bottom of the spiral staircase.
\item The assistant reaches the upper section of the spiral staircase, his clothes soaked through.
\item He pushes against the heavy hatch at the top of the spiral staircase with all his strength.
\item The metal of the spiral staircase groans under the pressure of the wind outside.
\item The assistant finally emerges from the spiral staircase into the upper deck.
\end{enumerate}

\textbf{Scenes 61-65 (Lantern room - Fighting the fire):}
\begin{enumerate}[leftmargin=*, itemsep=0pt, topsep=2pt, resume]
\item The lantern room is now filled with smoke from the shorted electrical panel and the windows are cracked, letting in freezing rain and wind.
\item The keeper uses a manual crank to keep the emergency light turning in the smoke-filled lantern room.
\item The assistant enters the lantern room and immediately grabs a fire extinguisher.
\item Together, they fight the small electrical fire in the corner of the lantern room.
\item The emergency red light reflects off the jagged cracks in the lantern room glass.
\end{enumerate}

\textbf{Scenes 66-75 (Lighthouse base - Storm aftermath):}
\begin{enumerate}[leftmargin=*, itemsep=0pt, topsep=2pt, resume]
\item At the lighthouse base, the iron door is partially buckled from repeated wave impacts and the floor is covered in several inches of seawater.
\item The supply crates in the lighthouse base have been knocked over by the force of the vibrations.
\item The orange tabby cat has climbed to the top of the highest shelf in the flooded lighthouse base.
\item The assistant arrives back at the lighthouse base to survey the damage.
\item He begins using a hand pump to clear the water from the lighthouse base.
\item The morning sun begins to peek through the buckled door of the lighthouse base.
\item The assistant finds the logbook floating in the puddle on the lighthouse base floor and rescues the shivering orange tabby cat from the shelf.
\item He begins to organize the debris in the ruined lighthouse base.
\item The wind has died down to a whisper around the exterior of the lighthouse base.
\item The assistant opens the door of the lighthouse base to see a calm, blue sea.
\end{enumerate}

\textbf{Scenes 76-80 (Lantern room - Morning after):}
\begin{enumerate}[leftmargin=*, itemsep=0pt, topsep=2pt, resume]
\item The lantern room is now calm in the bright morning light, with shattered glass on the floor and puddles of rainwater reflecting the sunrise.
\item The keeper sits on a stool in the lantern room, staring out at the horizon.
\item He begins the long process of sweeping up the glass shards in the lantern room.
\item The assistant enters the lantern room with two mugs of steaming coffee.
\item They both stand in the damaged lantern room, watching the first ship of the day pass safely by.
\end{enumerate}
\end{tcolorbox}
\captionof{figure}{\textbf{Example 10 minute (80 scenes) scenario from \dataset{}, Environment State Evolving}: The lantern room, lighthouse base, and spiral staircase appear and disappear across non-consecutive scenes, with each location's state evolving naturally through a storm sequence.}
\label{fig:example_10m_scenario}
% \end{figure}
%%%%%%%%%%%%%%%%%%%%%%%%%%%%%%

\section{Methodology Prompts} \label{appdx:prompts}

\subsection{\memory{} Initialization Prompts} \label{appdx:init-prompts}

\subsubsection{\memory{} Initialization: Reasoning, \Cref{longveo-eq:memory-initializations}}

\begin{lstlisting}
You are an expert cinematographer. Deeply analyze this video story:

User Prompt: 
{user_prompt}
    
Scenes:
{scenes}

Deep Analysis Task:
1. Determine the key characters and objects that will be shown in the video from the user prompt. Write a concise description of how each one originally looks. Focus on only the key characters and objects with motion.

For each character or object, provide:
- 'name': A unique identifier (e.g., "The Protagonist", "Red Silk Dress").
- 'description': A brief description alone without any other details (e.g., appearance, clothing, colors, style) consistent with relevant backgrounds.

2. Generate the list if entities strictly as a JSON object: 
```json
[{{"name": "...", "description": "..."}}]
```

3. Next, reflect back to the storyline. Are there any dynamic objects or characters in the storyline that were missed?

4. Finally, refine the list of characters and objects with their short descriptions as a JSON array.

Reasoning:
1. ...
2. ...  
3. ...
4. ...
```json
[{{"name": "...", "description": "..."}}]
```
\end{lstlisting}
\vspace{3mm}

\begin{lstlisting}
You are given a set of named visual references (characters, objects, environments) for a video.
For each reference, identify what single other reference it depends on for visual consistency during image synthesis.

References:
{json.dumps(refs, indent=2)}

Guidelines:
- Environments/backgrounds are often roots, but may depend on other environments.
- Characters typically depend on their primary environment.
- Objects typically depend on the environment or character they are associated with.
- No cycles allowed.

Return strictly as JSON list:
```json
[
  {{"name": "ref_name", "depends_on": "other_ref_name or null"}},
  ...
]
```  
\end{lstlisting}
\vspace{3mm}

\begin{lstlisting}
You are an expert spartial cinematographer. Deeply analyze this video story:

User Prompt: 
{user_prompt}
    
Scenes:
{scenes}

Deep Analysis Task:
Step 1. First, identify what are the main environments that the movie takes place in.

Step 2. Then, for each single environment, list what objects, furniture, and elements MUST be inside this space that facilitate all the scenes in the story smoothly. Do not combine multiple environments into one.

Step 3. Next, reflect back to the storyline. For all scenes within/related to that environment, is there anything missing from that environment?

Step 4. Finally, refine the list of environments with key and detailed descriptions of necessary things only.

Step 5. Return your step-level reasonings and deep analysis in detailed text form.

Notes: Do not create unnecessary/transition environments. Stick closely to the story context and planned scenes.

Reasoning:
1. ...
2. ...  
3. ...
4. ...
5. ...

```json
[{{"environment_name": "...", "detailed_spartial_description": "description of a single unified environment with all 
necessary elements"}}]
```
\end{lstlisting}

One note here is that we require the environments to be comprehensive at the global references, as most of environments' details are static. Meanwhile, entity descriptions are minimal at this stage to facilitate state evolution; their specific details are supplemented during the frame prompting steps.

\subsubsection{\memory{} Initialization: Environment Synthesis}

\begin{lstlisting}
Generate a single unified image viewing this entire environment with all its elements following the {env_description}. All elements must be in one cohesive scene, not split into multiple sub-images or panels. Do not include any humans unless explicitly specified.
\end{lstlisting}

\subsubsection{\memory{} Initialization: Entity Synthesis}

\begin{lstlisting}
Generate a professional, high-fidelity, single image of {name}: {description}. A single, centered, full image, cinematic lighting strictly on a solid, clean, all-white background. If {name} is an object, exclude all human or animal faces unless it's explicitly required.
\end{lstlisting}

\subsection{\model{} Prompts} \label{appdx:method-prompts}

\subsubsection{Prompt for Adaptive Segment Generation Mode (All Scenes), \Cref{longveo-eq:generation-mode}} \label{appdx:mode-prompt}

\begin{lstlisting}
Analyze the following video scenes and segment them into contiguous groups.

Scene Indices: {scene_indices}

Scenes:
{scenes}

A contiguous segment is a group of consecutive scenes that:
1. Share the same physical environment/location (e.g., interior vs exterior are different environments).
2. Are temporally continuous (no time jump between them).
3. Have action that flows directly from one scene to the next.

Segmentation Rules (Evaluate in order):
- Priority 1 (Moving Environments): In a moving environment (characters/subjects are moving, e.g., walking, driving, running), zoom in/out or framing changes remain continuous. 
- Priority 2 (Static Environments): In a static environment (characters/subjects remain in the same place), a significant zoom in/out or framing change (e.g., wide shot to close-up) starts a new segment. This explicitly forces the current segment to end in a static state.
- Hard Boundaries: A new segment always starts when the environment/location changes fundamentally, or there is a time jump. Note that interior and exterior are considered different environments even if related (e.g., inside car vs outside car).

Reason over all the scenes one-by-one first, then return as JSON (list of lists of scene indices):

Reasoning:...

```json
[[...], [...], ...]

Note: Every scene index from {scene_indices} must appear exactly once.
\end{lstlisting}

\subsubsection{Prompt to Obtain $\mathcal{V}_{\text{rel}}$ (All Scenes), \Cref{longveo-eq:retrieve}} \label{appdx:contiguous-scenes}

\begin{lstlisting}
Analyze the following video scenes and determine which previous scene (if any) is spatially and temporally contiguous with each scene.

Scenes:
{scenes}

For each scene, determine if there is AT MOST ONE previous scene that is contiguous with it. Two scenes are contiguous if:
1. They share the same physical environment/location
2. They are temporally continuous (no time jump between them)
3. The action flows directly from one to the other

If no previous scene is contiguous, return empty string "" for that scene.

Return as JSON:
```json
{{
  "0": "",
  "1": 0,
  "2": "",
  "3": 2,
  ...
}}
```

Note: Scene 0 always has empty string. Only select AT MOST ONE contiguous scene (the most recent one if multiple exist).
\end{lstlisting}

\subsubsection{Prompt to Obtain $\mathcal{S}_{\text{rel}}$ (All Scenes), \Cref{longveo-eq:retrieve}} \label{appdx:retrieve-relevant-scenes}

\begin{lstlisting}
Analyze the following video scenes and determine which PREVIOUS scenes contribute to the visual appearance of each scene.

Scenes:
{scenes}

For each scene, use chain-of-thought reasoning to identify the TOP 10 most relevant PREVIOUS scenes (with lower indices) for visual consistency.

Step 1: Reasoning
For each scene, think through:
- What objects appear in this scene? Which previous scenes first showed these objects?
- What characters appear in this scene? Which previous scenes established these characters?
- What physical spatial environment is this scene in? Which previous scenes showed this physical spatial arrangement?
- Which scenes are most visually important for maintaining consistency?

Step 2: Selection
Based on your reasoning, select the top 10 most relevant previous scenes for each category:
1. Objects - which previous scenes show objects that appear in this scene
2. Characters - which previous scenes show characters that appear in this scene  
3. Environment - which previous scenes show the same physical spatial arrangement and layout

Return as JSON:
```json
{{
  "reasoning": {{
    "0": "Scene 0 reasoning...",
    "1": "Scene 1 reasoning...",
    ...
  }},
  "relevant_scenes": {{
    "0": {{"objects": [], "characters": [], "environment": []}},
    "1": {{"objects": [0], "characters": [0], "environment": [0]}},
    "2": {{"objects": [0, 1], "characters": [1], "environment": [0, 1]}},
    ...
  }}
}}
```

Note: 
- Scene 0 always has empty lists.
- Each scene can only reference PREVIOUS scenes (lower indices).
- Each category MUST always include at most three immediately previous scenes/
- Limit to TOP 10 most relevant scenes per category, prioritizing most recent and most visually important.
\end{lstlisting}

\subsubsection{Prompt to Retrieve Relevant Global References (All Scenes), \Cref{longveo-eq:retrieve-global-refs}} \label{appdx:retrieve-global-references}

\begin{lstlisting}
Analyze the following video scenes and determine which reference entities and environments are relevant to each scene.

Scenes:
{scenes}

Reference Entities and Environments:
{json.dumps([f"{name}: {anchors[name]}" for name in anchor_names], indent=2)}

For each scene, identify which reference anchors (characters, objects, environments) appear or are relevant to that scene.

Return as JSON:
```{{
  "0": ["anchor_name1", "anchor_name2"],
  "1": ["anchor_name1", "anchor_name3"],
  ...
}}```

Note: 
- Only include anchors that are actually present or relevant in each scene.
- Anchor names must be selected from "{list(self.identity_anchors.keys())}".
\end{lstlisting}

\subsubsection{Prompt to Retrieve the Best End-of-Shot Frame for Scene Continuation, $\text{MLLM}^{\text{img}}_{\text{retr}}$, \Cref{longveo-eq:interpolate-extrapolate}} \label{appdx:retrieve-previous-shot}

\begin{lstlisting}
You are a video continuity judge. Given end-of-shot frames ({image_mapping}) from a multi-shot video and a new scene context, determine which frame is the best starting point for the new scene.

Original video context (contiguous scene):
{old_caption}

New scene context:
{curr_scene_description}

For each frame, analyze whether the subject, motion, and environment can naturally lead into the new scene. Then decide which frame (if any) is the best continuation point.

Return JSON: 
```json
{{"best_index": <{index_options} or null if none>, "reasoning": "..."}}
```
\end{lstlisting}

\subsubsection{Prompt to Generate Frame Prompts, $\text{MLLM}^{\text{img}}_{\text{pgen}}$ in \Cref{longveo-eq:interpolate-extrapolate}} \label{appdx:gen-frame-prompt}

We implement a two-step process for the $\text{MLLM}^{\text{img}}_{\text{pgen}}$ function in \Cref{longveo-eq:interpolate-extrapolate}. In the first step, we generate all frame prompts based on the ``All Scenes'' level to capture inter-scene dependencies. In the second step, during autoregressive generation, we refine each frame prompt given the memory context to integrate relevant historical details. The prompts are provided below.

\vspace{3mm}
\begin{lstlisting}
You are an expert Cinematographer creating visually consistent frames for a cohesive video story. Given the scenes, environment descriptions, generate detailed image prompts for the beginning frame of each scene and an ending frame for the final scene only.

Scenes: 
{scenes}

Reference Anchor Descriptions:
{ref_frame_descriptions}

Visual Consistency Guidelines: 
1. **Characters:** Maintain identical appearance (face, hair, body type, age) across all scenes unless the story explicitly describes changes. Clothing may change if narratively justified (e.g., changing for work, evening wear).
2. **Visual Style:** Establish consistent artistic direction, color grading, and cinematographic approach across all frames.
3. **Environments:** Each scene's setting must align with the reference environments. Spatial layout and static elements of recurring locations must remain identical.
4. **Camera Angles:** When continuing from a previous scene, preserve the identical camera angle. You must explicitly say: "This scene maintains the exact camera angle as scene(s) X...".
5. **Lighting:** If the current scene is a continuation of a previous one within the short period of time,  preserve the identical lighting conditions. Example: "The scene has the exact same lighting condition as scene(s) Y..."
6. **Strict Scene Adherence:** You MUST strictly follow the Scenes provided. DO NOT change:
   - Camera shot types (wide/medium/close-up/etc) specified in the scene
   - Actions or events described in the scene
   - Objects or props mentioned in the scene
   - Lighting conditions specified in the scene
   You may ONLY add visual details that enhance the scene WITHOUT contradicting any specifications.
7. **First Appearance Rule:** When a character or significant object appears for the FIRST time in the video, the frame prompt MUST include ALL explicit and implicit appearance details identified in Step 1 below, even if they are not the primary focus of that scene. This complete description serves as the visual reference for all subsequent scenes.
8. **Character Presence Rule:** If an established character is logically present in a scene, explicitly describe them in the frame even if the scene description focuses on an object or environment.

Each frame prompt must explicitly describe the following while strictly preserving all specifications from the original scene description:
- Visual style and artistic direction.
- Natural lighting effects that reflect the scene's temporal context (morning, noon, evening, etc).
- Character and object descriptions.
- Environment specifications matching reference descriptions.
- Camera angle details.

Step 1: Scene frame reasoning: Reasoning about the scenes one by one about entity and environment appearances to ensure smooth transitions and strict visual consistency.

For each scene, analyze:
1. **Explicit Character/Object/Environment Appearances:** What physical attributes are directly stated in the scenes? (e.g., "wearing a red jacket", "holding a coffee cup")
2. **Implied Appearances:** What can be logically inferred about the characters or objects based on the subsequent scenes? (e.g., if a character "adjusts their glasses" in scene 1, they must be wearing glasses in scene 0 (unless explicitly stated otherwise)).
3. **First Appearance Descriptions:** For the scene that a character or object or environment's first appearance or state changes, establish detailed visual descriptions that will serve as the reference for all subsequent scenes.

Step 2: Return a JSON dictionary with this structure. Scene indexes begin from 0:
```json
{{"frame_prompts": [{{"scene_index": 0, "begin_frame": "detailed image prompt with consistency reasoning"}}, {{"scene_index": {final_scene_index}, "end_frame": "detailed image prompt for final scene with consistency reasoning"}}]}}
```
Step 1 (at least 250 words):...
Step 2:...
\end{lstlisting}
\vspace{3mm}

\begin{lstlisting}
Analyze the video states and reference images to refine the current frame prompt for physical plausibility.

Scenes: 
{relevant_scenes}

Current Scene Description (Scene {scene_index}):
{curr_scene_description}

Current Frame Prompt:
{curr_prompt}

Video States Memory:
{relevant_video_states}

Reference Images:
{image_mappings}

Analyze and reason about:

Step 1. **Scene Analysis**: Read all the provided information and analyze the following one-by-one (at least 100 words each):
   - Is the scene contiguous with the previous scene? (Check if they share the same environment and have continuous action flow)
   - What is this scene trying to display according to the scene description?
   - What visual elements (objects, characters, environment) must be carried forward from reference images and Video States Memory for consistency?
   - What entities (characters and objects) must be carried forward from reference images and Video States Memory for consistency?
   - What relative positional and directional states in the Video States Memory must be strictly maintained in this scene?

Step 2. **Analyze Current Prompt - Camera Angle & Physical Plausibility**: Analyze the following one-by-one (at least 100 words each):
    - Where was the camera positioned relative to the main entities in the previous scene? 
    - This camera angle can be wrong. Under this camera angle, can all the relative positional and directional states identified in Step 1 be maintained absolutely?

Step 3. **Other Camera Angle Possibilities**: Analyze the following one-by-one (at least 100 words each):
   - If the current camera angle cannot maintain the strict positional states identitfied in Step 1, analyze alternative camera angles that must maintain them:
   - For each viable alternative (front view, left/right side, rear view), analyze:
     * Does this angle maintain visual continuity and smooth transition with the previous scene's camera position?
     * How smooth would the transition be from the previous camera angle to this angle?
     * What backgrounds and environments would be visible from this angle?
     * What entities would be visible vs hidden or obstructed?
     * What are the relative spatial relations among entities from this angle:
       - Relative entity orientations (which direction each entity faces relative to others)
       - Relative camera position (where camera is positioned relative to main subjects)
       - Relative positions between entities (distances and arrangements)
       - Depth layers (foreground/midground/background organization)
   
   - If the current camera angle maintains smooth continuity with previous scenes, skip this step and proceed to Step 4.

Step 4. **Refine the Prompt**. Discuss each of the following in at least 100 words:
   - If the scene is contiguous from the previous scene, explicitly add the instruction "Maintain exact camera angle and relative positional/directional states with Scene X"
   - Only change required elements that need correction for physical plausibility and consistency
   - Keep the rest of the original prompt unchanged to preserve the intended narrative and style
   
   - MUST specify the following spatial details in the prompt that must be consistent with Video States Memory:
        * Relative entity orientations (which directions the entities are facing relative to each other or environment features)
        * Relative camera position (camera position relative to the main subjects)
        * Relative positions between entities (with approximate distances if relevant, e.g., "the woman is 5 meters to the left of the car")
        * Depth layer organization (what's in foreground/midground/background)
    - For spatial issues, avoid adding instructions such as "facing to the right/left of the camera" as these are often impossible to control and can cause more issues. Focus on the relative to the environment or other entities instead, such as "the car is backed in with the rear facing the door" or "the car is on the left side of the road". This must be specific and descriptive enough.

- Refined prompt: MUST specify what environmental entities will be visible, will NOT be visible in the frame, and the spatial details.

Step 5. **Refine your refined prompt**: 
    - Make sure all the spatial details are cleary specified and consistent with the Video States Memory and reference images.

Provide the reasonings for above steps, and then finally return the refined prompt in the following JSON format:
Respond with JSON:
```
{{
  "refined_prompt": "the improved prompt"
}}
```
\end{lstlisting}

\subsubsection{Prompt to Synthesize Frames, TI2I in \Cref{longveo-eq:interpolate-extrapolate}} 

\begin{lstlisting}
You are an expert Cinematographer. Given the following storyline from user:

Scenes: 
{relevant_scenes}

Generate an image frame with strict visual consistency with reference images:
{final_frame_prompt}

Ensure Narrative Progression with the reference images as specified in the scenes. The reference images are provided in order with the following mapping: {reference_image_indexes}.
\end{lstlisting}

\subsubsection{Prompt to Generate Video Prompts, $\text{MLLM}_{\text{pgen-vid}}$ in \Cref{longveo-eq:video-syn}} \label{appdx:gen-vid-prompt}

\begin{lstlisting}
You are a professional video producer. Generate a detailed video prompt for a {scene_length}-second video{"that transitions from the beginning frame to the ending frame" if end_frame_path else "starting from the beginning frame"}.

Scenes:
{relevant_scenes}

Current Scene (Scene {scene_idx}):
{current_scene}

Video States Memory (for continuity):
{relevant_video_states}

Previous Scene Video Prompt (for camera/motion continuity):
{previous_video_prompt if previous_video_prompt else "None (this is the first scene)"}

Your task is to generate a comprehensive, narrative prompt for the next video segment that meaningfully progresses the story from the existing prompts and storyline.

Step 1. Read the scenes and understand the overall narrative arc.
Step 2. Start EXACTLY from the beginning frame (Image 1)
Step 3. {"End EXACTLY at the ending frame (Image 2). Visually inspect Image 2 and explicitly describe its composition, entity states, and environment in the video prompt as the final state the video must reach." if end_frame_path else "End naturally to best fulfill the Current Scene description"}
Step 4. Maintain continuity from Previous Video States and Previous Scene Video Prompt:
   - Entity states: character appearance, clothing, accessories must remain consistent
   - Environment states: ground texture, landmarks, lighting must remain consistent
   - Motion states: ongoing actions, camera movement direction, and environmental motion (e.g., train vibrations, car movement, boat rocking) must continue naturally from where the previous scene's prompt left off
   - CRITICAL: Camera direction must NOT reverse or contradict the previous scene. If the previous scene's camera was moving/orbiting in a certain direction, this scene must continue in the same direction or come to a natural stop, never reverse.
   **Explicitly specify these continuing elements in your video prompt**
Step 5. Ensure narrative/story logical progression: the actions and events in this scene must make sense as a continuation of what happened in the previous scenes. The characters' activities, positions, and interactions should flow naturally from the previous scene's ending state.
Step 6. If the scene involves multiple distinct actions or camera transitions, break the description into temporal segments (e.g., "Scene 1: ... Scene 2: ... Scene 3: ...") to clearly specify what happens in each part. 
   - IMPORTANT: If the scene involves showing content on a phone/tablet/screen, do NOT describe it as "camera zooms into the phone". Instead, treat it as two separate temporal segments: Segment 1 describes the character holding/interacting with the device, Segment 2 describes the screen content as if it fills the entire frame (a direct cut to the screen, not a zoom).

Image Mapping: [Begin Frame{", End Frame" if end_frame_path else ""}]
{'' if end_frame_path else 'IMPORTANT: There is NO end frame. Do NOT reference "Image 2" or any end frame constraint in the video prompt.'}

In the video prompt, cover the following:
{self.SCENE_PROMPT_TEMPLATE}

Example: 
Example: 
A professional 8-second video featuring [the entities]. 
The video opens with [opening frame description]... The [entities' activities]...
{"Finally, the video concludes with [closing frame description]..." if end_frame_path else "The video progresses naturally to fulfill the scene description..."}
[Any motions that must be continue from Previous Video States for continuity if applicable]...
The camera [camera movement/angle description, maintaining continuity with Previous Video States if applicable].
...

Respond with JSON:
```json
{{
  "video_reasoning": "explain your reasoning for how you structured the video prompt and whether temporal segments are needed",
  "video_prompt": "your detailed video prompt"
}}
```
\end{lstlisting}

\subsubsection{Prompt to Synthesize Video Segments, TI2V in \Cref{longveo-eq:video-syn}}
We use the generated prompt in above step to synthesize the video segment.

\subsection{HITS Self-Refinement Prompts: MLLM-Judge}

\subsubsection{Prompt to Extract Frame States,  $\text{MLLM}^{\text{img}}_{\text{ext}}$ in HITS, \Cref{subsec:memory-augmented-algorithms}} \label{appdx:extract-states-from-frame-prompt}

\begin{lstlisting}
Analyze this frame and extract the states of all entities.

Current Scene Description (Scene {scene_index}):
{curr_scene_description}

Frame Prompt:
{frame_prompt}

Instructions:
1. Identify every entity (characters, objects, environments) in the frame.

2. For each entity, extract:
   - type: "character", "object", or "environment"
   
   - identity: Features and appearance that identify this entity across scenes:
     * Physical features (body type, facial structure, distinctive marks)
     * Clothing items and colors
     * Hair style and color
     * Accessories
     * For objects: shape, design, color, material
     * For environments: architectural features, layout, defining characteristics
   
   - state: Current observable state that may change between scenes:
     * Position: Relative position to other entities ONLY (e.g., "on the left side of X", "behind/perpendicular to Y", "between X and Y"). DO NOT use frame-relative terms like "center", "foreground", "left side of frame".
     * Orientation: Which direction the entity faces relative to other entities or environment features ONLY (e.g., "facing toward the door", "facing away from X", "facing the same direction as Y"). DO NOT use frame-relative terms like "facing left", "facing the camera".
     * Depth: Relative depth to other entities (e.g., "closer to camera than X", "in front of Y", "behind Z")
     * Posture/arrangement (for characters: standing/sitting, limb positions; for objects: open/closed, arrangement)
     * Condition (clean/dirty, intact/damaged, etc.)
     
   Be specific with colors (exact shades) and directions. All positions and orientations must be relative to other entities, NOT to the frame or camera.

3. For spatial_relations, describe how entities relate spatially to each other.

Return the scene graph in this exact JSON format:
```json
{{
  "entities": {{
    "entity_name": {{
      "type": "character|object|environment",
      "identity": "features and appearance that identify this entity",
      "state": "current observable state",
    }}
  }},
  "spatial_relations": [
    {{"subject": "entity1", "relation": "relation_type", "object": "entity2"}}
  ]
}}
```
\end{lstlisting}

\subsubsection{Prompt to Extract Video States,  $\text{MLLM}^{\text{vid}}_{\text{ext}}$ in HITS, \Cref{subsec:memory-augmented-algorithms}} \label{appdx:extract-states-from-video-prompt}

\begin{lstlisting}
Analyze this video segment comprehensively. Provide at least 400 words total across all sections.

Current Scene Description (Scene {scene_index}):
{curr_scene_description}

Known Entity States (from begin frame):
{entities_json}

Extract the following from the full video segment:

1. **Supplementing Missing Elements**
- Identify any NEW entities (characters, objects, environments) that appear after the begin frame but were not in the known entity states.
- For each new entity, provide a full appearance description.

2. **Identity Changes**
- For all entities (known and new), describe any appearance details revealed throughout the video that were not visible in the begin frame (e.g., full body, accessories, clothing details, object condition).
- For characters: focus on hair, clothing, physical appearance.
- For objects: focus on color, shape, condition, visible attributes.
- For environments: focus on lighting, weather, background elements, and spatial arrangements.

3. **Motions**
- For each element (known and new), describe its motion using the template: "At the beginning, [...]. In the middle, [...]. At the end, [...]."

4. **Camera Dynamics**
- Describe camera movement, angle changes, zoom, and panning throughout the video using the same temporal template.

Return as JSON:
```json
{{
 "new_entities": {{
   "<entity_name>": "full appearance description"
 }},
 "identity_changes": {{
{entities_template}
 }},
 "motions": {{
{entities_template}
 }},
 "camera": "At the beginning, [...]. In the middle, [...]. At the end, [...].",
}}
```
\end{lstlisting}

\subsubsection{Prompt to Judge Frame - Consistency over Images (Entity, Environment, Narrative), \Cref{longveo-eq:frame-verifications}} \label{appdx:frame-judge-prompt}

\begin{lstlisting}
Evaluate the generated frame against reference images for consistency and progression.

Scenes:
{relevant_scenes}

Current Scene (Scene {scene_index}):
{scenes[scene_index]}

Frame Prompt:
{frame_prompt}

Image Mapping:
{image_mapping}

What's wrong with the current image compared to its references with mappings specified in Image Mapping?  

3. Entity Reference Consistency (1-10): First, identify which reference images (by Image number and label) show the same characters or objects that appear in the current frame. Then, using only those relevant references, check if any entity violates their appearance. If there are visual conflicts (e.g., object appearance changes, character look changes, clothing/design changes, distinctive feature changes) that are NOT explicitly described in the Frame Prompt, this is a violation. Compare each character and object one-by-one:
   - Focusing on every characters (emphasizing appearance, cloths, hair, accessories, etc) and objects (emphasizing furniture, devices, etc) visible in the frame. For each one, identify which reference images show that same entity, then check the EXACT appearance, colors, clothing/design, distinctive features against those references
   - State what matches and what differs from the reference images
Be very critical and detailed. Describe in detail in at least 200 words.

4. Environment Reference Consistency (1-10): First, identify which reference images (by Image number and label) show the same environment/location as the current frame. Then, using only those relevant references, check if the environment violates its appearance. If there are spatial conflicts (e.g., position inconsistencies, spatial arrangement changes, architectural changes) that are NOT explicitly described in the Frame Prompt, this is a violation:
   - Focusing on every spatial layout, architectural elements, lighting, overall setting
   - State what matches and what differs from the reference images
Be very critical and detailed. Describe in detail in at least 200 words.
        
5. Narrative Progression (1-10): What's wrong with the logical progression of the current frame compared to its prior frame regarding the entities, the environment, and logical lighting? The frame must NOT be identical or nearly identical to previous reference frames unless the Frame Prompt explicitly describes a static or unchanged scene. There must be visible change in entity positions/actions, environmental elements, or lighting conditions that aligns with the storyline progression. Lighting must progress consistently with time: if the previous frame shows late afternoon, the current frame must be late or even later unless the story explicitly involves a time jump. If the frame looks like a duplicate or shows no meaningful progression in entities or environment when the Frame Prompt indicates change should occur, this is a violation. Discuss in details in at least 250 words.

Respond with JSON:
```json
{{
  "entity_reference_consistency": <score 1-10>,
  "environment_reference_consistency": <score 1-10>,
  "narrative_progression": <score 1-10>,
  "reasoning": "detailed explanation of consistency and progression issues"
}}
```
\end{lstlisting}

\subsubsection{Prompt to Judge Frame - Consistency over Images (Spatial Logicalness), \Cref{longveo-eq:frame-verifications}} \label{appdx:frame-judge-prompt-2}

\begin{lstlisting}
Carefully examine the spatial arrangements and object positions across these images.

Scenes:
{relevant_scenes}

Current Scene (Scene {scene_index}):
{scenes[scene_index]}

Frame Prompt:
{frame_prompt}

Image Mapping:
{image_mapping}

Note: Images labeled "anchor_*" show the canonical spatial layout and object arrangements. The current frame MUST match these anchor layouts unless the Frame Prompt explicitly describes changes.

What's wrong with the current image compared to existing images regarding the spatial environment? Reasoning in at least 250 words.

Respond with JSON:
```json
{{
  "spatial_logicalness": <score 1-10, where 10 means nothing wrong>,
  "reasoning": "detailed analysis of spatial issues or 'nothing wrong'"
}}
```
\end{lstlisting}

\subsubsection{Prompt to Judge Frame - Textual States, \Cref{longveo-eq:frame-verifications}} \label{appdx:frame-judge-prompt-3}

\begin{lstlisting}
Compare the extracted states with relevant previous scenes and the storyline to identify discrepancies.

Scenes:
{relevant_scenes}

Current Scene (Scene {scene_index}):
{scenes[scene_index]}

Video States Memory:
{relevant_video_states}

Current Scene States:
{video_states}

Instructions:
Reasoning in at least 250 words to critically analyze consistency smartly:

Step 1: **Analyze Objects**
- What objects appear in the current scene's extracted states?
- How do these objects compare to their states in relevant previous scenes?
- What discrepancies exist (if any)? Note that if new objects appear in this scene that weren't in earlier scenes, and there's no narrative storyline for their appearance, this counts as a discrepancy.
- Are these discrepancies justified by the current scene description?
- Note that environmental changes due to natural video progression (e.g., spatial relationships changing as character moves, new buildings coming into view) should NOT be counted as discrepancies. Only flag static attributes that must be consistent.

Step 2: **Analyze Characters**
- What characters appear in the current scene's extracted states?
- How do these characters compare to their states in relevant previous scenes?
- What discrepancies exist (if any)? Note that if new characters appear in this scene that weren't in earlier scenes, and there's no narrative storyline for their appearance, this counts as a discrepancy.
- Are these discrepancies justified by the current scene description?

Step 3: **Analyze Environment**
- What physical spatial environment appears in the current scene's extracted states?
- How does this physical spatial environment compare to its state in relevant previous scenes?
- Strictly spot any spatial discrepancies or illogical environmental arrangements exited (if any). Note that if the physical spatial environment remains the same but new lights, atmosphere, or other environmental elements appear in this scene that weren't in earlier scenes, and there's no narrative storyline for their appearance, this counts as a discrepancy.
- Are these discrepancies justified by the current scene description? Normally, most of the spatial discrepancies are not justified unless explicitly described in the scene.

Step 1 (at least 200 words):...
Step 2 (at least 200 words):...
Step 3 (at least 200 words):...

Return as JSON:
```json
{{
  "objects": "detailed reasoning about objects consistency...",
  "objects_state_score": <score 1-10>,
  "characters": "detailed reasoning about characters consistency...",
  "characters_state_score": <score 1-10>,
  "environment": "detailed reasoning about environment consistency...",
  "environment_state_score": <score 1-10>
}}
```
\end{lstlisting}

\subsubsection{Prompt to Judge Frame - Basic Quality, \Cref{longveo-eq:frame-verifications}} \label{appdx:frame-judge-prompt-4}

\begin{lstlisting}
Evaluate if the generated frame successfully matches the following prompt and meets quality standards:

Frame Prompt:
{frame_prompt}

Current Frame States:
{video_states}

Evaluate and score each criterion on a scale of 1-10:

1. Instruction Following (1-10): Examine every single detail in the prompt. Does the frame faithfully capture every critical component specified in the prompt, especially lighting conditions and environmental details? Discuss in details.
2. Physical Plausibility (1-10): Is the frame physically realistic? All objects, characters, and environments must appear physically plausible and follow real-world physical laws unless the frame prompt specifies otherwise. The frame must represent one unified moment in space and time. NO "cinematic license" or artistic adjustments should excuse physical impossibilities. NO split screens, panels, or multiple disconnected scenes. NO collage-style layouts showing different moments or locations. Discuss in details.

Respond with JSON:
```json
{{
  "instruction_following": <score 1-10>,
  "physical_plausibility": <score 1-10>,
  "reasoning": "explanation of any issues found in detail"
}}
```
\end{lstlisting}

\vspace{5mm}
\subsubsection{Prompt to Judge Video - Inter Consistency, \Cref{longveo-eq:vid-verifications}} \label{appdx:vid-judge-prompt-1}

\begin{lstlisting}
Analyze event consistency between two contiguous scenes.

Scenes:
{relevant_scenes}

Previous Contiguous Scene (Scene {previous_scene_idx}):
{previous_scene}

Current Scene (Scene {current_scene_idx}):
{scenes[scene_index]}

Previous Contiguous Scene Video States:
{pre_contiguous_video_states}

Current Scene Video States:
{video_states}

Instructions:
These two scenes are spatially and temporally contiguous (same environment, continuous action flow). Compare the states of the current video versus previous video states to spot any inconsistencies regarding the states, motions, and camera:

Step 1: **Analyze Entity Appearance Consistency (at least 100 words)**
- For each entity in the current scene, check if it's new or appeared in the previous scene
- Are new entities properly justified by the scene description?
- For entities appearing in both scenes, do their visual appearance and states remain consistent?

Step 2: **Analyze Entity Motion Consistency (at least 100 words)**
- For entities appearing in both scenes, compare their motions at the end of Scene {previous_scene_idx} with the beginning of Scene {current_scene_idx}
- Do the motions form a smooth transition regarding activities and directions?
- Identify any abrupt changes in entity motions that aren't justified by the scene descriptions

Step 3: **Analyze Environment Appearance Consistency (at least 100 words)**
- Compare the environment layout and lighting at the end of Scene {previous_scene_idx} with the beginning of Scene {current_scene_idx}
- Does the spatial layout (room structure, architectural elements) remain consistent?
- Does the lighting (intensity, direction, color temperature) remain consistent?
- Note: Natural environmental changes due to video progression are acceptable (e.g., different parts of the same location becoming visible as characters move, background scenery shifting due to camera/character movement, gradual lighting changes)
- Only flag SIGNIFICANT inconsistencies: sudden teleportation to different locations, drastic lighting jumps without narrative justification, or architectural impossibilities
- Since these scenes are contiguous, major environment changes should be justified by the scene descriptions

Step 4: **Analyze Camera Consistency (at least 200 words)**
- Compare camera movement at the end of Scene {previous_scene_idx} with the beginning of Scene {current_scene_idx}
- Since these scenes are contiguous, the camera should maintain continuity - do they form a consistent flow?
- Identify any jarring camera transitions that break visual continuity.

Return as JSON:
```json
{{
  "entity_consistency_reasoning": "detailed analysis of entity appearance consistency",
  "entity_consistency_score": <score 1-10>,
  "entity_motion_reasoning": "detailed analysis of entity motion consistency",
  "entity_motion_score": <score 1-10>,
  "environment_consistency_reasoning": "detailed analysis of environment appearance consistency",
  "environment_consistency_score": <score 1-10>,
  "camera_reasoning": "detailed analysis of camera consistency",
  "camera_score": <score 1-10>
}}
```
\end{lstlisting}

\subsubsection{Prompt to Judge Video - Intra Consistency and Basic Video Quality, \Cref{longveo-eq:vid-verifications}} \label{appdx:vid-judge-prompt-2}

\begin{lstlisting}
Evaluate the generated video against the scene description.

Current Scene:
{scenes[scene_index]}

Current Video States:
{video_states}

Grading Guidelines (apply to all criteria):
- 1-3: Major issues with main entities/objects appearance (e.g. wrong character, wrong outfit, object replaced, impossible physics, jarring cut, different hairstyle)
- 4-6: Noticeable but minor issues with main entities/objects appearance (e.g. slight color shade difference, minor texture variation, small accessory missing)
- 7-10: Main entities/objects appearance is perfect; only background or negligible details may differ

Step 1: Understand the video: Watch the video carefully and read the 'Current Video State' to fully understand what happened in the video - what entities are present, how they look, how they move, and how the camera behaves.

Step 2: Judge against the scene description. Critically flag any error at each criterion on a scale of 1-10. For each criterion, provide at least 150 words of detailed analysis:

1. Instruction Following (1-10): Watch the video. Does it match the scene description? Check character appearance, entity appearance, character motions, entity motions, background and camera against the scene description as ground truth.

2. Physical Plausibility (1-10): Watch the video. Is it physically realistic? Check character physics, entity physics, visual artifacts, and lighting consistency purely from what you observe in the video.

3. Narrative Progression (1-10): Watch the video. Does it flow smoothly and continuously as a single professional shot?

4. Frame Fit (1-10): Watch the video. Does the story logically end with the provided end frame? The end frame is the last image in the prompt, describe it in detail, then judge whether the video leads naturally to it.

5. Character Consistency (1-10): Do NOT watch the video. Compare `identity` vs `identity_changes` in the 'Current Video State' for each character. Any mismatch not explicitly justified by the scene description is an inconsistency. Apply the grading guidelines based on severity of the mismatch.

6. Object Consistency (1-10): Do NOT watch the video. Compare `identity` vs `identity_changes` in the 'Current Video State' for each object. Any mismatch not explicitly justified by the scene description is an inconsistency. Apply the grading guidelines based on severity of the mismatch.

7. Environment Consistency (1-10): Do NOT watch the video. Compare `identity` vs `environment_changes` in the 'Current Video State' for each environment entity. Any mismatch not explicitly justified by the scene description is an inconsistency. Apply the grading guidelines based on severity of the mismatch.

Respond with the following format:

Step 1: Understand the video: Describe what you observe in the video and the 'Current Video State'...
Step 2: Judge each criterion with detailed analysis (at least 150 words each)...

Return the final scores in JSON format:
```json
{{
  "instruction_following": <score 1-10>,
  "physical_plausibility": <score 1-10>,
  "logical_progression": <score 1-10>,
  "frame_fit": <score 1-10>,
  "character_consistency": <score 1-10>,
  "object_consistency": <score 1-10>,
  "environment_consistency": <score 1-10>
}}
```
\end{lstlisting}

\subsection{MAPO Prompts} \label{appdx:mapo-prompt}

\subsubsection{Prompt to Reason Over Feedback (Frame Prompt)}

\begin{lstlisting}
You are analyzing an image generation prompt that failed validation.

Scene Description (MUST be fully satisfied):
{scene_description}

Original Prompt: {current_prompt}

Validation Scores: {scores}

Validation Feedback: {feedback}

Tasks:
1. READ AND UNDERSTAND the scene description. This is the ground truth that MUST be fully satisfied.

2. READ AND UNDERSTAND the original prompt in detail. What is it trying to achieve? What are ALL the key elements?

3. READ AND UNDERSTAND the validation feedback in detail. What specific issues were identified? Minor issues that that human can't see or barely notice can be disregarded.

4. ANALYZE DISCREPANCIES BETWEEN PROMPT AND GENERATION:
   - Where did the generation fail? Which parts of the prompt were missed or misinterpreted?
   - List specific elements described in the prompt that were not correctly generated
   - Are we drifting away from the scene description?

5. PROMPT REASONING AND REFINEMENT (at least 250-1000 words): This step is all about prompt reasoning and we must not discuss anything about the generation.
    
    Structure your analysis as follows:
    
    **Problematic phrases/parts:** For each issue discovered above, identify which phrases or instructions are causing the issues. Are there any ambiguities or contradictions in the prompt that could lead to misinterpretation?
    
    **What to add/clarify:** What specific clarifications or conflicts would resolve the ambiguities and help the model understand the intent better?
    
    **What to remove/simplify:** Are there overly specific constraints, redundant details, or non-essential elements that should be removed or simplified?

Example format:
    - **Problematic:** "The scientist holds a vial with a precise two-finger grip" - This is overly specific and impossible for the model to control exactly...
    - **Add/Clarify:** Specify that "the scientist carefully holds a small vial" to convey the careful handling without impossible constraints...
    - **Remove/Simplify:** Remove the detailed finger position requirements ("only distal pads of thumb and index finger") as these add complexity without improving the core scene...

Notes: 
    - This step 5 reasoning is critical: analyze every detail in the prompt that could cause the identified issues.
    - Focus entirely on prompt content and structure - do not mention generation quality or results in this step.
    - Pay special attention to the Scene Description and Physical Plausibility of the prompt's instructions.
    - Use the world "could cause" rather than "is causing" to avoid making assumptions about the generation - we are only analyzing the prompt here, not the generation results.

    
6. SPECIAL NOTES FOR SPATIAL/ENVIRONMENTAL INCONSISTENCIES:
    If the feedback mentions entity location or direction inconsistencies (e.g., "mannequin disappeared", "machine moved to wrong side", "table is now behind instead of left"), add explicit positional constraints to the refined prompt such as "[entity] remains at [location]" (e.g., "in the background", "on the table") or "[entity] stays [direction] of [reference]" (e.g., "left of the window", "behind A").
    Note: Avoid adding instructions such as facing to the right/left as these are often impossible to control and can cause more issues. Focus on the relative to the environment or other entities instead, such as "the car is backed in with the rear facing the door" or "the car is on the left side of the road". This must be specific and descriptive enough.

7. SPECIAL NOTES FOR CAMERA:
    Many times, the errors indicated in the feedback come from an impossible camera angle. In such cases, use a camera angle from the reference frames instead of the current prompt's camera specification.

8. DETERMINE MODE:
    Based on the errors identified, determine if the issues can be fixed with minor edits or require regeneration:
    - "edit": The errors are minor and can be fixed, such as entity states, small adjustments to positioning, or clarifications
    - "regenerate": The errors are major, such as fundamental scene misinterpretation, spatial or environmental or camera issues, missing key elements, or severe physical plausibility issues

Provide your step-by-step reasoning, at least 500 words:
Step 1: ...
Step 2: ...
Step 3: ...
Step 4: ...
Step 5: ...
Step 6: ...
Step 7: ...
Step 8: ...

Then return as JSON:
```json
{{
    "prompt_reasoning": "Copy exactly your Step 5 reasoning here.",
    "mode": "edit or regenerate",
    "edit_prompt": "if mode is edit, provide a one-sentence edit instruction with 'Edit the last reference image via...' (e.g., 'Edit the last reference image via adding the tools from Scene X on the table'). Otherwise, leave empty."
}}
```
\end{lstlisting}

\subsubsection{Prompt to Reason Over Feedback (Video Prompt)}

\begin{lstlisting}
You are analyzing a video generation prompt that failed validation.

Current Scene (MUST be fully satisfied):
{scenes[scene_index]}

Original Prompt: {current_prompt}

Validation Scores: {scores}

Validation Feedback: {feedback}

Tasks:
Step 1. READ AND UNDERSTAND the scene description. This is the ground truth that MUST be fully satisfied.

Step 2. READ AND UNDERSTAND the original prompt in detail. What is it trying to achieve? What are ALL the key elements and actions?

Step 3. READ AND UNDERSTAND the validation feedback in detail. What specific issues were identified?

Step 4. ANALYZE DISCREPANCIES BETWEEN PROMPT AND GENERATION:
   - Character identities: Did any character's appearance (hair, clothing, accessories, physical attributes) change unexpectedly or differ from what the scene description requires? Check ALL characters.
   - Object identities: Did any object's appearance (color, shape, condition, texture) change unexpectedly or differ from what the scene description requires? Check ALL objects.
   - Environment: Did the environment layout, lighting, or background change unexpectedly (sudden lighting shifts, background elements appearing/disappearing, layout changes) in ways not justified by the scene description?
   - Motions/directions: Which actions, movements, or transitions were missed or misinterpreted?

Step 5. PROMPT REASONING AND REFINEMENT (at least 250-1000 words): This step is all about prompt reasoning and we must not discuss anything about the generation.
    
    Structure your analysis as follows:
    
    **Problematic phrases/parts:** For each issue discovered above, identify which phrases or instructions are causing the issues. Are there any ambiguities or contradictions in the prompt that could lead to misinterpretation? For video prompts, pay special attention to:
    - Character identities: Does the prompt explicitly enforce that all characters maintain their appearance (hair, clothing, accessories) throughout the entire video?
    - Object identities: Does the prompt explicitly enforce that all objects maintain their appearance (color, shape, condition) throughout the entire video?
    - Environment: Does the prompt explicitly enforce that the environment layout, lighting, and background remain stable and consistent throughout the entire video?
    - Motions/directions: Are action descriptions clear, sequential, and free of impossible constraints or hyper-specific timing?
    
    **What to add/clarify:** What specific clarifications would resolve the identified issues? For video prompts, consider:
    - Character identities: If any character's appearance changed unexpectedly, explicitly add constraints such as "throughout the entire video, [character] must maintain [appearance]"
    - Object identities: If any object's appearance changed unexpectedly, explicitly add constraints such as "throughout the entire video, [object] must remain [appearance]"
    - Environment: If the environment layout or lighting changed unexpectedly, explicitly add constraints such as "throughout the entire video, the environment/lighting must remain [description]"
    - Motions/directions: Add motion flow, action sequencing, and transition cues to resolve ambiguities
    
    **What to remove/simplify:** Are there overly specific constraints, redundant details, or non-essential elements that should be removed or simplified? For video prompts, consider:
    - Character/object descriptions that are overly detailed and could cause the model to hallucinate inconsistencies
    - Overly specific motion constraints that are impossible to achieve
    - Redundant or conflicting descriptions
    - Technical specifications that could be replaced with simpler visual language

Notes: 
    - This step 5 reasoning is critical: analyze every detail in the prompt that could cause the identified issues.
    - Focus entirely on prompt content and structure — do not mention generation quality or results in this step.
    - Prioritize character identity, object identity, and environment consistency issues before motion issues.
    - Use "could cause" rather than "is causing" — we are only analyzing the prompt, not the generation results.

Provide your step-by-step reasoning, at least 500 words:
Step 1: ...
Step 2: ...
Step 3: ...
Step 4: ...
Step 5: ...

Then return as JSON:
```json
{{
    "prompt_reasoning": "Copy exactly your Step 5 reasoning here."
}}
```
\end{lstlisting}

\subsubsection{Prompt to Extract Contrastive Patterns (Frame Prompt)}

\begin{lstlisting}
Analyze successful and failed frame generation prompt refinements to extract actionable patterns.

Successful Refinements (what worked):
{positive_samples}

Failed Refinements (what didn't work):
{negative_samples}

By comparing the original vs refined prompts, identify what changes led to success or failure.

Extract 8-12 actionable patterns. Each pattern must:
1. Be specific and concrete (not vague advice)
2. Include a brief example (e.g., "use 'X' instead of 'Y'")
3. Be 1-2 sentences maximum

Focus on:
- **Structural changes**: How prompt organization affects results
- **Specificity**: When adding/removing detail improves generation
- **Terminology**: Which words/phrases work vs fail
- **Spatial/visual logic**: How to describe positions, angles, relationships
- **Common failures**: What consistently breaks generation

Respond with JSON:
{{
  "patterns": [
    "Pattern with brief example",
    ...
  ]
}}
\end{lstlisting}

\subsubsection{Prompt to Extract Contrastive Patterns (Video Prompt)}

\begin{lstlisting}
Analyze successful and failed video generation prompt refinements to extract key patterns.

Successful Refinements (what worked):
{positive_samples}

Failed Refinements (what didn't work):
{negative_samples}

Extract actionable patterns across these categories:

1. **Structural Patterns**: Optimal prompt organization for video, length, ordering
2. **Temporal/Motion Descriptions**: Effective ways to describe motion flow, action sequencing, timing
3. **Continuity Instructions**: How to ensure smooth transitions and consistent framing
4. **Specificity vs. Abstraction**: When to be precise vs. general for motion/timing
5. **Conflict Resolution**: Common contradictions in action descriptions, impossible constraints
6. **Domain-Specific Vocabulary**: Effective motion/temporal terms vs. confusing terms
7. **Negative Patterns**: Instructions that consistently fail for video, over-specifications

Return each pattern as a separate item in a list. Each pattern should be a concise, actionable guideline (1-2 sentences).

Respond with JSON:
{{
  "patterns": [
    "Pattern 1 guideline here",
    "Pattern 2 guideline here",
    ...
  ]
}}
\end{lstlisting}

\subsubsection{Prompt to Refine Frame Prompt}

\begin{lstlisting}
You are refining an image generation prompt based on analysis.

Current Scene (MUST be fully satisfied):
{scenes[scene_index]}

Original Prompt: {current_prompt}

Prompt Reasoning (Analysis):
{prompt_reasoning}

Learned Prompting Patterns (apply relevant ones):
{learned_patterns_from_successful_cases}

Refine the prompt targettedly strictly following the guidelines below:
- If the analysis indicates no issues or the prompt is already optimal, return the original prompt unchanged
- Otherwise, modify the specific parts via additions/clarifications and removals/simplifications identified as problematic
- Apply relevant learned patterns to improve the prompt
- Keep everything else from the original prompt unchanged
- For spatial issues, avoid adding instructions such as "facing to the right/left of the camera" as these are often impossible to control and can cause more issues. Focus on the relative to the environment or other entities instead, such as "the car is backed in with the rear facing the door" or "the car is on the left side of the road". This must be specific and descriptive enough.

Return as JSON:
```json
{{
    "refined_prompt": "Your refined prompt. Make only the specific changes identified in the analysis - strictly keep everything else from the original prompt unchanged."
}}
```
\end{lstlisting}

\subsubsection{Prompt to Refine Video Prompt}

\begin{lstlisting}
You are refining a video generation prompt based on analysis.

Current Scene (MUST be fully satisfied):
{scenes[scene_index]}

Original Prompt: {current_prompt}

Prompt Reasoning (Analysis):
{prompt_reasoning}

Learned Prompting Patterns (apply relevant ones):
{learned_patterns_from_successful_cases}

Refine the prompt targettedly strictly following the guidelines below:
- If the analysis indicates no issues or the prompt is already optimal, return the original prompt unchanged
- Otherwise, modify the specific parts via additions/clarifications and removals/simplifications identified as problematic
- Apply relevant learned patterns to improve the prompt
- Keep everything else from the original prompt unchanged

- CRITICAL: Do not change the beginning state of any entity! Their initial position, motion, and condition at the start of the video might be inherited from the previous scene and must be preserved exactly as in the original prompt. For example, if the car is moving at the beginning in the Original Prompt, you must keep the car moving instead of the car is accelerating.

- CRITICAL: By default, use a static camera. Do NOT use zoom, pan, tilt, dolly, crane, or orbital/circling movements unless the scene explicitly requires it. If camera changes are needed, they must be split into separate temporal segments with clear time ranges (e.g., "0-4s: static wide shot... 4-8s: tracking shot following subject...")

Return as JSON:
```json
{{
    "refined_prompt": "Your refined prompt. Make only the specific additions/removals identified in the analysis - strictly keep everything else from the original prompt unchanged."
}}
```
\end{lstlisting}

\subsection{\Cref{subsec:consistency-versus-horizons}'s Prompts} \label{subsec:dfirt-prompts}

\subsubsection{Prompt to Verify Character Consistency Drift, \Cref{longveo-eq:drift-consistency-verifications}}

\begin{lstlisting}
You are evaluating CHARACTER consistency in a long video generation.

The provided images are labeled as follows:
{image_labels}

Scene descriptions:
{past_scenes}

Current scene description (Scene {scene_idx}):
{current_scene}

Follow these steps:

Step 1: Identify visual changes: Compare the character in the reference frames to the character in the current scene frame. Focus on the character's face, hair, identity, body type, and overall appearance.

Step 2: Classify changes: For each difference, determine whether it is expected (justified by the story) or unexplained.

Step 3: Evaluate: Only flag as violated (0) for CLEAR appearance changes: a completely different person appears, the character's face/body type is obviously different, or the character's species changes. Do NOT flag minor variations like slight outfit dirt, accessories, lighting differences, or pose changes.

Output ONLY this JSON:
```json
{{
  "step1_visual_changes": "<list character differences>",
  "step2_classification": "<expected or unexplained for each>",
  "character_consistent": <1 if consistent, 0 if violated>,
  "reasoning": "<brief reason>"
}}
```
\end{lstlisting}

\subsubsection{Prompt to Verify Object Consistency Drift, \Cref{longveo-eq:drift-consistency-verifications}}

\begin{lstlisting}
You are evaluating OBJECT consistency in a long video generation.

The provided images are labeled as follows:
{image_labels}

Scene descriptions:
{past_scenes}

Current scene description (Scene {scene_idx}):
{current_scene}

Follow these steps:

Step 1: Identify visual changes: Compare the key objects in the reference frames to the current scene frame. Focus on the object's identity, shape, color, and size.

Step 2: Classify changes: For each difference, determine whether it is expected (justified by the story) or unexplained.

Step 3: Evaluate: Only flag as violated (0) for CLEAR appearance changes to key objects: object is a completely different item, shape is obviously wrong, or color is drastically different. Do NOT flag minor position changes, slight wear/damage, lighting differences, or partial occlusion.

Output ONLY this JSON:
```json
{{
  "step1_visual_changes": "<list object differences>",
  "step2_classification": "<expected or unexplained for each>",
  "object_consistent": <1 if consistent, 0 if violated>,
  "reasoning": "<brief reason>"
}}
```
\end{lstlisting}

\subsubsection{Prompt to Verify Environment Consistency Drift, \Cref{longveo-eq:drift-consistency-verifications}}

\begin{lstlisting}
You are evaluating BACKGROUND consistency in a long video generation.

The provided images are labeled as follows:
{image_labels}

Scene descriptions:
{past_scenes}

Current scene description (Scene {scene_idx}):
{current_scene}

Follow these steps:

Step 1: Identify visual changes: Compare the background/environment in the reference frames to the current scene frame. Pay close attention to:
- Architectural structure: walls, pillars, arches, ceiling, floor layout
- Spatial arrangement: relative positions of objects, furniture, doorways, ledges

Step 2: Classify changes: For each difference, determine whether it is expected (justified by the story) or unexplained.

Step 3: Evaluate: The background must be the SAME physical space as the reference frames, not just a similar-looking environment. Flag as violated (0) only if the room/location is clearly different: different architecture, different layout, or a different place entirely. Do NOT flag camera angle differences, story-driven events, or minor lighting changes.

Output ONLY this JSON:
```json
{{
  "step1_visual_changes": "<list background differences>",
  "step2_classification": "<expected or unexplained for each>",
  "background_consistent": <1 if consistent, 0 if violated>,
  "reasoning": "<brief reason>"
}}
```
\end{lstlisting}

%%%%%%%%%%%%%%%%%%%%%%%%%%%

\section{Full Storyboard Examples}\label{sec:storyboard-examples}

\subsection{\model{} Examples}

\begin{figure}[h!]
    \centering
\includegraphics[width=.9\textwidth]{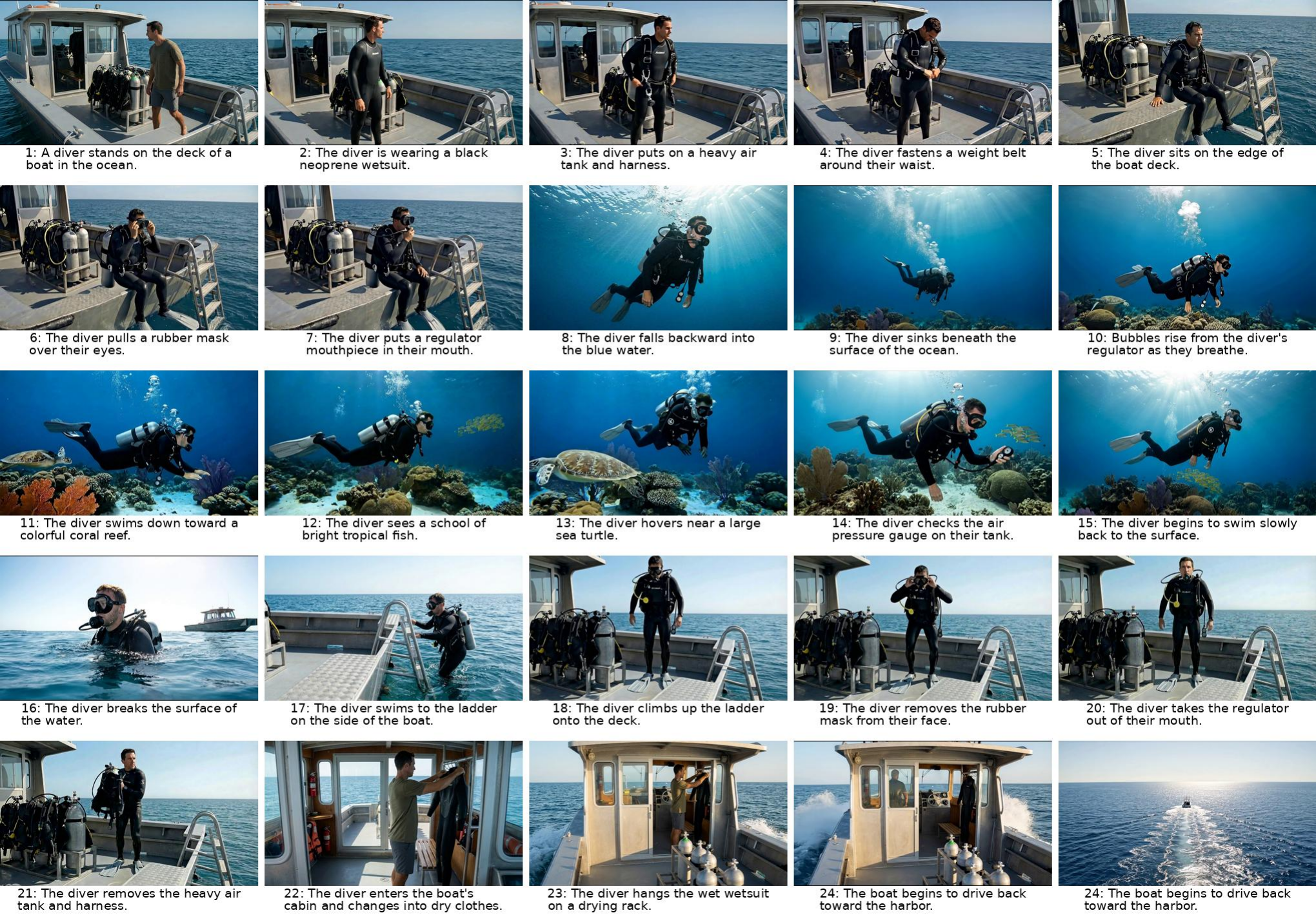}
\caption{Storyboard synthesized by \model{} of a 3-minute video from \dataset{}: \textit{The Scuba Diver's Reef Exploration} (24 scenes).}
\label{fig:3m-example-2}
\end{figure}

\begin{figure}[h!]
    \centering
\includegraphics[width=.9\textwidth]{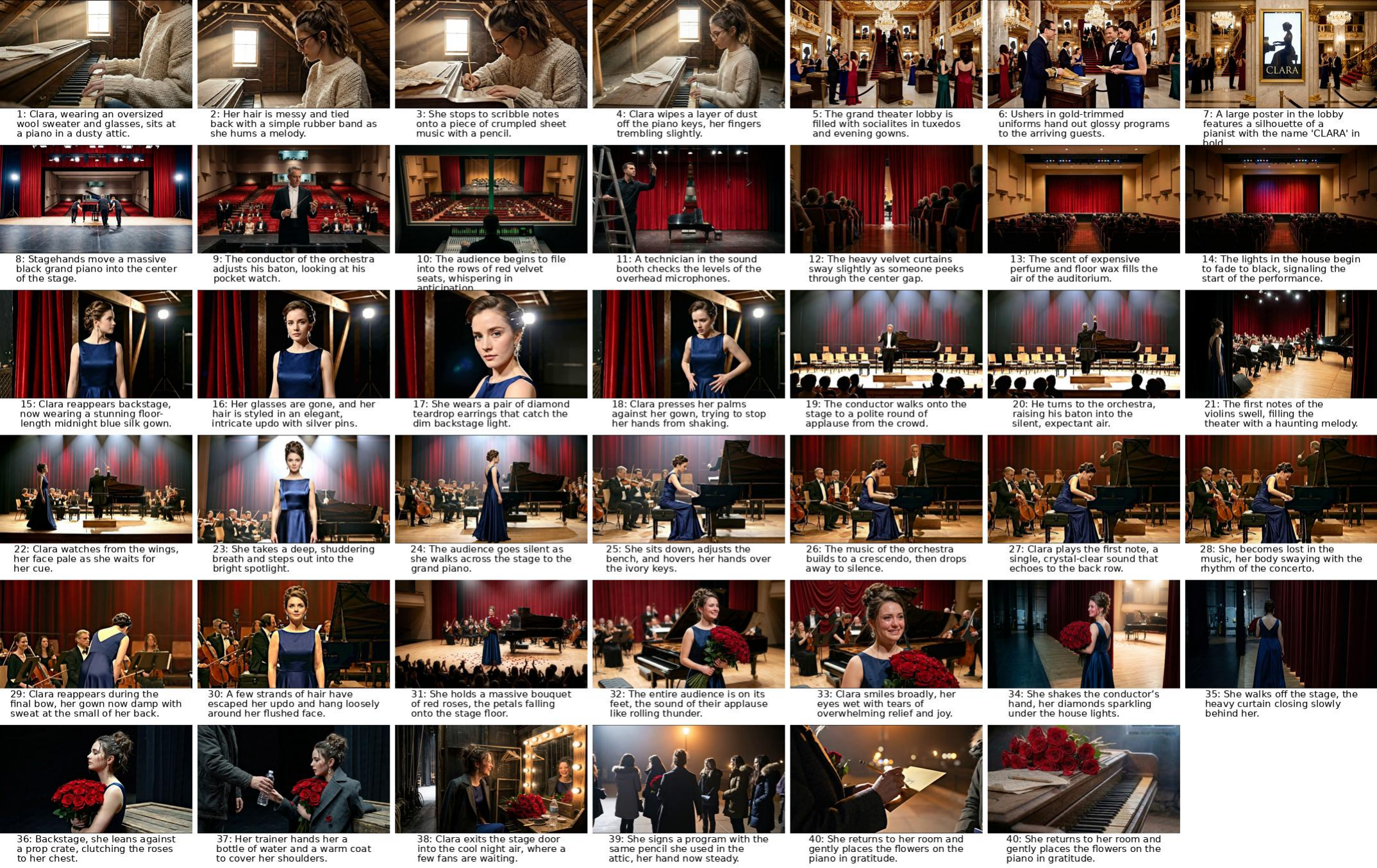}
\caption{Storyboard synthesized by \model{} of a 5-minute video from \dataset{}: \textit{The Stage Fright (Clara)} (40 scenes).}
\label{fig:5m-example}
\end{figure}

\begin{figure}[h!]
    \centering
\includegraphics[width=.9\textwidth]{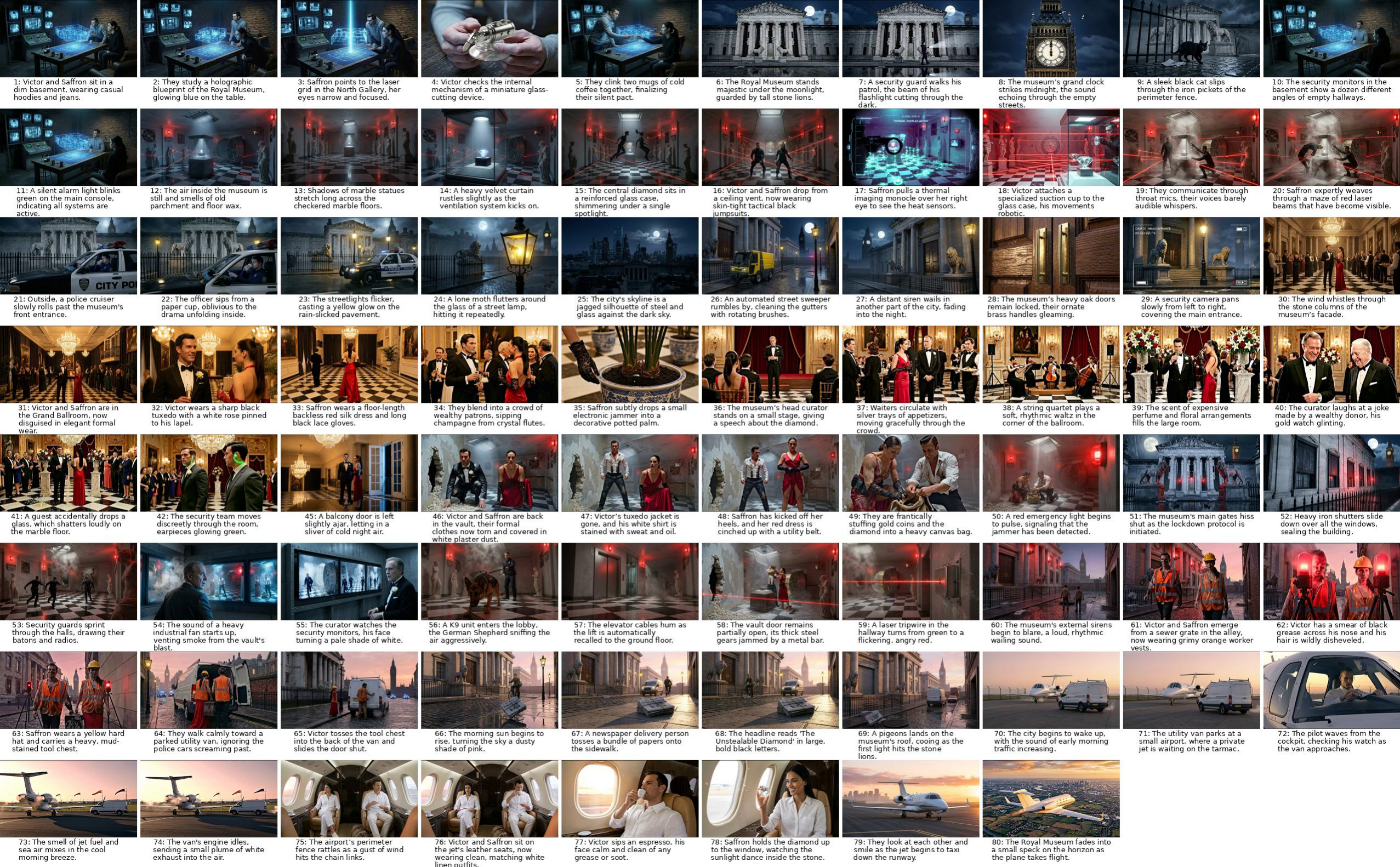}
\caption{Storyboard synthesized by \model{} of a 10-minute video from \dataset{}: \textit{The Great Museum Heist} (79 scenes).}
\label{fig:10m-example}
\end{figure}

\subsection{Baseline Examples} 

\begin{figure}[h!]
    \centering
\includegraphics[width=.9\textwidth]{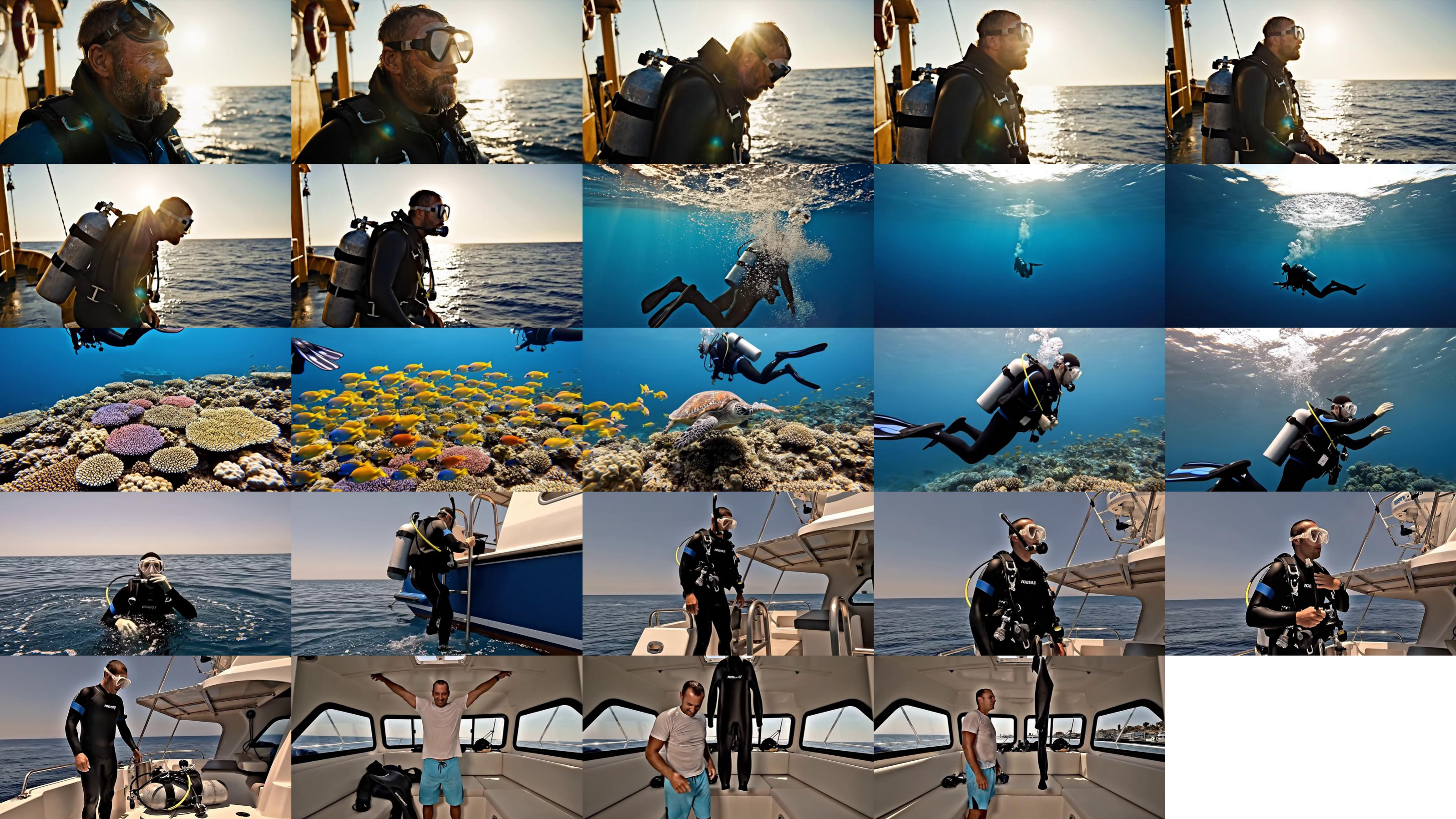}
\caption{Storyboard synthesized by Naive-AR of a 3-minute video from \dataset{}: \textit{The Scuba Diver’s Reef
Exploration} (24 scenes). Frequent inconsistencies in characters, objects, and environments accumulate throughout the video, severely degrading narrative coherence and visual continuity.}
\label{fig:naivear-3m-example}
\end{figure}

\begin{figure}[h!]
    \centering
\includegraphics[width=.9\textwidth]{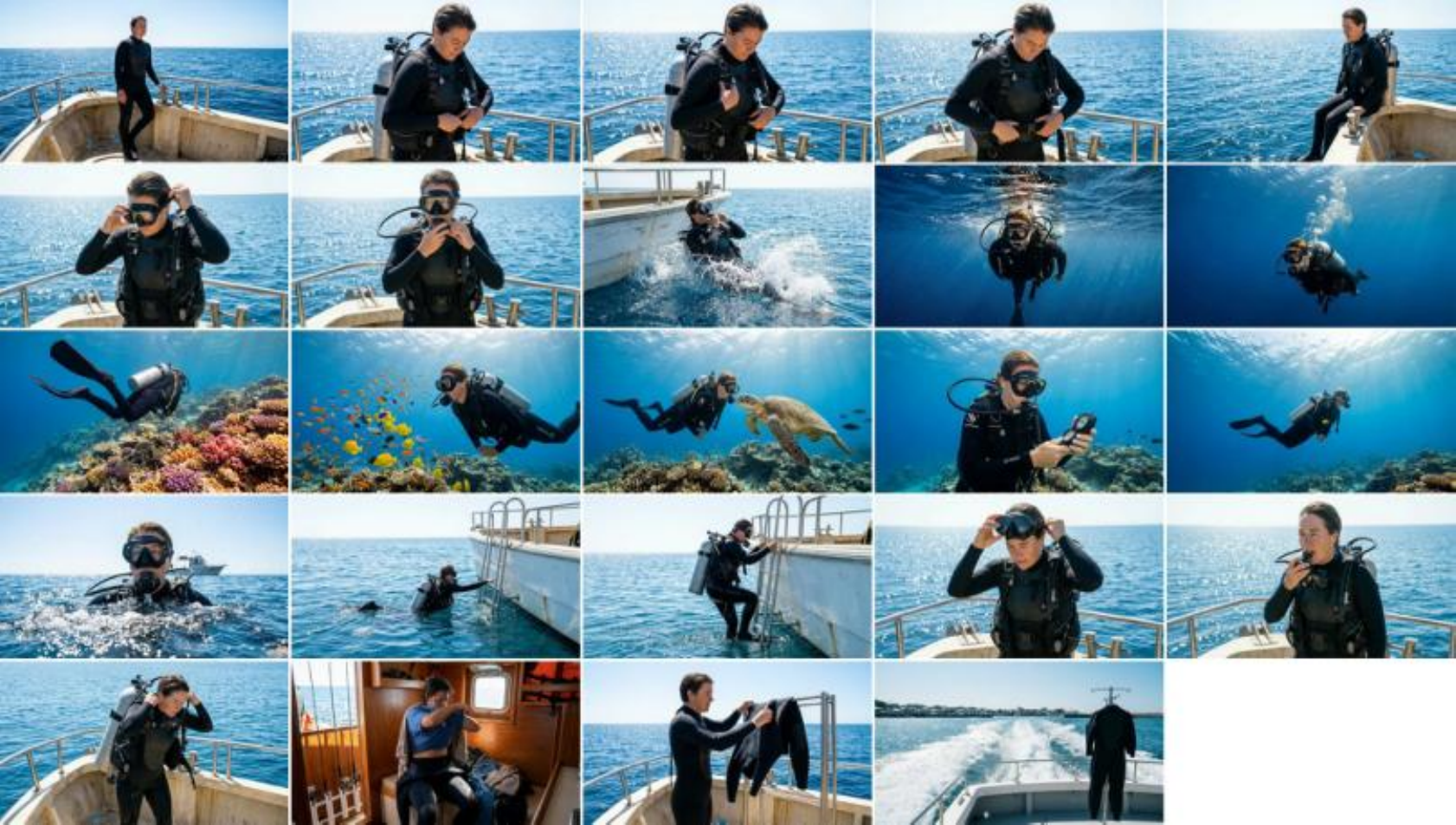}
\caption{Storyboard synthesized by Naive-Par of a 3-minute video from \dataset{}: \textit{The Scuba Diver’s Reef
Exploration} (24 scenes). Frequent inconsistencies in characters, objects, and environments accumulate throughout the video, severely degrading narrative coherence and visual continuity.}
\label{fig:naivepar-3m-example}
\end{figure}

\begin{figure}[h!]
    \centering
\includegraphics[width=.9\textwidth]{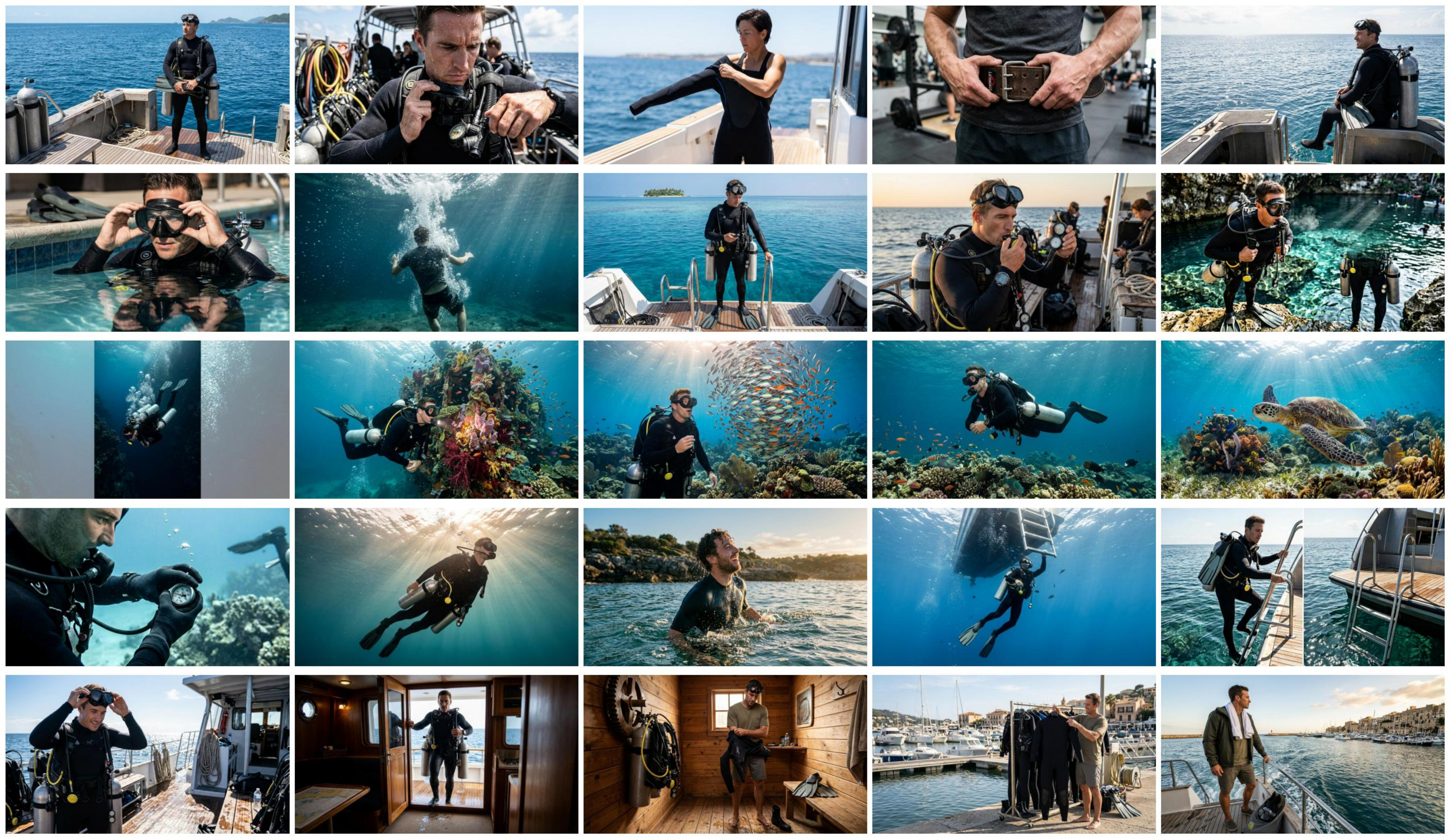}
\caption{Storyboard synthesized by MovieAgent of a 3-minute video from \dataset{}: \textit{The Scuba Diver’s Reef
Exploration} (24 scenes). Frequent inconsistencies in characters, objects, and environments accumulate throughout the video, severely degrading narrative coherence and visual continuity.}
\label{fig:movieagent-3m-example}
\end{figure}

\begin{figure}[h!]
    \centering
\includegraphics[width=.9\textwidth]{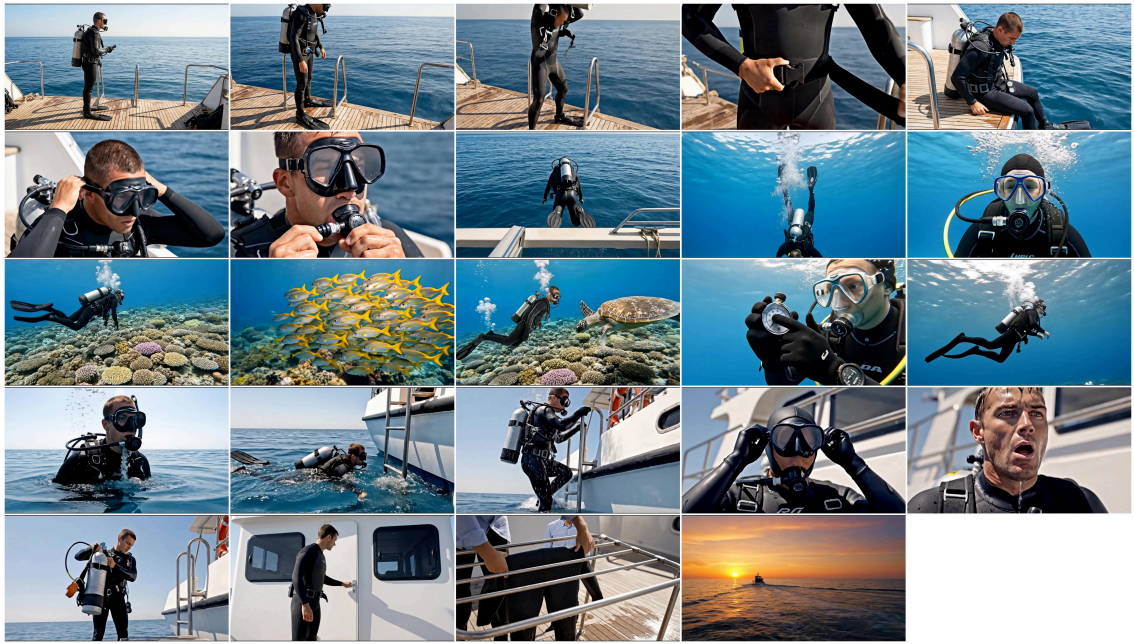}
\caption{Storyboard synthesized by ViMax of a 3-minute video from \dataset{}: \textit{The Scuba Diver’s Reef
Exploration} (24 scenes). Frequent inconsistencies in characters, objects, and environments accumulate throughout the video, severely degrading narrative coherence and visual continuity.}
\label{fig:vimax-3m-example}
\end{figure}

\end{document}